%% file: neurips_2020.tex
\documentclass{article}

 \usepackage[final]{neurips_2020}

\usepackage[utf8]{inputenc} 
\usepackage[T1]{fontenc}    
\usepackage{hyperref}       
\usepackage{url}            
\usepackage{booktabs}       
\usepackage{amsfonts}       
\usepackage{nicefrac}       
\usepackage{microtype}      
\usepackage{bm}             
\usepackage{amsmath}
\usepackage{natbib}
\usepackage{graphicx}
\usepackage{chngpage}
\usepackage{subcaption}
\usepackage[toc,page]{appendix}
\usepackage{algpseudocode}

\usepackage{pgfplots}
\DeclareUnicodeCharacter{2212}{−}
\usepgfplotslibrary{groupplots,dateplot}
\usetikzlibrary{patterns,shapes.arrows}
\pgfplotsset{compat=newest}

\usepackage{algorithm,algcompatible}
\algnewcommand\INPUT{\item[\textbf{Input:}]}%
\algnewcommand\OUTPUT{\item[\textbf{Output:}]}%

\title{Variational framework for partially-measured physical system control: examples of vision neuroscience and optical random media
}

\author{%
  Babak ~Rahmani\thanks{Department of Electrical Engineering.} \\
  EPFL\\
  \texttt{babak.rahmani@epfl.ch} \\
 
   \And
   Demetri Psaltis\\
   EPFL \\
   \texttt{demetri.psaltis@epfl.ch} \\
   
      \And
   Christophe Moser\\
   EPFL \\
   \texttt{christophe.moser@epfl.ch} \\
}

\begin{document}

\maketitle

\begin{abstract}

To characterize a physical system to behave as desired, either its underlying governing rules must be known a priori or the system itself be accurately measured. The complexity of full measurements of the system scales with its size. When exposed to real-world conditions, such as perturbations or time-varying settings, the system calibrated for a fixed working condition might require non-trivial re-calibration, a process that could be prohibitively expensive, inefficient and impractical for real-world use cases. In this work, we propose a learning procedure to obtain a desired target output from a physical system. We use Variational Auto-Encoders (VAE) to provide a generative model of the system function and use this model to obtain the required input of the system that produces the target output. We showcase the applicability of our method for two datasets in optical physics and neuroscience.
\end{abstract}

\section{Introduction}
In physical system characterization, a fundamental challenge is finding the proper continuous space input to a system that yields a desired functional output. For example, an open question in sensory/motor neuroscience is how to determine the input stimulation able to induce a desired behavior. So is controlling the output of an optical system, such as a turbid medium used for imaging, that could be non-linear and time-varying. In a linear physical system, the problem of finding the input that produces a desired output can be determined by monitoring its response to a series of arbitrary inputs and then computing the inverse of the system’s transmission matrix (a mapping from inputs to outputs). This entails measuring the responses of the system fully. In practice, physical systems can only be partially measured and, more importantly, are nonlinear. So the transmission matrix formalism cannot be used. Even though the forward path of the system could be fully characterized, obtaining its inverse for large scale systems involving millions of variables is computationally intensive if not entirely intractable. Hence, resorting to data-driven methods that do not require full-measurements or linear approximation of the system, such as deep learning approaches, is inevitable.
Deep learning techniques proposed  for these tasks \citet{mcintosh2016deep}, \citet{rahmani2018multimode} mostly take advantage of labeled data to do supervised training. For applications that require control over the response of one or an ensemble of targets, end-to-end supervised learning can fail due to the lack of labeled data within the distribution of desired target responses as well as inherent sensitivity of supervised approaches to perturbations in out-of-training-distribution data. Therefore, we instead propose a learning framework based on generative probabilistic models \citet{mirza2014conditional} and in specific, VAEs \citet{kingma2013auto} which involves in construction of a forward estimator of the possibly partially measured system. Once the forward model is obtained, a second estimator is trained to provide the required input of the system for producing the desired output. This latter estimator could be constraint so as to promote certain solutions. Therefore, contributions of this work are as follows:


\begin{itemize}

   \item Using the variational generative models, we provide a training algorithm for learning the distribution of the system's inputs that are needed to obtain a desired output of the system.
   \item Using the sampling feature of the learned forward VAE model, we illustrate how our training algorithm learns to iteratively move towards the correct distribution of the inputs.

\end{itemize}

\paragraph{Related works:}
As opposed to the inference problem of estimating the input of the system from noisy sensory outputs in experimental disciplines such as microscopy \citet{rivenson2017deep}, optical tomography \citet{wurfl2016deep} and neuroscience \citet{parthasarathy2017neural} that supervised deep learning approach is a fairly well-established technique, learning methods for control applications in these fields have yet to be matured. Closed-loop techniques based on deep networks have been proposed for a number of applications, such as for brain neuroscience \citet{bashivan2019neural} wherein authors control the activity of individual neuronal sites in V4 area by optimizing single input stimuli. Likewise, for optical turbid-medium imaging, authors have used ML-based estimators for controlling the optical fields \citet{rahmani2020actor}. 
As opposed to the previous works, we propose joint learning of the forward and backward models of the system with VAEs to implicitly impose compatibility of the sought-after solutions with the underlying physics of the problem. The latter, in essence, is akin to technique of untrained neural networks \citet{van2018compressed}, \citet{ulyanov2018deep}, \citet{heckel2018deep} in denoising and inpainting. 

\section{Generative modelling}
\paragraph{Problem scenario}
In the most general form, we assume that a given input of a system, $x_i$, is mapped to its output via the function $f$ as in $y_i=f(x_i)$. Therein, $x_i\in \mathbb{C}^n$ and $y_i\in \mathbb{C}^m$ in the most general case. All known about $f$ is that it could be a (non)linear time-varying function. We assume that all the noise sources are incorporated in $f$. Additionally, $f$ can be sampled as many times as needed. In other words, exact output of the system, i.e. $y_i$, is available for any given input $x_i$. Yet, $f$ is never measured nor analytically derived. Moreover, $f$ might only be partially measured for which, the input-output relationship is modified to $y_i=\varphi[f(x_i)]$  where $\varphi$ is either identity (fully measured system) or some other functions (for example modulus $|.|^2$ function).

We seek to find the $x_i^*$ that would produce a desired $y_i^*$. It is worth emphasizing that the experimentalist might only have access to the partially measured system while the objective is to obtain the desired output in the fully measured system. The problem, in its most general form, can be formulated as follows.
\begin{equation}
\label{eq:e1}
\mathcal{L}  = \min_{\mathbf{\xi,\zeta}} \, \operatorname{\mathbb{E}}_{x_i,y_i,z}{ \Bigl[D[y_i, M_{\zeta} (x_i,z)]\Bigr]}+\operatorname{\mathbb{E}}_{x_i^*,y_i^*,z}\Bigl[{\sigma [M_{\zeta} (A_{\xi}(x_i^*,y_i^*),z),y_i^*]}\Bigr]
\end{equation}

where $ M_{\zeta}: \mathbb{C}^{n \times l} \rightarrow \mathbb{C}^m $, referred to henceforth as the Model, is a differentiable representation of $f$ parameterized by $\zeta$ and $ A_{\xi}: \mathbb{C}^{n \times m} \rightarrow \mathbb{C}^n $, referred to henceforth as Actor, is a mapping that produces the input for $M_\zeta$. Therein, $D$ is the distance between outputs $y_i$ sampled (experimentally) from $y_i=f(x_i)$  and the output of $M_\zeta$. $\sigma$ is the distance between the desired target $y_i^*$ and predicted output of $M_\zeta$ given the output of $A_\xi$. $z=\{z_{i}\}_{i=1}^{l}$ is the latent space vector of size $l$. The two-term loss function $\mathcal{L}$ is then optimized with respect to the parameters $\zeta$ and $\xi$. The first RHS term in Eq. \ref{eq:e1} is further explained below.

\paragraph{Forward estimator learning}
The forward mapping $M_\zeta$ is estimated as a generative probabilistic VAE. The reason for this choice of model is two-fold. First, forward models that are fundamentally stochastic in nature (see example 2 in Results section) could be better represented by a probabilistic model rather than a ML estimator trained in a supervised learning manner. Additionally, even if $f$ is deterministic, noise sources incorporated into $f$ make it stochastic in practice. Second, the generative sampling feature of VAEs could be conveniently used to demonstrate how the correct control input $x_i^*$ (that is required to generate $y_i^*$) could be obtained iteratively.

The VAE $M_\zeta$ consists of two networks, an encoder and a decoder. The former is trained to transform input $x_i$ conditioned on the system's output $y_i$ onto the latent vector $z$ that is enforced to be close to a normal distribution $\mathcal{N}(0,\textnormal{I})$; effectively learning the conditional distribution $q_\Phi(z|y_i)$ parameterized by $\Phi$. The decoder, on the other hand, takes the latent vector $z$- drawn from the encoder distribution $\mathcal{N}(\mu_{enc},\sigma_{enc})$ using reparameterization trick- to generate output $\hat y_i$; effectively learning the conditional distribution $p_\theta(y_i|z,x_i)$ parameterized by $\theta$. The training of the VAE is carried out by optimizing the following loss function w.r.t. $\zeta : \{\theta,\Phi\}$ \citet{higgins2016beta}.

\begin{equation}
\label{eq:e2}
\mathcal{L}_{M_\zeta}  = \min_{\mathbf{\zeta: \{\theta,\Phi\}}} \, \operatorname{\mathbb{E}}_{x_i\sim \rho(x)}\Bigl[\operatorname{\mathbb{E}}_{z\sim q(.|y_i)}{ [\log [p_\theta(y_i|z,x_i)]]}
-
\beta \operatorname{\mathbb{E}}_{y_i\sim \rho(.|x_i)}{ [D_\textnormal{KL}(q_\Phi(z|y_i)\,||\,\mathcal{N}(0,\textnormal{I}))]}\Bigr]
\end{equation}

where $\beta$ is the weighting factor between the two terms in the loss function and $\rho$ is to denote a general purpose probability distribution.

\paragraph{Training algorithm}
A sketch of the networks and gradient flows is depicted below. Algorithm \ref{alg1} presents the learning procedure for the system control. It involves in computing the variational updates of the forward model followed by training of the backward mapping. We compute an empirical performance metric between the outputs generated through the experimental system by the control inputs provided by the algorithm and the targets and reiterate if the performance is not satisfactory.

\begin{tabular}{cc}
\raisebox{-.1cm}{\begin{minipage}{.8\textwidth}
\input{algorithm1}
\end{minipage}}

&
\begin{minipage}{.4\textwidth}
\includegraphics[width=\textwidth]{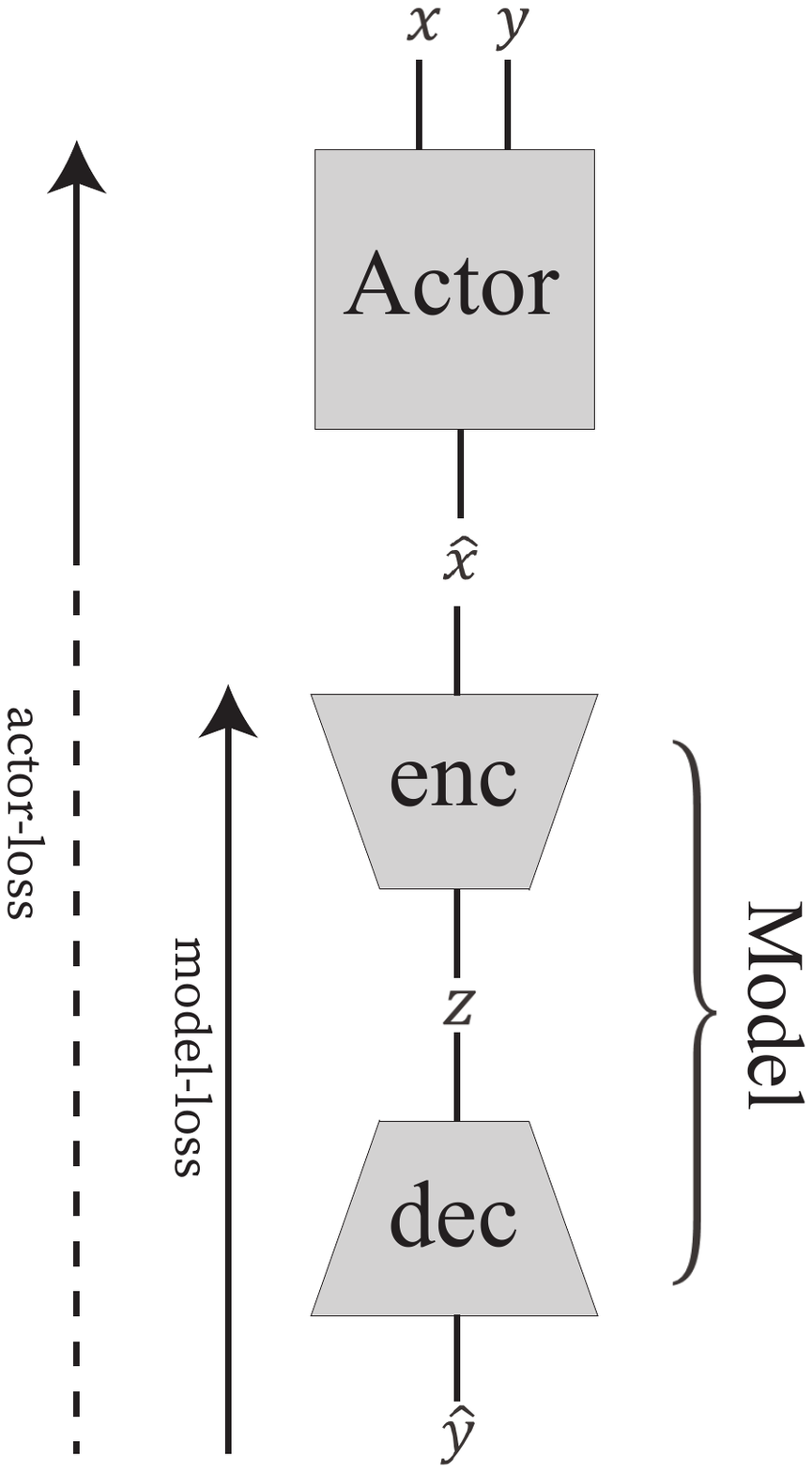}
\end{minipage}

\end{tabular}

\section{Results}

\paragraph{Phase retrieval for optical system control}
The first example involves in characterization of a slowly time-varying, nonlinear physical system featuring random scramblers. The objective of this experiment is to find the appropriate complex input vector of the system, $\textbf{X}^*=\{x_i^* \}$ $\in{\mathbb{C}^n}$, that produces a desired target output, $\textbf{Y}^*= \{ y_{\mu}^*\}$ $\in{\mathbb{R}^m}$, given the partial measurements of the system as in $y_{\mu}^*=\Big|\sum_{i}^{ }F_{\mu i}x^{*}_i\Big|^2$, where $x^{*}_i\!$ (and respectively $y^*_{\mu}$) are elements of the input (output) vector and $F_{\mu i}$ is the complex-value measurement matrix.
Although the problem in essence is a phase retrieval (PR) of the system's input, key differences with the conventional PR settings renders it more challenging. In particular, in the the original PR problem, $F$ is entirely known \textit{a priori}. In the current setting, $F$ is not measured and therefore is unknown. Instead, tuples of an arbitrary input $\textbf{X}$ and its corresponding output $\textbf{Y}$ is available. Secondly, while in the conventional PR, outputs $\textbf{Y}$ (generated via a teacher model) always belong to the support of $F$, the target output $\textbf{Y}^*$ may not belong to the support of $F$ which requires finding the input that produces the closest output to the target in some metric. 
The optimization problem can then be re-written as:

\begin{equation}
\label{eq:e3}
\mathcal{L}  = \min_{\mathbf{\xi,\zeta}} \, \operatorname{\mathbb{E}}_{\textbf{X},\textbf{Y},z}{ \Bigl\Vert \textbf{Y}- M_{\zeta} (\textbf{X},z)\Bigr\Vert^{2}_{l_2}}+\operatorname{\mathbb{E}}_{\textbf{X}^*,\textbf{Y}^*,z}\Bigl[{ \Vert \textbf{Y}^*-M_{\zeta} (A_{\xi}(\textbf{Y}^*),z)\Vert^2_{l_2}}\Bigr]
\end{equation}

where we choose $l_2$ norm for the forward and backward metrics. The network architecture and optimization scheme is further explained in the Appendices. We tested our algorithm with MNIST dataset \citet{cohen2017emnist} as the target outputs. An example of the input-output of the system is shown in Appendix \ref{appendix:A}. Fig. \ref{fig:Figure2} (a) plots empirical $l_2$ norm as well as 2D Pearson correlation between the system's outputs and targets versus the iteration number. It can be seen that the algorithm almost reaches the 2D correlation ($\sim 0.9$) obtained with full-measurement techniques. Examples of the experimentally generated outputs using the proposed algorithm are also provided in Appendix \ref{appendix:A}.

\begin{figure*}[h]
\centering
\begin{tikzpicture}[yscale=0.35,xscale=0.35]

  \input{pgf/loss-fiber}
  \input{pgf/corr-fiber}
  \input{pgf/loss-vision}
  \input{pgf/corr-vision}
  
\begin{scope}[yscale=.75,xscale=.75]
   \input{pgf/emb0}
   \input{pgf/emb1}
   \input{pgf/emb2}
   \input{pgf/emb3}

   \end{scope}
   
  \end{tikzpicture}
  \caption{Performance metric of the algorithm (loss: left axis and Pearson correlation: right axis) versus iteration number for phase retrieval (a) and vision neuroscience task (b). The  Latent vector evolution of the latter task as 2D embedding (blue). Orange dots denote the latent vector of the true system $f$.}
  \label{fig:Figure2}
\end{figure*}
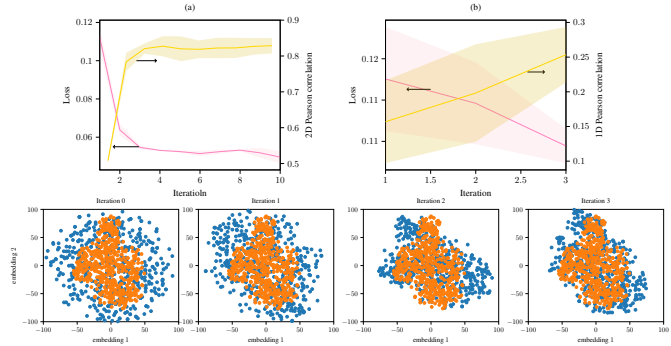

\paragraph{Vision neuroscience}
In the second example, we apply the algorithm to a dataset comprising sequences of natural images $x_{ij}^* \in \mathbb{R}^{n\times t}$ and their corresponding count data $y_{ij}^* \in \mathbb{N}^{m\times t}$. These images and the count data are, respectively, the stimuli entering the retina in Salamander and the elicited time-series count responses of a number of Retina Ganglion Cells (RGCs). Approximating the system as a Poisson process, the system $f$ is defined as the function that takes the image sequences as input and gives a time-varying posterior mean as output. Models based on Convolutional Neural Networks (CNNs) have been recently proposed for this modeling \citet{mcintosh2016deep}. Given $f$, we intend to find a transformed version of the input images that while are constraint to be of lower resolution, still elicit similar neuronal responses (in some metric) to those of the original input images (refer to appendix \ref{appendix:C}). This constraint is imposed implicitly by architecture of the Actor network explained in more details in the appendix. The loss function for this optimization problem reads as follows:

\begin{equation}
\label{eq:e4}
\mathcal{L}  = \min_{\mathbf{\xi,\zeta}} \, \operatorname{\mathbb{E}}_{x_{ij}^*,y_{ij}^*,z}\Bigl[\log \textnormal{Po}(y_{ij}^*| M_{\zeta} (x_{ij}^*,z))\Bigr]+\operatorname{\mathbb{E}}_{x_{ij}^*,y_{ij}^*,z}\Bigl[{\sigma [M_{\zeta} (A_{\xi}(x_{ij}^*),z),y_{ij}^*]}\Bigr]
\end{equation}

where we choose Poisson loss both for the forward and backward mappings. Details of the true system $f$ is given in the Appendix \ref{appendix:B}. Fig. \ref{fig:Figure2} plots the performance metric evolution of this task. It can be seen that the algorithm almost reaches the maximum possible performance of the system (2D correlation $\sim 0.3$) within three iterations. The latent vector of the forward VAE Model of our algorithm is sampled at each iteration and projected to a 2-dimensional (2D) embedding using t-SNE. The true latent vector distribution required for obtaining the desired outputs is also shown. The network architecture and optimization scheme is further explained in the Appendices.

\section{Discussion and conclusion}
We proposed a framework based on VAEs for system control. We also demonstrated how VAEs could illustratively show iterative convergence of the posterior latent variables to those required for obtaining the target outputs. The relevance of the approach was showcased for two applications. The applicability of the method to problems that are chaotic or rapidly time-varying is interesting and perhaps more challenging due to their difficulty of latent space learning. We note that black-box treatment of the physical system by the algorithm should be treated with caution and further studied in future work.




\medskip

\small

\bibliography{ref.bib}
\bibliographystyle{abbrvnat}

\begin{appendices}

\section{Optical control task true system}
\label{appendix:A}
\begin{figure*}[!h]
 \centering
  \includegraphics[scale=.5]{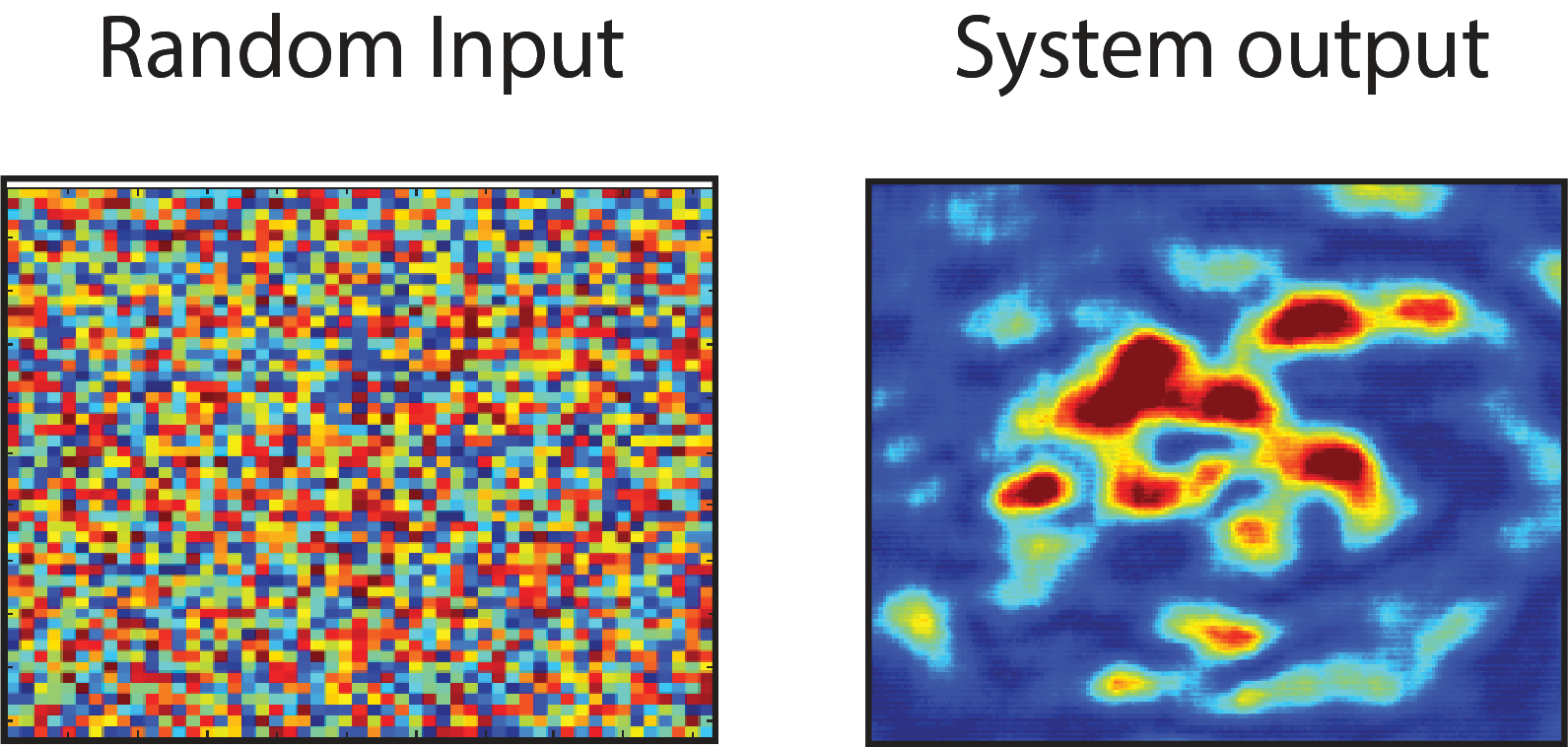}
  \caption{An example of a random input to the system and its corresponding output.}
  \label{fig:Figure3}
\end{figure*}

\begin{figure*}[!h]
 \centering
  \includegraphics[scale=.42]{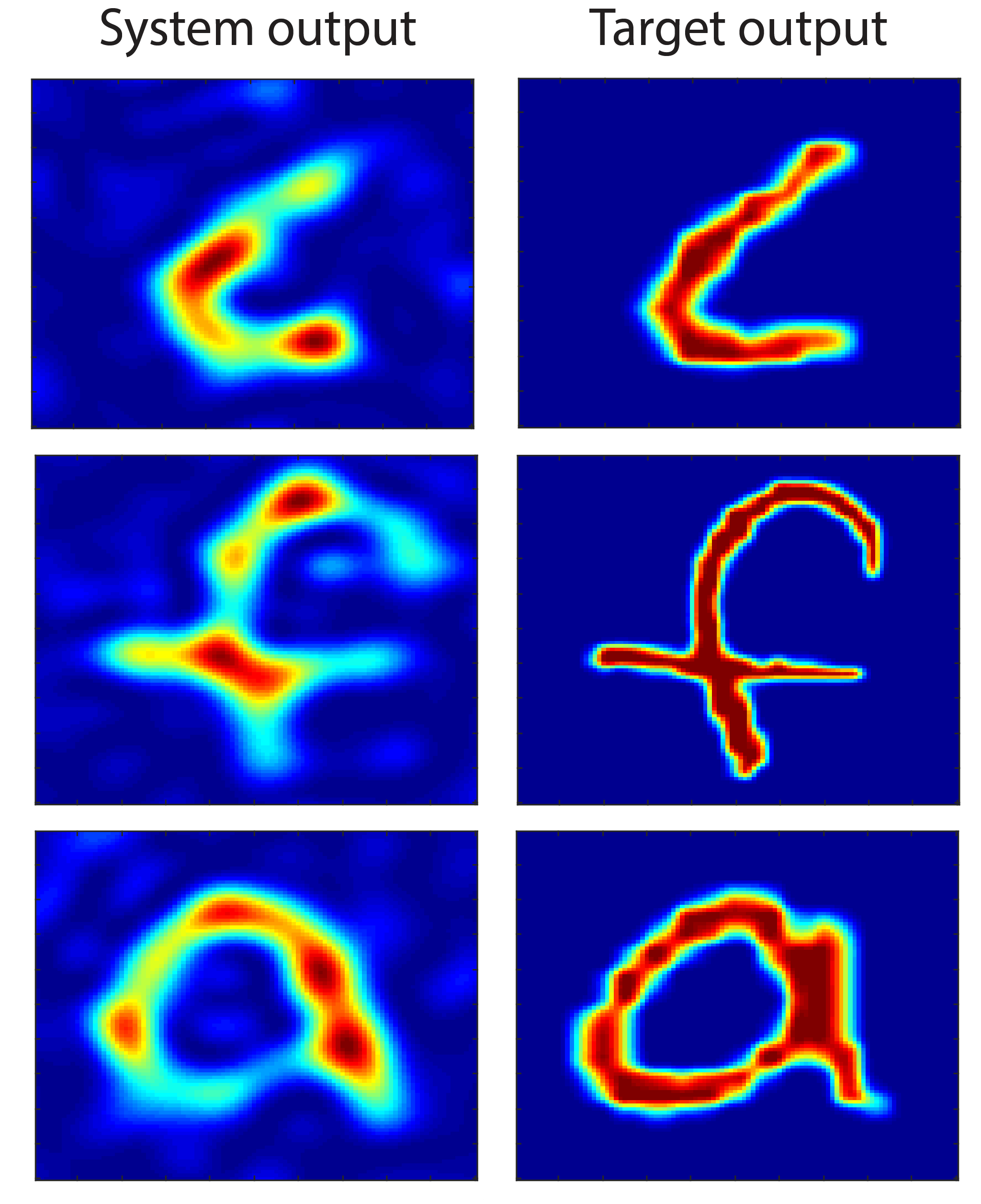}
  \caption{Examples of the experimental system's outputs obtained with the input solutions found by the algorithm.}
  \label{fig:Figure4}
\end{figure*}

The optical control task system is the experimental setup that consists of an input modulator (spatial light modulator), turbid medium (a 50 \textnormal{$\mu$}m core size step-index multimode fiber of length 75 cm) and a receiver (a CMOS camera) working at light wavelength 532 nm. An example of a random input and its corresponding system's output is depicted in Fig. \ref{fig:Figure3}. Some examples of system's output obtained with the input found with the proposed algorithm is also depicted in Fig. \ref{fig:Figure4}.

\section{Vision neuroscience task true system.}
\label{appendix:B}
\begin{figure*}[!h]
 \centering
  \includegraphics[scale=.5]{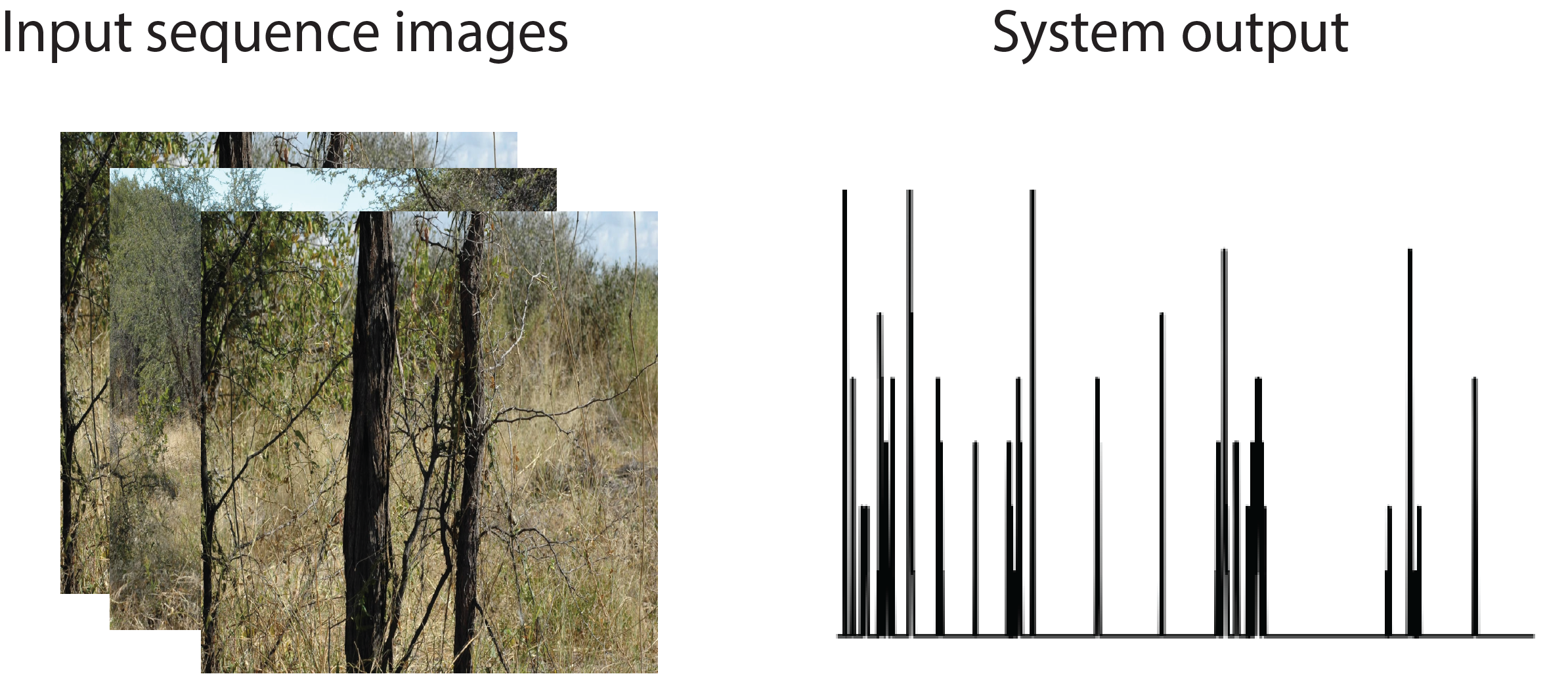}
  \caption{An example of the input image sequence and its corresponding output spike train of the vision system.}
  \label{fig:Figure5}
\end{figure*}

Instead of the experimental system, We used a CNN-based network trained with the \textit{entire} dataset of the input image stimuli and their corresponding neuronal responses as the proxy for the true system $f$. Therefore, to be fair, only a third of the same dataset, randomly selected, is made available to our training algorithm. The architecture of $f$ is identical to that of the forward model. An example of the input-output of this system is depicted in Fig. \ref{fig:Figure5}.

\section{Low-resolution constraint of vision neuroscience backward model}
\label{appendix:C}
\begin{figure*}[h!]
 \centering
  \includegraphics[scale=.75]{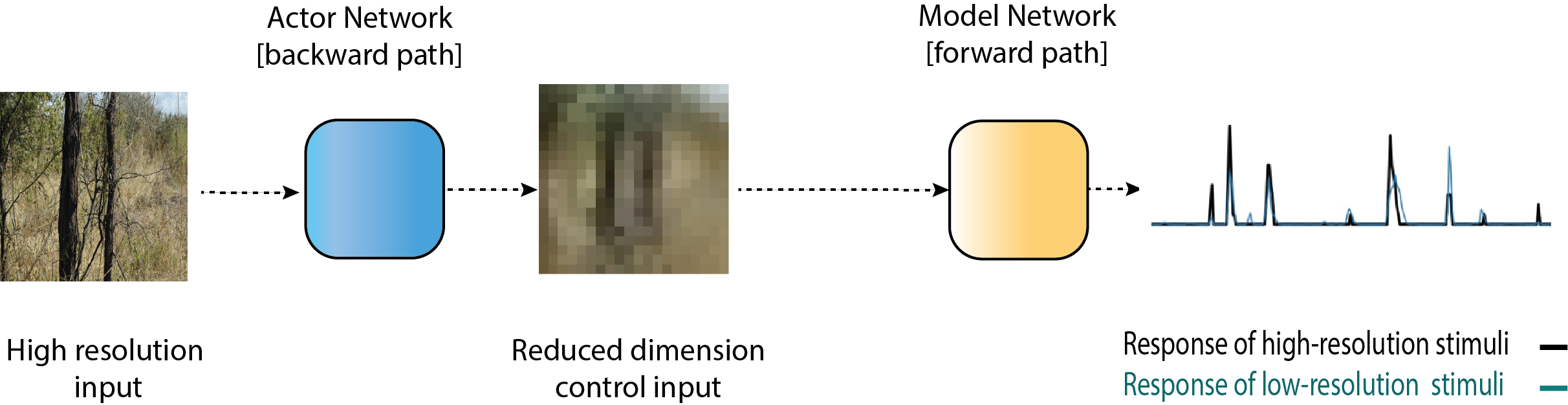}
  \caption{Dimensionality reduction of the input stimuli in the vision neuroscience task.}
  \label{fig:Figure6}
\end{figure*}

The backward model in the vision task, which has a U-net architecture \citet{ronneberger2019u}, is constraint to find solutions that are of lower resolutions than the original high resolution stimuli. This is achieved by adjusting the bottleneck size in the network architecture (denoted in Table \ref{tab:3}). Lower sizes for the bottleneck provide lower resolution solutions. The dimensionality reduction process is depicted in Fig. \ref{fig:Figure6}.

\section{Network architecture and optimization}
\label{appendix:D}
The hyperparameters of the forward and backward networks, optimizers as well as training epochs used for training is summarized in Table \ref{tab:1}. Architecture of the networks is presented in Table \ref{tab:2} and \ref{tab:3}. Hyperparameters were chosen such that a balance between the two terms of losses in Eq.  \ref{eq:e1} is achieved.

\begin{table}[b!]
\begin{center}
 \begin{tabular}{lll}
 \toprule
& Task 1 & Task 2 \\
\midrule
Optimizer & Adam & Adam \\
Learning rate & 1e-4 &1e-4 \\
VAE's $\beta$ & 500/450 & 10   \\
Latent space dim.  & 100 & 15 \\
Train/val/test batch size & 20/-\,/$10^3$ & $10^3$/$10^3$/$10^3$ \\
Train/val/test batch num. & $10^3$/-\,/$10^3$ & 287/72/5  \\
\bottomrule
\end{tabular}
\caption{Training details}
\label{tab:1}
\end{center}
\end{table}

\begin{table}[t!]
 \begin{tabular}{c|c|c}
 \toprule
 
Actor & Encoder & Decoder \\
\midrule
Input $100\times 100$ imgs &
Input $51\times 51$ imgs & 
Input latent dim. vector 
\\

F.C. output $51 \times 51$ Sigmoid &
F.C. output $2 \times$ latent dim. no activ.  &
F.C. output $51 \times 51$ Sigmoid
\\

& 
& 
F.C. output $100 \times 100$ Sigmoid
\\

\bottomrule
\end{tabular}
\caption{Task 1 network architecture}
\label{tab:2}
\end{table}

\begin{table}[h!]
    \begin{adjustwidth}{-.5in}{-.5in} 
\begin{center}
 \begin{tabular}{c|c|c}
 \toprule
 
Actor & Encoder & Decoder \\
\midrule
Input $50\times 50 \times 1000$ seq. of imgs &
Input $50\times 50 \times 1000$ seq. of imgs & 

F.C. output $16\times 12 \times 12$ Relu \\

$3\times 3$ conv. 64 s. 1 same Relu &
$3\times 3$ conv. 64 s. 1 same Relu &
$3\times 3$ conv. 32 s. 1 same Relu \\

$2\times 2$ maxpool & 
$2\times 2$ maxpool& 
$2\times 2$ Upsampling \\

$3\times 3$ conv. 32 s. 1 same Relu & 
$3\times 3$ conv. 32 s. 1 same Relu  &
4 sided zero pad.\\

$2\times 2$ maxpool & 
$2\times 2$ maxpool & 
$3\times 3$ conv. 64 s. 1 same Relu \\

$3\times 3$ conv. 16 s. 1 same Relu &
$3\times 3$ conv. 16 s. 1 same Relu&
$2\times 2$ Upsampling \\

F.C. output Bottleneck(1/4/9) No activ. & 

F.C. output $2 \times$ Latent dim. No activ. &

$3\times 3$ conv. 1 s. 1 same Sigmoid \\


F.C. output $16\times 12 \times 12$ &
&
$21\times 21$ conv. 4 s. 1 no pad. no activ.
\\

$3\times 3$ conv. 32 s. 1 same Relu & 

&
$40 \times 1$ 1D-conv. 4 s. 1 same Relu
\\

$2\times 2$ Upsampling &

&
$15\times 15$ conv. 4 s. 1 no pad. Relu
\\

4 sided zero pad. &

&
F.C. output $9$ Exponential activ.
\\

$3\times 3$ conv. 64 s. 1 same Relu & 

&
\\

$2\times 2$ Upsampling &
 &
 \\

$3\times 3$ conv. 1 s. 1 same Sigmoid & 
 &
\\

\bottomrule
\end{tabular}
\caption{Task 2 network architecture}
\label{tab:3}
\end{center}
\end{adjustwidth}
\end{table}

\section{Code repository}
\label{appendix:E}
All models, implemented in Tensorflow v. 2.1. on Nvidia GPU 2080 Ti, will be available upon publication.

\end{appendices}

\end{document}

%% file: algorithm1.tex
\begin{algorithm}[H]\small
\begin{algorithmic}[1]
    \INPUT Data tuples $(x_i,y_i)$ sampled randomly from partially measured system $y_i=\varphi[f(x_i)]$, target outputs $y_i^*$, $K_1$ and $K_2$ (number of training steps for $M_\zeta$ and $A_\xi$, respectively)
    \OUTPUT The control input $x_i^*$ required for generating $y_i^*$ 
    \newline
    \STATE \textbf{Initialization} Variational parameters $\zeta: \{\theta,\Phi\}$ and $\xi$
    \WHILE{the system's desired performance is not achieved}

      \FOR {i\,$\in \{1,2,3,...K_1\}$}
      \STATE $\zeta \gets \zeta-\alpha\nabla_{\zeta} \mathcal{L}_{M_\zeta}(x_i,y_i,z)$
      \ENDFOR
      
       \FOR {i$\in \{1,2,3,...K_2\}$}
      \STATE $\xi \gets \xi-\alpha\nabla_{\xi} \mathcal{L}_{A_\xi}(x_i^*,y_i^*)$
      \ENDFOR     
      
      \STATE Sample new $(x_i,y_i)$ from $x_i \gets A_\xi(x_i^*,y_i^*)$, and $y_i \gets \hat y_i=f(A_\xi(y_i^*))$
      \STATE Calculate empirical performance metric $\frac {1}{N} \sum_{i=1}^N \sigma(\hat{y_i},y_i^*)$
    \ENDWHILE
\end{algorithmic}
        \caption{\,}
        \label{alg1} 
\end{algorithm}

%% file: pgf/loss-fiber.tex

\definecolor{color0}{rgb}{1,0.752941176470588,0.796078431372549}
\definecolor{color1}{rgb}{1,0.505882352941176,0.752941176470588}

\begin{axis}[name=plot1,
tick align=outside,
tick pos=left,
x grid style={white!69.0196078431373!black},
xmin=1, xmax=10,
xtick style={color=black},
y grid style={white!69.0196078431373!black},
ymin=0.04288261839689, ymax=0.121208410492727,
ytick style={color=black},
yticklabel style={/pgf/number format/fixed},
xlabel={Iteratioln},
ylabel={Loss},
title={(a)}
]
\path [fill=color0, fill opacity=0.2]
(axis cs:1,0.117648147215644)
--(axis cs:1,0.106576416127831)
--(axis cs:2,0.0608079464110327)
--(axis cs:3,0.0536388564483482)
--(axis cs:4,0.052801721313163)
--(axis cs:5,0.0518741541571985)
--(axis cs:6,0.050065014619681)
--(axis cs:7,0.0514373772261845)
--(axis cs:8,0.0527187011579637)
--(axis cs:9,0.0492338428931663)
--(axis cs:10,0.0464428816739735)
--(axis cs:10,0.0527187533664191)
--(axis cs:10,0.0527187533664191)
--(axis cs:9,0.0544050388742609)
--(axis cs:8,0.053707859180997)
--(axis cs:7,0.0534811436312434)
--(axis cs:6,0.0528218208518788)
--(axis cs:5,0.0530503427943816)
--(axis cs:4,0.0534130921528301)
--(axis cs:3,0.0557094333805194)
--(axis cs:2,0.0668622280722111)
--(axis cs:1,0.117648147215644)
--cycle;

\addplot [semithick, color1]
table {%
1 0.112112281671738
2 0.0638350872416219
3 0.0546741449144338
4 0.0531074067329965
5 0.0524622484757901
6 0.0514434177357799
7 0.052459260428714
8 0.0532132801694803
9 0.0518194408837136
10 0.0495808175201963
};
\node[anchor=west] (source) at (axis cs:2.5,.1){};
       \node (destination) at (axis cs:4,.1){};
       \draw[->](source)--(destination);
\node[anchor=west] (source) at (axis cs:2.95,.055){};
       \node (destination) at (axis cs:1.5,.055){};
       \draw[->](source)--(destination);

\end{axis}

%% file: pgf/corr-fiber.tex

\definecolor{color0}{rgb}{0.858823529411765,0.705882352941177,0.0470588235294118}
\definecolor{color1}{rgb}{1,0.843137254901961,0}

\begin{axis}[
tick align=outside,
ytick pos=right,
x grid style={white!69.0196078431373!black},
xtick=\empty,
xmin=0.55, xmax=10.45,
xtick style={draw=none},
y grid style={white!69.0196078431373!black},
ymin=0.482177900618951, ymax=0.9,
ytick style={color=black},
ylabel={2D Pearson correlation},
]
\path [fill=color0, fill opacity=0.2]
(axis cs:1,0.516261494688943)
--(axis cs:1,0.49995194672624)
--(axis cs:2,0.758333892087984)
--(axis cs:3,0.805894120014802)
--(axis cs:4,0.799248235848761)
--(axis cs:5,0.785200279470376)
--(axis cs:6,0.794055585979773)
--(axis cs:7,0.793760650482152)
--(axis cs:8,0.794396558851922)
--(axis cs:9,0.803617074278898)
--(axis cs:10,0.80723608270037)
--(axis cs:10,0.850275803439317)
--(axis cs:10,0.850275803439317)
--(axis cs:9,0.8499972015217)
--(axis cs:8,0.850850522262356)
--(axis cs:7,0.850763092334862)
--(axis cs:6,0.843291776359654)
--(axis cs:5,0.854721608546447)
--(axis cs:4,0.855432868872025)
--(axis cs:3,0.834815811941297)
--(axis cs:2,0.809084788200774)
--(axis cs:1,0.516261494688943)
--cycle;

\addplot [semithick, color1]
table {%
1 0.508106720707591
2 0.783709340144379
3 0.820354965978049
4 0.827340552360393
5 0.819960944008412
6 0.818673681169713
7 0.822261871408507
8 0.822623540557139
9 0.826807137900299
10 0.828755943069843
};
\end{axis}

%% file: pgf/loss-vision.tex
\definecolor{color0}{rgb}{1,0.752941176470588,0.796078431372549}
\definecolor{color1}{rgb}{1,0.505882352941176,0.752941176470588}

\begin{axis}[name=plot2,at={($(plot1.east)+(4cm,0)$)},anchor=west,
tick align=outside,
tick pos=left,
x grid style={white!69.0196078431373!black},
xmin=1, xmax=3,
xtick style={color=black},
y grid style={white!69.0196078431373!black},
ymin=0.101545704660811, ymax=0.119675206943269,
ytick style={color=black},
ylabel={Loss},
xlabel={Iteration},
title={(b)}
]
\path [fill=color0, fill opacity=0.2]
(axis cs:1,0.118851138657703)
--(axis cs:1,0.106253646307812)
--(axis cs:2,0.104648907557412)
--(axis cs:3,0.102369772946377)
--(axis cs:3,0.106518629634838)
--(axis cs:3,0.106518629634838)
--(axis cs:2,0.114564122780875)
--(axis cs:1,0.118851138657703)
--cycle;

\addplot [semithick, color1]
table {%
1 0.112552392482758
2 0.109606515169144
3 0.104444201290607
};

\node[anchor=west] (source) at (axis cs:2.5,.1134){};
       \node (destination) at (axis cs:2.8,.1134){};
       \draw[->](source)--(destination);
\node[anchor=west] (source) at (axis cs:1.5,.1113){};
       \node (destination) at (axis cs:1.2,.1113){};
       \draw[->](source)--(destination);

\end{axis}

%% file: pgf/corr-vision.tex

\definecolor{color0}{rgb}{0.858823529411765,0.705882352941177,0.0470588235294118}
\definecolor{color1}{rgb}{1,0.843137254901961,0}

\begin{scope}[yscale=1,xscale=1]
\begin{axis}[at={($(plot1.east)+(4cm,0)$)},anchor=west,
tick align=outside,
ytick pos=right,
xtick=\empty,
x grid style={white!69.0196078431373!black},
xmin=1, xmax=3,
xtick style={draw=none},
y grid style={white!69.0196078431373!black},
ymin=0.0873679209519637, ymax=0.303337814193494,
ytick style={color=black},
ylabel={1D Pearson correlation},
]
\path [fill=color0, fill opacity=0.2]
(axis cs:1,0.216455815050779)
--(axis cs:1,0.0971847342811241)
--(axis cs:2,0.128043281546461)
--(axis cs:3,0.21344541120308)
--(axis cs:3,0.293521000864334)
--(axis cs:3,0.293521000864334)
--(axis cs:2,0.268522098847998)
--(axis cs:1,0.216455815050779)
--cycle;

\addplot [semithick, color1]
table {%
1 0.156820274665952
2 0.198282690197229
3 0.253483206033707
};
\end{axis}
\end{scope}

%% file: pgf/emb0.tex

\definecolor{color0}{rgb}{0.12156862745098,0.466666666666667,0.705882352941177}
\definecolor{color1}{rgb}{1,0.498039215686275,0.0549019607843137}

\begin{axis}[name=emb0,at={($(plot1.south)+(-4cm,-2cm)$)},anchor=north,
tick align=outside,
tick pos=left,
title={Iteration 0},
x grid style={white!69.0196078431373!black},
xmin=-100, xmax=100,
xtick style={color=black},
y grid style={white!69.0196078431373!black},
ymin=-100, ymax=100,
ytick style={color=black},
ylabel={embedding $2$},
xlabel={embedding $1$},
]
\addplot [draw=color0, fill=color0, mark=*, only marks]
table{%
x  y
21.1500949859619 -22.5011024475098
16.6191120147705 -55.3013038635254
-88.6034240722656 5.73891687393188
-23.0090217590332 -43.2068901062012
49.7083625793457 6.33428144454956
29.0422458648682 82.9691390991211
-68.403678894043 12.7039747238159
50.7699394226074 -8.40394592285156
-1.70406234264374 -25.1554565429688
4.33218574523926 56.5958251953125
8.27231407165527 -44.7257461547852
33.0033149719238 -41.4221305847168
-13.1977405548096 -88.661994934082
-65.8494415283203 33.7700653076172
-56.0236358642578 -51.0484504699707
6.90684700012207 -68.6029357910156
27.5841636657715 -4.36586141586304
56.7395820617676 10.4644155502319
67.6385955810547 64.9821548461914
2.00158286094666 -66.3843307495117
-40.6994972229004 22.1128406524658
-0.328835129737854 -17.8626899719238
-85.1968002319336 38.3350067138672
24.7759780883789 27.6306781768799
81.4989166259766 -3.63760209083557
-14.8966836929321 -36.2158012390137
-43.011890411377 44.7462997436523
39.5894050598145 -30.2418098449707
11.9842929840088 -6.71130037307739
42.2109832763672 57.5758590698242
29.058126449585 -77.1032333374023
46.3608474731445 -22.7671203613281
33.0165252685547 30.4815006256104
-48.6030006408691 43.1410369873047
50.3255081176758 -26.7602806091309
-58.9843368530273 -24.7647857666016
-73.3526916503906 16.4686069488525
15.9004173278809 71.7661437988281
15.1128063201904 67.2534332275391
51.2593650817871 51.3417129516602
-95.5780639648438 14.1938848495483
33.0202560424805 30.4837551116943
42.2243461608887 57.565357208252
-17.1081657409668 -32.7111549377441
61.1512222290039 26.3118515014648
-17.4231872558594 -70.5640411376953
8.18441581726074 83.6281127929688
60.5479888916016 -62.2427635192871
37.6044425964355 27.6864318847656
-33.98876953125 41.3595085144043
-22.8231830596924 -75.4150543212891
-1.43841409683228 -102.151893615723
38.5893440246582 -13.6810426712036
-16.4975280761719 -39.3823165893555
55.5405502319336 44.3064956665039
14.7915229797363 -39.514289855957
-31.5110855102539 77.3052673339844
32.6286773681641 43.6584701538086
-22.7315673828125 76.9272766113281
-23.0078029632568 -43.2052116394043
-62.270809173584 -0.311387032270432
-18.8846759796143 40.8141136169434
17.2401142120361 77.3921203613281
-81.6279983520508 11.2325086593628
37.085132598877 48.717113494873
-56.1809234619141 -20.9248218536377
-55.5926246643066 -34.9346199035645
56.4404716491699 -32.5075149536133
21.0119152069092 33.4822006225586
-38.0850143432617 -19.0480670928955
51.3852806091309 0.226901769638062
51.0247917175293 -46.3693199157715
8.46057605743408 1.75162661075592
12.9210357666016 -31.1383361816406
-64.995475769043 14.2207355499268
-59.5787506103516 0.124428801238537
-64.9930114746094 14.2371444702148
-53.3449363708496 58.9861869812012
1.87235832214355 -98.3153991699219
50.0277900695801 -20.7771892547607
-7.21460151672363 -39.6274719238281
-7.56423854827881 64.4165878295898
-48.4932670593262 -6.96436548233032
-35.1425285339355 -10.0422945022583
-48.977954864502 12.245099067688
-2.81372332572937 53.9565544128418
-22.2474746704102 -1.33123958110809
61.4583587646484 -2.16919350624084
42.8760147094727 64.7778015136719
9.27387046813965 42.1645584106445
27.9242382049561 33.0617980957031
-35.1330375671387 -24.733865737915
-62.7820281982422 -69.4281997680664
-53.2435684204102 68.9651794433594
27.1807498931885 58.3144187927246
76.3477401733398 -25.113037109375
-25.2571792602539 68.186882019043
-49.8262138366699 56.9819717407227
-6.11153602600098 -13.2199335098267
13.3201017379761 -46.0413589477539
-40.5655708312988 62.9804306030273
1.98765587806702 -56.0737648010254
31.7725639343262 62.5756759643555
61.6963958740234 39.9091148376465
26.3997230529785 -56.4978828430176
13.7520980834961 0.165823012590408
7.72943305969238 32.8821640014648
-35.6641654968262 -16.5921192169189
-35.5046844482422 -89.7029266357422
-1.43775761127472 -102.150978088379
-48.7919807434082 9.40722751617432
68.1814575195312 31.5063095092773
34.8923110961914 9.76666641235352
-3.07737588882446 82.6568832397461
41.6059875488281 34.249439239502
35.1201858520508 -21.5875492095947
75.8407592773438 -26.8820209503174
70.6434707641602 -42.2818717956543
4.33107376098633 56.5953712463379
-55.5895309448242 -3.94103407859802
-24.11181640625 -28.3717555999756
-8.0656566619873 -22.3367729187012
66.2538070678711 -13.2430067062378
-80.2746658325195 -24.7477531433105
64.5163803100586 -46.2083053588867
58.3306427001953 13.2043123245239
74.9374313354492 -13.8494729995728
61.1476821899414 26.308895111084
17.9797496795654 55.7092742919922
43.5440635681152 46.0707511901855
-29.9248371124268 42.6758842468262
-3.38869190216064 -27.845911026001
17.7852764129639 3.60319590568542
-19.0168476104736 -58.3244438171387
77.7368850708008 -6.82841634750366
-19.9574928283691 -95.2697677612305
-18.0505390167236 43.314395904541
59.8629608154297 -45.0763320922852
-19.3152618408203 -47.0948066711426
10.5884885787964 -64.8203964233398
-33.4135437011719 -56.3028564453125
68.0221405029297 -29.7244777679443
-46.0941543579102 26.9536514282227
-61.5308685302734 -38.059513092041
-46.2558555603027 -61.1163902282715
-36.6248054504395 71.1921920776367
81.1510162353516 -42.5347671508789
70.3054809570312 46.7288055419922
41.8933525085449 -40.0525131225586
86.0825729370117 29.2621116638184
39.7029037475586 -93.9907455444336
7.06648921966553 -64.0708999633789
-54.7220764160156 31.7875709533691
49.2374305725098 -86.3716354370117
-78.1117095947266 -6.59471940994263
-56.2816963195801 45.5150947570801
-15.8343839645386 -94.829460144043
35.0394401550293 -30.5745697021484
-78.1189041137695 -6.59759664535522
-5.122633934021 18.5098705291748
-44.568302154541 13.7358283996582
3.41126155853271 -25.9032001495361
24.7753429412842 27.6284942626953
37.7390670776367 -92.0137329101562
3.95927357673645 -48.0937423706055
-81.627685546875 11.2329473495483
12.5246105194092 -59.6002426147461
10.5905771255493 38.6500930786133
76.6747589111328 3.97887825965881
-46.6020851135254 86.5274429321289
18.9809284210205 65.7343902587891
-17.395923614502 31.1032619476318
-43.7489013671875 -0.968255281448364
-6.96268701553345 61.7266731262207
-35.9189033508301 62.5714530944824
51.1905136108398 -13.8723917007446
-19.8990631103516 -80.6267242431641
-0.328573256731033 -56.653564453125
24.5155448913574 -95.5054779052734
-11.9723463058472 91.4416580200195
-1.34916996955872 -5.8978123664856
-36.6258201599121 71.1933135986328
33.5148696899414 28.0982685089111
-10.1418142318726 -77.8275375366211
-1.01555216312408 -72.5897369384766
1.97162127494812 67.7962341308594
-88.9382171630859 0.374693095684052
59.9104652404785 -59.0852813720703
2.11882615089417 -52.8746490478516
16.8329982757568 -45.3905372619629
-3.35637187957764 60.7370910644531
7.30264043807983 41.816764831543
-39.9686813354492 4.82787227630615
-22.6513652801514 73.8871536254883
-14.2024993896484 -45.787296295166
23.872486114502 11.27161693573
25.4992771148682 31.6981048583984
-6.96539449691772 -81.2283096313477
20.1691856384277 6.10875940322876
-64.9897384643555 14.224027633667
-35.4553413391113 -36.1114501953125
-42.3222923278809 -50.0125503540039
-32.2418022155762 66.4328994750977
24.4646663665771 -9.0407133102417
-65.941650390625 25.2191638946533
12.9210052490234 -31.1385707855225
-48.7388229370117 -59.2428398132324
-16.0291023254395 2.33989262580872
44.7072105407715 79.053337097168
-78.5981903076172 28.470178604126
0.328640043735504 35.2093772888184
-57.0190238952637 24.5535335540771
-73.6380462646484 4.8135290145874
28.3831005096436 -86.8002700805664
-28.5480461120605 11.3547267913818
14.5298595428467 -68.0594940185547
44.7271461486816 -10.987642288208
-44.5268135070801 59.7973442077637
-28.6346168518066 -59.3394317626953
76.2260284423828 -34.7858543395996
84.3069152832031 1.83728897571564
-24.8009166717529 -62.3949966430664
-58.1382255554199 -13.9216470718384
50.7708930969238 -8.38622188568115
65.8145217895508 23.5081558227539
56.1214027404785 -28.6252517700195
-57.2193984985352 -0.167600199580193
-53.9551658630371 -4.58554553985596
-29.2907390594482 33.3599166870117
5.56126642227173 74.399299621582
-55.9243850708008 55.3096160888672
-3.12474656105042 2.77323460578918
33.6201629638672 -27.7200508117676
-37.4998168945312 -7.85673570632935
-7.65825462341309 8.03091812133789
46.4261474609375 -3.03215146064758
3.77290558815002 80.5998306274414
-31.0028705596924 -73.4079132080078
17.1814212799072 9.04356288909912
-31.5135898590088 77.3065643310547
38.3886756896973 41.4174346923828
45.045955657959 37.9724082946777
-77.0038986206055 -17.4617691040039
1.61067926883698 52.4295043945312
-61.048942565918 33.0834808349609
18.5431728363037 30.3352336883545
29.1286659240723 51.7178955078125
2.00226140022278 -66.382698059082
36.7378387451172 -90.6232070922852
-24.3083686828613 54.155330657959
12.8119125366211 52.2794532775879
-41.3373336791992 30.1891193389893
-9.49973964691162 -97.5412521362305
50.7709808349609 -8.38623332977295
2.48457169532776 8.26243686676025
14.3568086624146 -78.4820175170898
-46.4164237976074 17.0235023498535
-45.1472091674805 5.26138114929199
68.5030517578125 1.16542148590088
6.9039249420166 -68.6065292358398
-19.0854072570801 -24.1010456085205
61.3369102478027 30.1326580047607
67.0801086425781 60.9168434143066
-28.2008590698242 -27.504711151123
-31.7768421173096 -19.3513431549072
5.14915704727173 -15.2436742782593
-11.5458555221558 -91.4554672241211
-28.5643844604492 18.9065742492676
76.7467346191406 -31.8688068389893
7.25611972808838 -82.5008392333984
-77.2702178955078 31.5822525024414
-44.0165252685547 -45.4921875
-7.02245330810547 75.6312637329102
-69.9969329833984 15.1483755111694
61.464973449707 -2.18863224983215
-46.9357643127441 45.4412879943848
5.55597734451294 74.3954315185547
-1.15884959697723 6.32083415985107
-33.1814994812012 75.175407409668
57.3590354919434 -8.942702293396
-77.2702026367188 31.5817222595215
22.183780670166 14.3121995925903
4.5495343208313 -88.5026245117188
-19.7501277923584 24.2495765686035
43.7003898620605 -57.9849166870117
66.4016342163086 -49.4254608154297
25.4506549835205 -12.8427686691284
14.3561172485352 -78.4836883544922
24.515552520752 -95.5054626464844
-11.7274007797241 -18.0168628692627
64.646728515625 -6.27377796173096
5.92423152923584 58.9829063415527
40.1144714355469 54.7469062805176
-74.7816314697266 -13.0865573883057
-36.0863494873047 31.4809589385986
-10.921778678894 44.7296752929688
14.7499589920044 -39.5515327453613
-3.35494899749756 60.7401809692383
12.8101587295532 -94.2077178955078
-37.6587066650391 -14.207423210144
-77.4443054199219 39.645824432373
-10.8620281219482 23.4934711456299
-81.6269226074219 11.2330322265625
70.9015884399414 -52.5406951904297
44.9242935180664 79.2066421508789
-9.4996976852417 -97.5412445068359
21.0450897216797 -47.9238586425781
-3.07760119438171 82.6570510864258
-59.0036239624023 53.1763458251953
-70.9782485961914 -15.3763399124146
-52.8890419006348 22.6675453186035
64.683479309082 -40.31640625
-46.6025543212891 86.5268630981445
56.6806640625 76.6010513305664
-61.5318641662598 -38.0586013793945
-56.7103881835938 -59.6947708129883
-31.8454189300537 -15.4859380722046
-66.356559753418 -65.4269638061523
68.1808700561523 31.5060043334961
-15.8240528106689 -87.1667861938477
59.1292686462402 -81.2135467529297
-34.9618530273438 -18.6811847686768
26.6791954040527 -30.4465599060059
45.3058929443359 -20.5306377410889
41.8945236206055 85.2559967041016
-9.0821475982666 57.6790504455566
23.7090911865234 -82.5844573974609
18.8728370666504 -51.7649459838867
11.279333114624 57.8180274963379
-36.5310554504395 37.8952713012695
-19.7518653869629 24.2511100769043
65.8137435913086 23.5082340240479
54.277702331543 -73.4088287353516
-42.3213005065918 -50.0153007507324
-5.72124433517456 41.0401039123535
42.8756637573242 64.7792205810547
-14.7689046859741 -70.0563888549805
11.4052953720093 29.9739398956299
-17.1091270446777 -32.7105522155762
-0.0240040868520737 26.7842388153076
-34.9644927978516 -18.6806259155273
-23.8671588897705 62.361572265625
-33.1852416992188 75.1765975952148
-69.7765121459961 25.3742980957031
-43.9346046447754 -28.4523315429688
24.7822170257568 -18.8895950317383
39.8574562072754 -34.9591026306152
42.687686920166 51.3472442626953
-3.05136871337891 11.6232948303223
58.2389183044434 -20.0805053710938
26.6838283538818 -30.4528160095215
-7.429527759552 30.1163501739502
4.94722890853882 -1.93569588661194
-44.2075958251953 -24.1683406829834
76.3632736206055 -25.1178379058838
15.2915735244751 -27.2979488372803
-40.6991844177246 22.1127490997314
-85.3312911987305 7.55935192108154
-4.27953290939331 -72.3161468505859
20.4649982452393 -41.889705657959
1.98746144771576 -56.0769577026367
-57.5380935668945 -18.4491367340088
18.8728866577148 -51.765209197998
-17.5985908508301 63.2956352233887
2.48445177078247 8.26147079467773
-57.2183227539062 -0.164321601390839
-53.2437477111816 68.9668350219727
-25.1715049743652 -70.6716918945312
17.1783123016357 9.04111099243164
52.7357368469238 37.7216262817383
-31.9160900115967 45.6129684448242
-20.5534210205078 -33.584171295166
-10.1058340072632 -16.7077541351318
-46.6020126342773 86.527587890625
15.0622053146362 -51.5480041503906
-17.7981605529785 17.2342395782471
18.5460338592529 -65.0013046264648
23.1978912353516 -73.3820037841797
-15.68226146698 -12.8858528137207
19.3751335144043 -11.3641777038574
-53.6722640991211 -41.9655532836914
77.7352294921875 -6.8282322883606
-2.81228852272034 53.9606285095215
-44.205451965332 54.7616653442383
39.5912590026855 -30.2410449981689
-3.01974368095398 -81.6346054077148
10.6247453689575 76.3539276123047
16.9946327209473 13.228928565979
-20.2585334777832 33.3196792602539
20.8575305938721 -84.5068054199219
35.5631141662598 -76.9230575561523
-56.4409370422363 14.3861808776855
-85.55419921875 40.3450012207031
13.1213788986206 71.5030975341797
-1.30332326889038 -66.4766845703125
-37.4917449951172 -7.8549542427063
-54.4865837097168 -49.9855804443359
9.01939487457275 66.0593032836914
10.656436920166 -44.9478340148926
-7.4106183052063 50.626636505127
26.7445945739746 81.4882354736328
-32.0852165222168 -1.5681357383728
-62.6843338012695 25.3254890441895
13.7537403106689 0.166785776615143
-4.28262996673584 -72.3417129516602
-38.1854133605957 15.5000238418579
46.2158737182617 -55.8326721191406
-78.1125869750977 -6.59434604644775
15.1096591949463 18.4111366271973
-73.9196929931641 62.0004692077637
-29.2907238006592 33.3625068664551
-49.5818862915039 -53.3055648803711
51.2402153015137 -40.3995399475098
8.4651050567627 1.75011050701141
90.90283203125 -18.4318161010742
51.385498046875 0.221478581428528
48.5365180969238 43.0263290405273
90.5971145629883 16.8000106811523
-23.0086555480957 -43.2064094543457
5.77266693115234 -85.3339233398438
67.8432693481445 3.7289342880249
-69.4964065551758 -65.5995941162109
35.3266944885254 -7.25010871887207
-48.6028594970703 43.1411781311035
0.126820355653763 70.2942886352539
-14.4762945175171 69.6383666992188
51.2401809692383 -40.4000282287598
-59.0027847290039 53.1747550964355
-26.61203956604 45.3414497375488
51.5339202880859 -35.5219116210938
15.4266376495361 73.9575347900391
22.2371921539307 -3.67488169670105
-24.3795547485352 -55.9067802429199
-41.3351860046387 30.1870861053467
-43.1012306213379 35.5629501342773
10.1139802932739 -99.4064559936523
42.8655433654785 -14.9204244613647
1.97653877735138 67.8043365478516
-49.4140777587891 23.763500213623
74.1035842895508 12.437123298645
-0.326650410890579 -17.8624000549316
16.9469947814941 95.9907150268555
29.0432033538818 82.9695663452148
-61.0691986083984 18.7536602020264
-20.022891998291 -22.1382122039795
-3.14007425308228 -92.3139190673828
41.0333023071289 20.4912815093994
42.8642082214355 -14.9196634292603
-48.4962272644043 -6.96518421173096
16.3245410919189 42.5537300109863
-49.5822067260742 -53.3058128356934
33.0441703796387 -72.2859039306641
10.1298704147339 60.8395462036133
19.2336463928223 -5.23422241210938
61.3369064331055 30.13232421875
-13.1837892532349 -53.0705718994141
-37.4385948181152 -46.253345489502
35.3268699645996 -7.25023555755615
52.044750213623 -72.1877746582031
-65.4698944091797 45.8860092163086
81.1509323120117 -42.534797668457
-6.96240711212158 61.7262001037598
-22.2647018432617 49.9613189697266
-88.6027221679688 5.73215055465698
76.6028823852539 -51.499080657959
58.0229339599609 41.7488708496094
56.9711837768555 -16.9041919708252
37.9919929504395 -11.9333276748657
-69.8509292602539 -35.5823631286621
-6.40222358703613 -35.272331237793
4.94808530807495 -1.93559181690216
28.3867702484131 -86.8064651489258
52.044849395752 -72.1878433227539
59.9058265686035 -32.3837127685547
-16.27663230896 -65.9089660644531
-31.2831974029541 -4.39884090423584
24.4633388519287 -9.04107475280762
20.7877407073975 -59.114818572998
-93.5070114135742 26.076545715332
-62.0948867797852 -26.3198051452637
42.6879005432129 51.3471031188965
43.7020149230957 -57.9846649169922
-13.2003173828125 103.126762390137
7.25477981567383 -82.5028533935547
-33.0571823120117 89.2600860595703
11.4057550430298 29.9739665985107
-30.9002132415771 4.65049409866333
60.1746673583984 -21.8173160552979
67.0805358886719 60.9165191650391
-1.43796348571777 -102.151206970215
11.898060798645 -11.1211729049683
19.4402542114258 -31.3543472290039
-57.0183486938477 24.5517673492432
-3.05132460594177 11.6232957839966
-59.6362571716309 -52.1943283081055
11.5595426559448 85.1811218261719
12.81032371521 -94.2078247070312
-1.03176307678223 -72.5977783203125
-14.5826587677002 -23.4674587249756
-38.3712005615234 -4.90947103500366
};
\addplot [draw=color1, fill=color1, mark=*, only marks]
table{%
x  y
-14.1562461853027 16.2737731933594
15.9061107635498 0.851377665996552
-14.5903673171997 32.9087791442871
24.7583923339844 -52.4581413269043
1.76987397670746 -18.1802673339844
12.3498573303223 -70.0374374389648
-10.3571968078613 74.6969299316406
-49.0271110534668 -17.9125480651855
22.1610584259033 -49.2137107849121
36.1605529785156 -17.489049911499
-18.2265892028809 -11.663556098938
-5.55481910705566 73.7786407470703
13.1064291000366 -44.5872497558594
46.4859504699707 -35.5428199768066
14.6336479187012 -65.2039794921875
-10.5388031005859 65.3314514160156
-6.78564691543579 26.3422622680664
0.797948896884918 76.6471176147461
35.6403961181641 -61.0181159973145
28.7877178192139 -49.7694931030273
-12.8304452896118 -14.748987197876
6.28536701202393 -48.0450553894043
4.0225887298584 26.9253406524658
49.203483581543 -13.8960771560669
10.4839687347412 79.8198547363281
1.43978202342987 -64.7868728637695
9.88216018676758 44.1859855651855
-54.984489440918 -9.38840389251709
-41.3730010986328 -12.4610357284546
-23.3574848175049 18.0422458648682
33.5815658569336 16.6903343200684
45.8693923950195 -34.5644378662109
-3.67634797096252 73.5894012451172
29.9919891357422 -14.1625623703003
18.2956447601318 -43.659553527832
-23.6551246643066 17.2828235626221
-0.523545205593109 -33.7187957763672
-3.46174550056458 25.0130176544189
-13.303617477417 61.720516204834
8.45828437805176 57.3310928344727
25.74001121521 -3.36165690422058
30.3130989074707 -61.7557830810547
-9.81210994720459 -66.5528717041016
-7.22028207778931 41.8842697143555
27.6234664916992 -26.0225315093994
-23.6735897064209 23.3374004364014
25.8328628540039 -44.4100875854492
-47.7951469421387 -8.54372787475586
-9.06405830383301 -16.9985446929932
3.21282887458801 -58.4978675842285
-43.9842987060547 -6.30048370361328
2.99738717079163 23.0593242645264
15.7514410018921 10.014009475708
-42.0615005493164 15.3201522827148
-0.909237623214722 -9.83153057098389
-10.9196891784668 -22.6806621551514
-28.9560070037842 -38.3873100280762
-21.7516994476318 37.6689529418945
-0.90399307012558 -12.597393989563
-10.7759971618652 79.7270126342773
6.95509386062622 -31.4997692108154
-1.24866032600403 -48.9488410949707
1.08647096157074 -73.665901184082
31.7287845611572 -2.52046322822571
31.3026351928711 -4.13925886154175
-9.25134658813477 -46.2501754760742
38.2947578430176 -5.81643438339233
0.778681576251984 -26.6699714660645
13.3758659362793 -49.5960464477539
-2.4627513885498 -37.982551574707
18.2552795410156 -45.910270690918
36.0335083007812 -43.0681915283203
-45.2637023925781 0.117252051830292
-20.7245254516602 27.6910781860352
4.71429300308228 43.474235534668
-8.54316520690918 29.0524082183838
-10.6342191696167 62.4147338867188
-25.5382080078125 2.76588034629822
-20.9436187744141 -10.1609754562378
33.8535652160645 -8.42477703094482
-41.7106094360352 17.5954570770264
-42.6732711791992 30.4063396453857
6.28234338760376 7.22087717056274
-4.92992496490479 -56.8990135192871
-11.8856220245361 38.4550132751465
8.65184688568115 0.349969923496246
30.8050785064697 -12.5851440429688
-8.84524250030518 21.4997024536133
8.73474788665771 78.5139312744141
13.9132356643677 -56.6436729431152
15.94407081604 19.3828907012939
19.3564701080322 -4.94871473312378
-48.4773941040039 -23.1676368713379
-11.1856412887573 7.2453989982605
-1.52089750766754 -30.7742691040039
1.84971594810486 37.3751029968262
-5.7638635635376 8.2220287322998
10.7414426803589 1.55196952819824
-15.4240303039551 -29.0445556640625
22.3318481445312 27.5244808197021
37.5331916809082 -62.9953689575195
27.2599811553955 -48.8286933898926
-9.75624656677246 65.166145324707
-0.649470210075378 13.6928176879883
-28.4043292999268 20.3937892913818
39.9670448303223 27.7662830352783
19.4506244659424 -2.32283234596252
-16.0902843475342 -60.8875999450684
8.23221015930176 55.2768592834473
7.78045177459717 63.0177726745605
33.5156669616699 -37.2890853881836
24.4366474151611 13.5936975479126
30.8764934539795 -20.917142868042
45.1145668029785 20.4626750946045
25.9215526580811 -48.9129219055176
11.8244342803955 -50.2865715026855
-17.0146446228027 -19.9688949584961
11.3605842590332 -39.7386283874512
-31.621208190918 4.6064395904541
2.80386805534363 -40.207405090332
-8.1839075088501 30.9129390716553
13.5803813934326 -66.5836944580078
-17.5457954406738 24.6219291687012
-30.2811622619629 -15.9232082366943
-10.8703584671021 -2.81906247138977
-2.95088005065918 -38.0437088012695
21.6684036254883 -66.0403289794922
-16.8177680969238 -2.28541779518127
11.2980442047119 -4.48066854476929
16.5028839111328 -10.287504196167
-32.2802047729492 -13.6332426071167
-47.3634262084961 -11.6237506866455
17.3558483123779 -76.4602737426758
4.3209376335144 -12.1043319702148
-21.5901622772217 3.09332180023193
21.5516510009766 25.0099468231201
-24.203426361084 25.5471076965332
-8.62772560119629 67.5758895874023
8.06674289703369 33.1486778259277
-4.65688323974609 -1.56843042373657
6.29899215698242 67.7486572265625
-10.5629920959473 -22.7748241424561
26.409646987915 -51.4696846008301
12.7960977554321 76.7983474731445
-39.6081924438477 -1.63019394874573
2.3358690738678 -28.6222839355469
44.9914283752441 -53.0035552978516
-47.1891784667969 -3.51396441459656
-17.2614307403564 14.4307317733765
27.7461719512939 -12.9975690841675
-14.8937530517578 70.0156936645508
-19.1048164367676 15.080099105835
29.5145778656006 -40.4345779418945
-11.9497404098511 66.8091354370117
31.4417381286621 -9.81137275695801
40.2962608337402 11.2909832000732
-4.69818592071533 -16.7863788604736
25.4435920715332 -46.8182754516602
-13.724157333374 -19.9166126251221
4.88891696929932 24.9782314300537
17.5421257019043 25.3467330932617
-1.54870390892029 27.5834064483643
-35.4704093933105 -5.87784862518311
-12.174446105957 11.2807664871216
-17.2362785339355 55.619499206543
-22.6500720977783 -28.7325820922852
-38.5535430908203 12.7513599395752
15.0858554840088 -60.9970588684082
40.2745246887207 10.7555618286133
2.10408425331116 -43.9000816345215
2.43483734130859 15.0179214477539
3.69317245483398 -32.3478851318359
-12.121976852417 82.5328826904297
29.1320610046387 1.27591419219971
32.0222129821777 25.2803649902344
-46.8451690673828 4.17712068557739
36.4329147338867 -49.6340103149414
-12.5656356811523 -47.868968963623
2.3995053768158 3.609938621521
-14.246506690979 -22.5724544525146
25.0124397277832 -25.8872509002686
-0.254881918430328 31.9352989196777
-12.4840927124023 23.8146667480469
-31.6277961730957 -15.385142326355
-35.646427154541 -21.0412502288818
9.02097988128662 61.0992965698242
23.1384620666504 -61.568775177002
5.37125873565674 3.65267586708069
1.76970970630646 -8.3620548248291
-11.1585922241211 80.6489181518555
-34.6866912841797 -28.5431785583496
-2.43847513198853 -54.868049621582
-2.0417959690094 -1.06925773620605
40.1018676757812 4.54694747924805
31.3149070739746 -5.08900690078735
2.91170954704285 46.0739555358887
-16.4742374420166 19.0722599029541
4.82551193237305 -25.9199619293213
-48.4812698364258 1.47474801540375
-22.6564407348633 -5.44893312454224
-39.1985664367676 -8.78360557556152
-41.8020248413086 15.4914474487305
38.3638916015625 -26.1201820373535
-16.8518733978271 77.0881042480469
-35.7138023376465 16.9489593505859
-7.21964645385742 -45.8839645385742
34.8170280456543 -12.6864891052246
26.6888599395752 -59.169303894043
-19.6623344421387 19.8142642974854
-7.39723062515259 13.1846542358398
45.6438179016113 -4.15494441986084
0.111052051186562 -33.7903213500977
-3.46785235404968 -9.97296714782715
26.0671405792236 0.227064251899719
26.5278739929199 4.71392631530762
-7.33442544937134 81.3036270141602
-35.3015899658203 31.5906181335449
1.35896062850952 58.5727119445801
1.5297212600708 71.1305923461914
-10.010570526123 23.5384063720703
18.9875087738037 -6.82228851318359
6.80627346038818 23.0084571838379
4.11152935028076 -34.8415184020996
-12.7648687362671 -47.0462188720703
15.9349784851074 -21.9359931945801
28.6154327392578 3.10245203971863
-28.0933208465576 1.6114673614502
9.18317985534668 47.0225067138672
-12.45578956604 68.5944442749023
21.3322200775146 -70.7522659301758
30.226448059082 2.85382008552551
-0.521462440490723 -47.6239852905273
37.0535774230957 23.7878074645996
-1.67553639411926 61.4517555236816
34.4644203186035 -12.1803646087646
28.5689296722412 -22.4197235107422
-32.4310493469238 -18.4039688110352
39.0668754577637 -12.8985385894775
45.6493759155273 4.99477672576904
11.5254974365234 41.074592590332
-8.21777820587158 38.5960464477539
21.1782970428467 7.94684505462646
-15.5823822021484 20.4763259887695
-7.84456348419189 33.6721115112305
14.9224367141724 -44.2074966430664
14.3978500366211 -32.9720993041992
-43.2316627502441 -20.4248332977295
-29.3081855773926 20.3523845672607
-20.6603851318359 -3.9294376373291
-1.38323223590851 47.1030387878418
-27.3588199615479 39.5297660827637
24.0691299438477 -50.2185173034668
-45.5678291320801 -25.494592666626
-8.38327789306641 -2.32737755775452
0.616246342658997 -39.3792991638184
-6.9112434387207 35.1056709289551
36.5442810058594 18.2156276702881
-6.54022693634033 22.064790725708
-42.7417068481445 27.1025981903076
-2.0668785572052 -15.1366767883301
-32.1374626159668 -23.2694110870361
-51.4727516174316 -14.646125793457
9.67971038818359 10.4945592880249
34.4755973815918 12.4918699264526
-50.3344345092773 -4.68861103057861
19.0577411651611 13.9575262069702
11.7904024124146 -37.3484649658203
10.4059038162231 -48.5614318847656
-6.00288057327271 -8.90501689910889
34.664249420166 12.0318851470947
17.9467601776123 -66.4420394897461
21.5221843719482 -59.7803802490234
11.1702671051025 74.4532623291016
15.7115955352783 -64.0511245727539
6.62478828430176 38.1891136169434
-39.778018951416 -25.5806522369385
-39.5218887329102 14.0598936080933
1.42190909385681 -53.7553443908691
-10.9876232147217 69.3366317749023
-7.99409246444702 20.985725402832
-33.2599143981934 27.005298614502
-21.3398609161377 23.3442535400391
18.7808227539062 -75.8148574829102
-9.13871479034424 49.2513580322266
-22.5335121154785 -29.2492485046387
11.6988506317139 75.239616394043
-6.66132497787476 69.9680023193359
20.1837615966797 -72.3973007202148
4.53920030593872 63.4861526489258
27.4785194396973 -56.5249290466309
31.1390323638916 7.47712659835815
-2.16996502876282 -58.0512161254883
32.9060592651367 -49.1143951416016
-13.7541513442993 -4.27445602416992
24.6208076477051 4.66902160644531
-20.3429908752441 3.30881142616272
-29.2848739624023 -26.0461730957031
26.9002685546875 -46.6657180786133
-17.4212207794189 34.3454704284668
0.481921970844269 5.13405418395996
-6.70317602157593 16.1393127441406
-14.9225292205811 -52.051643371582
-12.7421054840088 5.92016887664795
-44.0190124511719 -1.4125919342041
5.69720506668091 56.2550811767578
-5.33053112030029 -28.6760196685791
-43.5013122558594 1.70440280437469
-16.7730884552002 -50.5043869018555
-54.7032814025879 -14.4245710372925
-16.3418769836426 25.6480083465576
-53.2781944274902 -15.987491607666
23.5342407226562 -35.0567817687988
6.20785188674927 64.8274536132812
-51.5338935852051 -10.556471824646
5.68330335617065 19.8132553100586
15.3247947692871 -3.25546836853027
-26.7605438232422 12.6324501037598
-17.1904792785645 -1.1943895816803
-5.54897451400757 63.4965934753418
35.7032318115234 12.815821647644
5.95640516281128 -67.5386276245117
-17.6672134399414 44.9177665710449
4.53430223464966 -28.5955810546875
14.6802082061768 -20.8955821990967
-14.3905334472656 11.6104049682617
9.05849838256836 74.4791107177734
-36.0064239501953 1.14881777763367
37.4585456848145 -18.2941722869873
-14.5936479568481 61.6761436462402
-20.5579738616943 -20.6917114257812
17.1725883483887 -6.50464487075806
-6.07322120666504 -24.980224609375
-3.1837592124939 13.4426889419556
6.84798860549927 -43.4136352539062
0.68505322933197 44.5727310180664
17.0947704315186 -64.4154815673828
-2.5180516242981 -64.8597412109375
8.40808582305908 -50.0519676208496
-53.5180053710938 -16.5135402679443
-9.29903316497803 51.9289207458496
23.4430599212646 -1.8377548456192
29.9761981964111 -53.7202911376953
-6.89573097229004 -49.0051002502441
17.1100425720215 -58.1905899047852
11.3685636520386 -11.402551651001
-3.10602593421936 65.9060668945312
-39.5226516723633 13.0065221786499
-1.43580591678619 69.1976776123047
-4.13788604736328 30.8474102020264
15.560097694397 -49.7316246032715
-39.9275054931641 11.8809366226196
2.13128209114075 64.247444152832
-12.0617790222168 48.4711875915527
4.42822170257568 -52.078182220459
35.5346374511719 -20.2437839508057
2.57543063163757 59.2436637878418
-39.2273483276367 13.9640045166016
-39.0242462158203 22.9863262176514
-12.306568145752 3.77273058891296
-30.7462024688721 1.61940360069275
35.7655181884766 -44.487865447998
1.52739584445953 66.1834182739258
10.0763473510742 -49.0631675720215
0.760641396045685 -60.115837097168
35.1040725708008 -53.1094245910645
4.08598947525024 84.7822799682617
36.7877464294434 -42.2259330749512
-9.05131912231445 -54.4618797302246
11.1937694549561 24.7227592468262
28.5881576538086 -40.3376121520996
19.4687156677246 -25.4935531616211
-31.2322673797607 25.4531726837158
-38.107479095459 -9.05575561523438
40.2524681091309 -55.6437835693359
-38.8869400024414 -24.7076225280762
-51.5100479125977 -23.7850437164307
14.2239561080933 -1.5298445224762
3.43313956260681 -26.457239151001
-14.7513008117676 59.8690528869629
3.2206859588623 11.7626342773438
-11.1817827224731 7.24995851516724
-13.5163869857788 34.2488327026367
-3.65907788276672 -27.7040939331055
-13.2587223052979 40.5102844238281
-25.1148815155029 -1.82188069820404
42.614875793457 -55.443660736084
-26.7906265258789 24.2013778686523
27.493932723999 -41.520191192627
0.206918761134148 -42.6391639709473
-36.6595687866211 -24.2476444244385
-3.31427526473999 -6.8505425453186
10.03773021698 -37.2256546020508
22.3766479492188 -10.2656364440918
-2.72840595245361 46.2721061706543
26.7294540405273 -30.5466651916504
38.8881492614746 -44.215404510498
14.4415721893311 -63.5295181274414
-37.0411911010742 20.3775463104248
15.9359560012817 -32.611888885498
-1.22140038013458 8.00126934051514
2.23746418952942 54.3661193847656
-32.4431686401367 5.31508731842041
-6.95743560791016 -32.4719734191895
-3.33114576339722 41.5269088745117
5.78294277191162 -27.7950344085693
-15.9254083633423 -12.0912179946899
-1.86363780498505 32.3100433349609
-18.5041275024414 -34.950080871582
-19.3557910919189 58.7318649291992
-12.6471309661865 -28.8995933532715
-29.3336219787598 -6.75573492050171
-12.5778541564941 -29.3947620391846
-0.755422353744507 21.0363597869873
-53.4739990234375 -11.9781694412231
9.53775215148926 1.18791878223419
-5.44722270965576 3.71906733512878
9.39107418060303 -62.8244514465332
25.2141914367676 -40.1923370361328
-23.5486850738525 -55.1394195556641
36.624927520752 -51.204517364502
-1.41397559642792 64.4651794433594
31.1220169067383 32.1042709350586
28.4181709289551 -61.9741172790527
41.0682106018066 -6.71240520477295
7.91067838668823 39.9108734130859
-1.29828250408173 -3.32939910888672
-9.48875617980957 72.2016906738281
-10.4276132583618 38.6326713562012
-49.5756340026855 -17.0155220031738
-1.23361301422119 85.5276947021484
-9.40144634246826 62.9727249145508
1.43499612808228 59.0431213378906
-6.21810626983643 53.1744232177734
-40.3940620422363 29.7566719055176
25.2305107116699 -40.0765686035156
14.0889110565186 -52.4261283874512
-4.53575658798218 25.081392288208
-13.8788919448853 23.9843482971191
-36.7962684631348 9.24637508392334
15.9282217025757 31.1022338867188
11.8129873275757 73.0501480102539
14.2319955825806 -63.4841346740723
-1.97849798202515 66.7942733764648
-23.7449798583984 -56.3668365478516
-48.5560188293457 7.43255281448364
-20.9436798095703 -41.8449440002441
-14.2904224395752 23.3435115814209
10.4972915649414 49.6534957885742
-3.99147343635559 -50.0040283203125
18.211742401123 -51.417423248291
-5.88881301879883 -8.80561828613281
-1.51741051673889 -47.4201202392578
4.32817792892456 31.6414756774902
-24.871789932251 -7.62801742553711
-20.4118881225586 45.7798843383789
27.4483509063721 -26.8081016540527
-5.56537246704102 -29.4588851928711
-16.0065059661865 -65.9480972290039
-5.21497488021851 -25.0645942687988
40.4045066833496 -20.6414604187012
-45.9770278930664 22.8458423614502
-26.5258178710938 -0.407515108585358
28.5873775482178 -46.7295265197754
-44.2382164001465 10.3070650100708
-50.5521125793457 5.60726404190063
-7.19392871856689 -42.993709564209
-9.5423583984375 -38.5154151916504
-43.3997917175293 -7.75084924697876
9.01693916320801 -51.2871055603027
25.0602397918701 13.2396965026855
-29.0721530914307 11.8301572799683
-27.9322357177734 -3.03503251075745
1.96551692485809 36.5476722717285
-7.27517414093018 82.8970489501953
0.202581092715263 59.3200950622559
33.6214828491211 4.04944849014282
34.3057708740234 -22.4389762878418
-23.6837158203125 -9.76546096801758
-34.6332054138184 0.0546604692935944
39.3979530334473 -44.4877548217773
-22.6435832977295 7.02360010147095
29.4733123779297 -25.9379234313965
-33.2400817871094 5.08824300765991
-38.6765251159668 4.40700483322144
9.50541591644287 18.0882816314697
10.8913688659668 -52.0722389221191
32.0797653198242 -33.75390625
-6.8299732208252 1.5664496421814
27.8091373443604 -42.6121864318848
7.99870491027832 47.2746238708496
-47.7646560668945 19.1078987121582
-0.175539493560791 -6.24241399765015
-15.9347543716431 -23.7865829467773
3.07636833190918 -4.72971487045288
37.8014335632324 -23.3688488006592
18.0060062408447 42.4501457214355
13.2788887023926 11.630054473877
-7.13648319244385 -29.9890613555908
0.0621891058981419 87.6787338256836
7.98978853225708 -35.625072479248
};
\end{axis}

%% file: pgf/emb1.tex

\definecolor{color0}{rgb}{0.12156862745098,0.466666666666667,0.705882352941177}
\definecolor{color1}{rgb}{1,0.498039215686275,0.0549019607843137}

\begin{axis}[name=emb1, at={($(emb0.east)+(1cm,0)$)},anchor=west,
tick align=outside,
tick pos=left,
title={Iteration 1},
x grid style={white!69.0196078431373!black},
xmin=-100, xmax=100,
xtick style={color=black},
y grid style={white!69.0196078431373!black},
ymin=-100, ymax=100,
ytick style={color=black},
xlabel={embedding $1$},
]
\addplot [draw=color0, fill=color0, mark=*, only marks]
table{%
x  y
9.32280254364014 -32.946460723877
34.526496887207 -15.1829023361206
31.9747467041016 -3.11646389961243
41.560375213623 53.2561988830566
-4.45113325119019 29.7055625915527
-80.122802734375 -10.0564661026001
51.6337394714355 -20.3851928710938
-25.6542205810547 -42.2791709899902
6.75061893463135 10.8114166259766
-45.2362022399902 -13.0009365081787
10.4404497146606 -10.5440664291382
32.0250434875488 -72.8527374267578
-46.854118347168 80.6315994262695
-49.8105430603027 63.9247207641602
-54.5593757629395 83.9864883422852
16.6458778381348 -85.2203140258789
-10.1818437576294 -1.41159915924072
17.076810836792 -71.412971496582
-2.86121582984924 29.7463092803955
-10.102822303772 81.2242889404297
-4.1091890335083 38.2712287902832
-85.0359268188477 40.3795547485352
76.3347015380859 -19.8100242614746
-33.5404472351074 55.5038833618164
55.3413352966309 -60.556770324707
0.998445570468903 77.1460647583008
18.5814228057861 -66.8374786376953
39.5954170227051 -66.680046081543
42.4923362731934 -18.7368469238281
-6.67772006988525 -45.0622100830078
0.322501450777054 -56.4279937744141
-29.8038997650146 -29.8366966247559
47.4103622436523 -37.8953475952148
-29.8565406799316 51.0941886901855
-52.6826934814453 9.93048000335693
68.1349487304688 -41.9615478515625
61.3903694152832 23.6721248626709
14.0493593215942 -68.0750732421875
52.6380653381348 -75.1964340209961
35.7198028564453 -68.0569152832031
-18.2849864959717 -9.28180694580078
-17.3924655914307 75.4655990600586
-31.0464363098145 -25.6747245788574
-2.93645858764648 -9.14064693450928
18.4116134643555 22.0908966064453
-27.5344047546387 -20.000675201416
-55.4290237426758 74.3099975585938
22.4228591918945 -91.769416809082
-44.7375144958496 25.3065032958984
-59.0680885314941 23.7977828979492
-67.6604690551758 -15.0737895965576
62.597785949707 -0.960226237773895
-43.6413383483887 -28.2213172912598
24.2127437591553 -23.4485492706299
63.818172454834 -16.5176525115967
-49.2001152038574 -23.4509658813477
16.3691730499268 28.8970909118652
-44.4911880493164 14.281943321228
73.2066955566406 18.0678005218506
52.8835372924805 -81.6611633300781
-55.8699722290039 0.525313019752502
-57.8021926879883 33.9127464294434
-8.58901691436768 60.2787704467773
-9.08064556121826 -4.24012517929077
-42.6634635925293 29.0260066986084
18.1355533599854 65.687614440918
-31.076488494873 24.5561504364014
-57.6602897644043 27.4743824005127
-20.1648368835449 47.7934303283691
-60.5689468383789 -25.2738609313965
-30.0559501647949 22.1438770294189
-44.320011138916 44.4069328308105
-7.67110824584961 3.07240700721741
19.5080451965332 -39.8024559020996
54.9878883361816 -36.5339698791504
-64.5148162841797 -37.4931411743164
54.1023483276367 -38.0459175109863
36.5586090087891 -46.9642677307129
-32.8492774963379 -2.34557580947876
33.9863700866699 73.5102310180664
-36.4442901611328 -7.13967895507812
25.8564281463623 -38.6079177856445
9.46086597442627 31.5260009765625
-36.824634552002 29.4553604125977
-72.1629180908203 56.5709495544434
-76.9896774291992 48.4354400634766
-10.019359588623 -9.48846244812012
-29.0381031036377 -40.6472320556641
25.3106575012207 -61.2886238098145
-60.5094413757324 4.34746980667114
30.8753681182861 74.0744705200195
-11.8877573013306 -90.1683349609375
48.1179695129395 8.89127349853516
39.6276893615723 -42.2327461242676
-1.54018652439117 -1.91422176361084
15.1592197418213 -42.6865234375
45.6581840515137 4.34763669967651
67.2036056518555 28.8080348968506
-56.9002151489258 59.9553031921387
33.1799240112305 -11.5996761322021
-39.6856689453125 53.4896087646484
-72.5668411254883 41.6779289245605
56.2286491394043 -82.3176651000977
-46.7246932983398 0.815215528011322
-0.702592790126801 -30.9812774658203
2.73739290237427 -38.4423675537109
-77.0647048950195 35.6938667297363
15.7170715332031 48.4100914001465
40.7115364074707 -47.3840599060059
62.5555267333984 -2.11803984642029
-7.99502277374268 24.4965686798096
-22.1945381164551 -47.2731781005859
-66.998908996582 -38.1692008972168
7.62552738189697 -11.0215129852295
-73.2337341308594 57.6109199523926
-47.3230209350586 74.9012756347656
-63.2450370788574 40.842227935791
-74.1817855834961 -18.0148544311523
27.8757286071777 -58.9619255065918
-49.7687835693359 -4.36497640609741
16.7516899108887 -65.4715042114258
45.9593544006348 74.3016815185547
10.4519863128662 11.6184167861938
5.08792352676392 15.8655118942261
-42.1859817504883 16.8958320617676
-38.5915679931641 43.8222351074219
-15.5753555297852 54.1078834533691
22.2897491455078 -48.4592628479004
-4.66438436508179 12.3368940353394
-58.6211013793945 48.2821960449219
-6.87540864944458 73.8236541748047
-31.7859592437744 -49.7100410461426
-59.8760223388672 67.9472427368164
-77.9946517944336 58.2399368286133
-38.3803176879883 -69.9543685913086
40.5032768249512 25.8013820648193
46.5774192810059 2.71805047988892
17.7148876190186 -76.5003967285156
17.0730323791504 -0.442331612110138
4.61158180236816 -67.901237487793
40.9189796447754 -68.0545501708984
-58.4160919189453 43.9162292480469
11.0030708312988 46.3272895812988
38.7281150817871 -98.0504455566406
2.25405859947205 85.4136810302734
-8.34746932983398 71.824577331543
-23.8171310424805 95.2674407958984
-48.4916839599609 37.1843376159668
47.8225784301758 -8.50154399871826
6.11261987686157 61.6497421264648
37.644603729248 -75.8468475341797
-61.8102493286133 45.9109191894531
-32.1789360046387 36.998119354248
23.1389198303223 -41.4457092285156
-29.244047164917 7.64552593231201
55.1877517700195 0.012973764911294
37.0749816894531 67.6428985595703
-26.4021148681641 97.7504119873047
-29.2760601043701 7.64208650588989
-14.1388053894043 -17.3400859832764
12.807822227478 -77.3144073486328
23.3334980010986 -100.742523193359
44.7215156555176 -5.84346532821655
39.5326690673828 -11.6018238067627
39.5796012878418 -52.3999099731445
-9.57880115509033 -4.10242605209351
26.4311084747314 9.64022731781006
-8.75598335266113 30.2857818603516
18.9326305389404 32.1063079833984
-89.1660842895508 -9.93179321289062
34.6748924255371 -12.0530385971069
-27.8246192932129 47.8424873352051
48.839542388916 -81.3272323608398
-56.3789863586426 41.5656394958496
58.175838470459 1.22972071170807
-11.4945106506348 -49.6822929382324
36.3821487426758 59.8043212890625
-26.0002269744873 5.17541122436523
10.6128025054932 -46.1345138549805
-57.6747093200684 8.8924732208252
-17.7005348205566 100.392456054688
38.9869995117188 11.6083936691284
9.00825691223145 -37.7638359069824
23.9840202331543 7.06391668319702
-50.3615493774414 -29.7436428070068
24.4667015075684 -66.6996688842773
-15.7088613510132 28.9768295288086
52.5972366333008 7.75130987167358
-15.0335779190063 -61.3109130859375
43.2971534729004 -76.0288467407227
-53.1288909912109 -47.9343643188477
-70.8812255859375 38.4930686950684
19.8220520019531 14.1212663650513
-12.5076360702515 51.4227600097656
-36.5508842468262 54.5609817504883
22.6140575408936 -95.5808944702148
52.1194190979004 -7.45278739929199
18.1957988739014 53.0053863525391
39.1872177124023 -13.7225399017334
-40.9753532409668 70.8468475341797
4.93563652038574 -28.5882625579834
-0.051065731793642 -33.4864921569824
-59.9695587158203 5.48137187957764
23.7179489135742 -79.3755645751953
36.4333343505859 -80.0461959838867
-17.2168407440186 -11.8173274993896
46.427059173584 14.3625335693359
-44.5677299499512 51.9777679443359
-13.2518033981323 -58.3168907165527
24.2211093902588 1.62640416622162
-22.3068370819092 -2.50088286399841
-53.762149810791 -21.5643692016602
-7.14675855636597 55.0070266723633
-57.8178329467773 23.6023216247559
-67.9904098510742 41.4016876220703
49.7790756225586 -86.0553817749023
-3.26024842262268 -30.0372104644775
43.4421234130859 -52.3907585144043
60.7736320495605 -37.5528259277344
38.357551574707 38.8813095092773
-3.84183549880981 63.7218589782715
14.0033006668091 -2.41481804847717
-22.2731399536133 32.1628341674805
-46.001522064209 5.24034214019775
-25.8593044281006 0.40646967291832
-3.92954993247986 61.2227935791016
3.23953700065613 -13.0238084793091
-66.1395263671875 63.0974006652832
54.428638458252 -80.1040725708008
-50.0697326660156 -4.86638116836548
-42.2293167114258 10.03648853302
-70.2580947875977 65.8434295654297
1.25217258930206 -57.3466491699219
28.7533512115479 -51.9016494750977
-32.3381118774414 49.9543075561523
10.8719816207886 -94.3409271240234
28.439884185791 39.6895370483398
23.7415351867676 24.8886566162109
0.0586681813001633 56.130485534668
17.1885299682617 28.374547958374
-17.7821788787842 5.67174959182739
3.72641706466675 56.4627532958984
18.4866104125977 8.01149940490723
28.2743129730225 -53.6302642822266
17.3729267120361 78.051383972168
-71.8167877197266 -30.3573780059814
0.466324150562286 -17.2937717437744
-14.6221370697021 -47.1876106262207
-28.7602310180664 -42.5260925292969
76.0337677001953 -62.4969062805176
-24.4743404388428 -35.5742454528809
-36.3379287719727 61.3513298034668
13.1980152130127 69.9492797851562
-46.0035781860352 5.25451135635376
-45.4236297607422 72.0156021118164
-8.15412902832031 -54.5452194213867
-21.3845844268799 7.49844408035278
60.1095161437988 29.9567756652832
63.6686210632324 -77.4356079101562
-29.6128692626953 -69.2815170288086
36.9900970458984 48.1465034484863
25.3369636535645 -44.7625961303711
-14.6749649047852 -79.906623840332
5.54153108596802 -39.2567138671875
24.8304920196533 -18.0145168304443
9.78762912750244 73.7536087036133
-3.67364406585693 45.735725402832
-51.3585777282715 -5.92134237289429
-16.9776191711426 8.20602130889893
-53.319091796875 -7.95559024810791
-73.8429260253906 7.86787414550781
-59.1021842956543 -0.544853091239929
32.4971313476562 -67.3413162231445
6.27641439437866 83.9837265014648
-30.3210067749023 -39.5978240966797
-24.1576404571533 -76.6713638305664
7.12114286422729 -56.7941970825195
-49.2938385009766 31.9205760955811
55.4296722412109 -83.7873458862305
-15.7996969223022 -64.0383834838867
-25.9619789123535 -39.2251434326172
-26.0711364746094 -12.706579208374
-34.624584197998 -23.7667274475098
33.2251510620117 -52.8151664733887
-21.3781490325928 -9.24608993530273
37.9572410583496 -70.4272232055664
51.6496772766113 17.2448329925537
-64.1011199951172 -20.632942199707
30.4975986480713 -78.3243637084961
-55.9125099182129 51.3213577270508
-11.4702939987183 -6.29994058609009
-44.0872802734375 75.3114624023438
-63.1407432556152 73.891716003418
-19.8314743041992 -39.1781120300293
-47.9503898620605 -44.2970428466797
34.6040458679199 8.9972505569458
-49.074951171875 -24.2052040100098
-75.7253112792969 -4.53796625137329
-11.2654390335083 -24.4615707397461
53.3841094970703 -32.0169105529785
21.7070999145508 51.8484535217285
-31.894832611084 45.1505966186523
-9.3747444152832 -3.83961653709412
63.4889984130859 -8.28697395324707
26.7562599182129 -81.4492416381836
13.3997783660889 69.928825378418
14.1910028457642 20.9513912200928
-25.7290420532227 28.6516094207764
24.5123729705811 -47.6246643066406
-20.6032485961914 -19.600866317749
35.7622108459473 -30.691068649292
11.2961492538452 62.3837013244629
22.6704998016357 -78.5138320922852
20.4097270965576 -89.4226455688477
38.2782135009766 -99.0074844360352
11.5944595336914 58.0836486816406
-55.543285369873 5.34963512420654
-12.8041677474976 -56.8891143798828
5.01171064376831 -67.4069900512695
4.4493932723999 -34.1462745666504
-43.0261726379395 62.7176132202148
81.9225387573242 17.2141590118408
-55.6576309204102 -2.49507784843445
-26.8722286224365 30.1592025756836
55.6324653625488 -24.7440643310547
55.5868148803711 -39.8163070678711
29.0934066772461 -45.9067497253418
8.31020641326904 27.7116851806641
32.8439865112305 -72.0264739990234
-49.4609794616699 -11.1300554275513
-65.2707443237305 28.8614368438721
-25.9632720947266 0.414748281240463
-9.26763153076172 51.3041610717773
-62.8284378051758 29.3487243652344
45.9782409667969 -56.8976287841797
-52.1288604736328 -1.34856688976288
-12.1957082748413 4.52983570098877
18.3101348876953 57.4100189208984
21.946704864502 -44.9192504882812
-10.7081670761108 -81.8503036499023
81.5499420166016 17.6880187988281
-19.8329334259033 32.0615196228027
4.76231908798218 46.5767707824707
-60.5884284973145 55.0180740356445
73.8064346313477 2.57429122924805
26.9220504760742 -63.4641609191895
64.9911270141602 -46.4762878417969
18.8365936279297 -35.8444480895996
-17.9828472137451 -36.6518249511719
-67.9799499511719 18.5042362213135
11.0271167755127 -52.942211151123
45.1596527099609 -89.1565704345703
-36.0112953186035 -51.9523811340332
33.091194152832 46.4251480102539
2.02315449714661 38.223575592041
43.2191276550293 -35.567066192627
-4.41271543502808 38.3772888183594
-12.487606048584 -28.9089832305908
28.4084949493408 -28.5274753570557
65.2153930664062 -17.393045425415
-72.5526351928711 41.677734375
16.7910861968994 -77.0765228271484
8.30664348602295 27.7140789031982
56.2462959289551 31.0230674743652
-45.4076805114746 72.0212554931641
31.0547161102295 -91.0861358642578
-14.2770929336548 84.14501953125
-14.9639596939087 47.0697937011719
58.2144241333008 -3.85522627830505
-6.6051173210144 -24.2289066314697
-28.2932167053223 85.2200317382812
2.00552368164062 0.0662505030632019
33.1275253295898 -24.2090129852295
-89.1660232543945 -9.93119144439697
-15.1216163635254 -5.86972427368164
-15.1458797454834 25.197208404541
-6.32946586608887 63.026798248291
-43.4332084655762 63.8251762390137
52.6423110961914 -44.4961395263672
0.0434559881687164 2.36362075805664
11.6089563369751 -20.1500816345215
6.99877452850342 -87.8842620849609
-76.8720474243164 48.7785911560059
17.5766277313232 -26.5456848144531
-40.5418090820312 -27.3065147399902
-14.2864074707031 65.1830215454102
8.81477642059326 -43.040412902832
14.1704206466675 87.8577194213867
8.28959560394287 80.2759628295898
35.3916168212891 -19.6984729766846
45.8179588317871 27.0349159240723
-22.7057037353516 29.6437377929688
2.81312036514282 75.3235626220703
9.66703224182129 -86.9072265625
-39.0341453552246 91.1105575561523
13.5970678329468 36.7189826965332
-37.9823188781738 -22.2792720794678
73.7068328857422 -2.93283677101135
27.9424781799316 68.8108520507812
55.2946014404297 -36.0313758850098
1.96595585346222 -84.1399154663086
-21.2088317871094 -49.3823890686035
-35.5941429138184 15.8527851104736
-65.0111541748047 -0.357852727174759
12.1014566421509 27.9099388122559
6.33086919784546 21.878438949585
-23.3059406280518 -9.85410499572754
-29.389368057251 7.57904529571533
49.2490310668945 -23.3911933898926
-1.79422342777252 -50.7530899047852
19.5035419464111 -28.9388523101807
14.2423524856567 17.6848850250244
37.1326942443848 6.48196172714233
21.303337097168 -90.2865524291992
54.5601196289062 1.17305445671082
-29.7344779968262 21.9483032226562
-31.0668640136719 55.3086013793945
70.7160110473633 -27.8938140869141
41.6217918395996 53.2141532897949
75.6269989013672 7.94197988510132
22.7414150238037 -56.7673149108887
57.7181434631348 15.3867511749268
3.03051352500916 28.1730213165283
-29.4750480651855 51.431266784668
-6.65818357467651 -33.0915641784668
17.1596031188965 11.6644353866577
42.4692115783691 -81.1018753051758
16.6977043151855 17.8535594940186
10.6137256622314 -64.8331680297852
50.085277557373 -70.5462646484375
36.3580322265625 -43.2707443237305
36.7844276428223 -31.2320289611816
0.281022757291794 -73.6572952270508
58.3553733825684 -61.0205574035645
-49.7326812744141 89.933837890625
-35.6634254455566 69.9238128662109
8.23991394042969 35.0352897644043
-13.2226133346558 -38.4834289550781
26.4546546936035 -13.6904716491699
40.4109420776367 -16.4407520294189
-1.80339789390564 -77.3722763061523
80.1193161010742 27.0829067230225
31.4442653656006 -56.8373985290527
-33.1684226989746 14.0345106124878
7.79058790206909 -47.0205001831055
-9.39495468139648 27.1643447875977
51.1937103271484 11.7813119888306
18.2617721557617 -45.2603530883789
9.524094581604 31.4447708129883
-47.1327362060547 27.1800746917725
14.5505867004395 17.6927051544189
-29.2135219573975 61.9171943664551
45.6076583862305 -10.6896390914917
41.0364265441895 -10.9034194946289
7.22548484802246 32.7723770141602
-6.32699584960938 33.3250427246094
26.3122825622559 87.7534561157227
3.02424883842468 28.1737651824951
-75.136116027832 67.3862915039062
20.2654190063477 58.4522514343262
49.9269599914551 26.7808971405029
0.883104503154755 -26.9777374267578
-55.4030838012695 67.0867385864258
31.9592018127441 -2.96563220024109
-8.32564735412598 -14.5378675460815
-10.6732664108276 66.0868988037109
-58.3036041259766 19.7138328552246
46.1164321899414 31.0418109893799
9.21284770965576 66.4373397827148
24.0178565979004 81.3976135253906
-36.005916595459 -51.9089546203613
-0.197995573282242 -42.3744888305664
-76.401725769043 67.4477844238281
48.5367736816406 -76.9530715942383
6.76680374145508 -82.063606262207
-43.0726051330566 -19.7742710113525
-23.7456111907959 21.3240699768066
-24.5760326385498 73.1255340576172
-14.5638523101807 -30.3424205780029
-24.6573791503906 7.04412078857422
-4.18410968780518 -20.142448425293
-67.9572982788086 70.3490829467773
-36.8516578674316 78.7099685668945
13.2844362258911 -51.1179809570312
30.1599388122559 -8.57676315307617
18.3770198822021 57.3541450500488
-20.4185771942139 39.6341934204102
-36.2313804626465 -80.9286575317383
-55.6520767211914 -20.773983001709
62.6732978820801 -0.654761493206024
-10.5286340713501 -22.2649097442627
-67.2915191650391 49.4610366821289
-53.4997787475586 -21.717622756958
-43.0177841186523 7.37294387817383
-22.9521636962891 47.7305946350098
55.795036315918 -54.2499237060547
-68.0024871826172 31.3510627746582
-67.4125671386719 8.34248924255371
39.4701499938965 -80.5195693969727
-72.53564453125 21.0604114532471
};
\addplot [draw=color1, fill=color1, mark=*, only marks]
table{%
x  y
-14.1562461853027 16.2737731933594
15.9061107635498 0.851377665996552
-14.5903673171997 32.9087791442871
24.7583923339844 -52.4581413269043
1.76987397670746 -18.1802673339844
12.3498573303223 -70.0374374389648
-10.3571968078613 74.6969299316406
-49.0271110534668 -17.9125480651855
22.1610584259033 -49.2137107849121
36.1605529785156 -17.489049911499
-18.2265892028809 -11.663556098938
-5.55481910705566 73.7786407470703
13.1064291000366 -44.5872497558594
46.4859504699707 -35.5428199768066
14.6336479187012 -65.2039794921875
-10.5388031005859 65.3314514160156
-6.78564691543579 26.3422622680664
0.797948896884918 76.6471176147461
35.6403961181641 -61.0181159973145
28.7877178192139 -49.7694931030273
-12.8304452896118 -14.748987197876
6.28536701202393 -48.0450553894043
4.0225887298584 26.9253406524658
49.203483581543 -13.8960771560669
10.4839687347412 79.8198547363281
1.43978202342987 -64.7868728637695
9.88216018676758 44.1859855651855
-54.984489440918 -9.38840389251709
-41.3730010986328 -12.4610357284546
-23.3574848175049 18.0422458648682
33.5815658569336 16.6903343200684
45.8693923950195 -34.5644378662109
-3.67634797096252 73.5894012451172
29.9919891357422 -14.1625623703003
18.2956447601318 -43.659553527832
-23.6551246643066 17.2828235626221
-0.523545205593109 -33.7187957763672
-3.46174550056458 25.0130176544189
-13.303617477417 61.720516204834
8.45828437805176 57.3310928344727
25.74001121521 -3.36165690422058
30.3130989074707 -61.7557830810547
-9.81210994720459 -66.5528717041016
-7.22028207778931 41.8842697143555
27.6234664916992 -26.0225315093994
-23.6735897064209 23.3374004364014
25.8328628540039 -44.4100875854492
-47.7951469421387 -8.54372787475586
-9.06405830383301 -16.9985446929932
3.21282887458801 -58.4978675842285
-43.9842987060547 -6.30048370361328
2.99738717079163 23.0593242645264
15.7514410018921 10.014009475708
-42.0615005493164 15.3201522827148
-0.909237623214722 -9.83153057098389
-10.9196891784668 -22.6806621551514
-28.9560070037842 -38.3873100280762
-21.7516994476318 37.6689529418945
-0.90399307012558 -12.597393989563
-10.7759971618652 79.7270126342773
6.95509386062622 -31.4997692108154
-1.24866032600403 -48.9488410949707
1.08647096157074 -73.665901184082
31.7287845611572 -2.52046322822571
31.3026351928711 -4.13925886154175
-9.25134658813477 -46.2501754760742
38.2947578430176 -5.81643438339233
0.778681576251984 -26.6699714660645
13.3758659362793 -49.5960464477539
-2.4627513885498 -37.982551574707
18.2552795410156 -45.910270690918
36.0335083007812 -43.0681915283203
-45.2637023925781 0.117252051830292
-20.7245254516602 27.6910781860352
4.71429300308228 43.474235534668
-8.54316520690918 29.0524082183838
-10.6342191696167 62.4147338867188
-25.5382080078125 2.76588034629822
-20.9436187744141 -10.1609754562378
33.8535652160645 -8.42477703094482
-41.7106094360352 17.5954570770264
-42.6732711791992 30.4063396453857
6.28234338760376 7.22087717056274
-4.92992496490479 -56.8990135192871
-11.8856220245361 38.4550132751465
8.65184688568115 0.349969923496246
30.8050785064697 -12.5851440429688
-8.84524250030518 21.4997024536133
8.73474788665771 78.5139312744141
13.9132356643677 -56.6436729431152
15.94407081604 19.3828907012939
19.3564701080322 -4.94871473312378
-48.4773941040039 -23.1676368713379
-11.1856412887573 7.2453989982605
-1.52089750766754 -30.7742691040039
1.84971594810486 37.3751029968262
-5.7638635635376 8.2220287322998
10.7414426803589 1.55196952819824
-15.4240303039551 -29.0445556640625
22.3318481445312 27.5244808197021
37.5331916809082 -62.9953689575195
27.2599811553955 -48.8286933898926
-9.75624656677246 65.166145324707
-0.649470210075378 13.6928176879883
-28.4043292999268 20.3937892913818
39.9670448303223 27.7662830352783
19.4506244659424 -2.32283234596252
-16.0902843475342 -60.8875999450684
8.23221015930176 55.2768592834473
7.78045177459717 63.0177726745605
33.5156669616699 -37.2890853881836
24.4366474151611 13.5936975479126
30.8764934539795 -20.917142868042
45.1145668029785 20.4626750946045
25.9215526580811 -48.9129219055176
11.8244342803955 -50.2865715026855
-17.0146446228027 -19.9688949584961
11.3605842590332 -39.7386283874512
-31.621208190918 4.6064395904541
2.80386805534363 -40.207405090332
-8.1839075088501 30.9129390716553
13.5803813934326 -66.5836944580078
-17.5457954406738 24.6219291687012
-30.2811622619629 -15.9232082366943
-10.8703584671021 -2.81906247138977
-2.95088005065918 -38.0437088012695
21.6684036254883 -66.0403289794922
-16.8177680969238 -2.28541779518127
11.2980442047119 -4.48066854476929
16.5028839111328 -10.287504196167
-32.2802047729492 -13.6332426071167
-47.3634262084961 -11.6237506866455
17.3558483123779 -76.4602737426758
4.3209376335144 -12.1043319702148
-21.5901622772217 3.09332180023193
21.5516510009766 25.0099468231201
-24.203426361084 25.5471076965332
-8.62772560119629 67.5758895874023
8.06674289703369 33.1486778259277
-4.65688323974609 -1.56843042373657
6.29899215698242 67.7486572265625
-10.5629920959473 -22.7748241424561
26.409646987915 -51.4696846008301
12.7960977554321 76.7983474731445
-39.6081924438477 -1.63019394874573
2.3358690738678 -28.6222839355469
44.9914283752441 -53.0035552978516
-47.1891784667969 -3.51396441459656
-17.2614307403564 14.4307317733765
27.7461719512939 -12.9975690841675
-14.8937530517578 70.0156936645508
-19.1048164367676 15.080099105835
29.5145778656006 -40.4345779418945
-11.9497404098511 66.8091354370117
31.4417381286621 -9.81137275695801
40.2962608337402 11.2909832000732
-4.69818592071533 -16.7863788604736
25.4435920715332 -46.8182754516602
-13.724157333374 -19.9166126251221
4.88891696929932 24.9782314300537
17.5421257019043 25.3467330932617
-1.54870390892029 27.5834064483643
-35.4704093933105 -5.87784862518311
-12.174446105957 11.2807664871216
-17.2362785339355 55.619499206543
-22.6500720977783 -28.7325820922852
-38.5535430908203 12.7513599395752
15.0858554840088 -60.9970588684082
40.2745246887207 10.7555618286133
2.10408425331116 -43.9000816345215
2.43483734130859 15.0179214477539
3.69317245483398 -32.3478851318359
-12.121976852417 82.5328826904297
29.1320610046387 1.27591419219971
32.0222129821777 25.2803649902344
-46.8451690673828 4.17712068557739
36.4329147338867 -49.6340103149414
-12.5656356811523 -47.868968963623
2.3995053768158 3.609938621521
-14.246506690979 -22.5724544525146
25.0124397277832 -25.8872509002686
-0.254881918430328 31.9352989196777
-12.4840927124023 23.8146667480469
-31.6277961730957 -15.385142326355
-35.646427154541 -21.0412502288818
9.02097988128662 61.0992965698242
23.1384620666504 -61.568775177002
5.37125873565674 3.65267586708069
1.76970970630646 -8.3620548248291
-11.1585922241211 80.6489181518555
-34.6866912841797 -28.5431785583496
-2.43847513198853 -54.868049621582
-2.0417959690094 -1.06925773620605
40.1018676757812 4.54694747924805
31.3149070739746 -5.08900690078735
2.91170954704285 46.0739555358887
-16.4742374420166 19.0722599029541
4.82551193237305 -25.9199619293213
-48.4812698364258 1.47474801540375
-22.6564407348633 -5.44893312454224
-39.1985664367676 -8.78360557556152
-41.8020248413086 15.4914474487305
38.3638916015625 -26.1201820373535
-16.8518733978271 77.0881042480469
-35.7138023376465 16.9489593505859
-7.21964645385742 -45.8839645385742
34.8170280456543 -12.6864891052246
26.6888599395752 -59.169303894043
-19.6623344421387 19.8142642974854
-7.39723062515259 13.1846542358398
45.6438179016113 -4.15494441986084
0.111052051186562 -33.7903213500977
-3.46785235404968 -9.97296714782715
26.0671405792236 0.227064251899719
26.5278739929199 4.71392631530762
-7.33442544937134 81.3036270141602
-35.3015899658203 31.5906181335449
1.35896062850952 58.5727119445801
1.5297212600708 71.1305923461914
-10.010570526123 23.5384063720703
18.9875087738037 -6.82228851318359
6.80627346038818 23.0084571838379
4.11152935028076 -34.8415184020996
-12.7648687362671 -47.0462188720703
15.9349784851074 -21.9359931945801
28.6154327392578 3.10245203971863
-28.0933208465576 1.6114673614502
9.18317985534668 47.0225067138672
-12.45578956604 68.5944442749023
21.3322200775146 -70.7522659301758
30.226448059082 2.85382008552551
-0.521462440490723 -47.6239852905273
37.0535774230957 23.7878074645996
-1.67553639411926 61.4517555236816
34.4644203186035 -12.1803646087646
28.5689296722412 -22.4197235107422
-32.4310493469238 -18.4039688110352
39.0668754577637 -12.8985385894775
45.6493759155273 4.99477672576904
11.5254974365234 41.074592590332
-8.21777820587158 38.5960464477539
21.1782970428467 7.94684505462646
-15.5823822021484 20.4763259887695
-7.84456348419189 33.6721115112305
14.9224367141724 -44.2074966430664
14.3978500366211 -32.9720993041992
-43.2316627502441 -20.4248332977295
-29.3081855773926 20.3523845672607
-20.6603851318359 -3.9294376373291
-1.38323223590851 47.1030387878418
-27.3588199615479 39.5297660827637
24.0691299438477 -50.2185173034668
-45.5678291320801 -25.494592666626
-8.38327789306641 -2.32737755775452
0.616246342658997 -39.3792991638184
-6.9112434387207 35.1056709289551
36.5442810058594 18.2156276702881
-6.54022693634033 22.064790725708
-42.7417068481445 27.1025981903076
-2.0668785572052 -15.1366767883301
-32.1374626159668 -23.2694110870361
-51.4727516174316 -14.646125793457
9.67971038818359 10.4945592880249
34.4755973815918 12.4918699264526
-50.3344345092773 -4.68861103057861
19.0577411651611 13.9575262069702
11.7904024124146 -37.3484649658203
10.4059038162231 -48.5614318847656
-6.00288057327271 -8.90501689910889
34.664249420166 12.0318851470947
17.9467601776123 -66.4420394897461
21.5221843719482 -59.7803802490234
11.1702671051025 74.4532623291016
15.7115955352783 -64.0511245727539
6.62478828430176 38.1891136169434
-39.778018951416 -25.5806522369385
-39.5218887329102 14.0598936080933
1.42190909385681 -53.7553443908691
-10.9876232147217 69.3366317749023
-7.99409246444702 20.985725402832
-33.2599143981934 27.005298614502
-21.3398609161377 23.3442535400391
18.7808227539062 -75.8148574829102
-9.13871479034424 49.2513580322266
-22.5335121154785 -29.2492485046387
11.6988506317139 75.239616394043
-6.66132497787476 69.9680023193359
20.1837615966797 -72.3973007202148
4.53920030593872 63.4861526489258
27.4785194396973 -56.5249290466309
31.1390323638916 7.47712659835815
-2.16996502876282 -58.0512161254883
32.9060592651367 -49.1143951416016
-13.7541513442993 -4.27445602416992
24.6208076477051 4.66902160644531
-20.3429908752441 3.30881142616272
-29.2848739624023 -26.0461730957031
26.9002685546875 -46.6657180786133
-17.4212207794189 34.3454704284668
0.481921970844269 5.13405418395996
-6.70317602157593 16.1393127441406
-14.9225292205811 -52.051643371582
-12.7421054840088 5.92016887664795
-44.0190124511719 -1.4125919342041
5.69720506668091 56.2550811767578
-5.33053112030029 -28.6760196685791
-43.5013122558594 1.70440280437469
-16.7730884552002 -50.5043869018555
-54.7032814025879 -14.4245710372925
-16.3418769836426 25.6480083465576
-53.2781944274902 -15.987491607666
23.5342407226562 -35.0567817687988
6.20785188674927 64.8274536132812
-51.5338935852051 -10.556471824646
5.68330335617065 19.8132553100586
15.3247947692871 -3.25546836853027
-26.7605438232422 12.6324501037598
-17.1904792785645 -1.1943895816803
-5.54897451400757 63.4965934753418
35.7032318115234 12.815821647644
5.95640516281128 -67.5386276245117
-17.6672134399414 44.9177665710449
4.53430223464966 -28.5955810546875
14.6802082061768 -20.8955821990967
-14.3905334472656 11.6104049682617
9.05849838256836 74.4791107177734
-36.0064239501953 1.14881777763367
37.4585456848145 -18.2941722869873
-14.5936479568481 61.6761436462402
-20.5579738616943 -20.6917114257812
17.1725883483887 -6.50464487075806
-6.07322120666504 -24.980224609375
-3.1837592124939 13.4426889419556
6.84798860549927 -43.4136352539062
0.68505322933197 44.5727310180664
17.0947704315186 -64.4154815673828
-2.5180516242981 -64.8597412109375
8.40808582305908 -50.0519676208496
-53.5180053710938 -16.5135402679443
-9.29903316497803 51.9289207458496
23.4430599212646 -1.8377548456192
29.9761981964111 -53.7202911376953
-6.89573097229004 -49.0051002502441
17.1100425720215 -58.1905899047852
11.3685636520386 -11.402551651001
-3.10602593421936 65.9060668945312
-39.5226516723633 13.0065221786499
-1.43580591678619 69.1976776123047
-4.13788604736328 30.8474102020264
15.560097694397 -49.7316246032715
-39.9275054931641 11.8809366226196
2.13128209114075 64.247444152832
-12.0617790222168 48.4711875915527
4.42822170257568 -52.078182220459
35.5346374511719 -20.2437839508057
2.57543063163757 59.2436637878418
-39.2273483276367 13.9640045166016
-39.0242462158203 22.9863262176514
-12.306568145752 3.77273058891296
-30.7462024688721 1.61940360069275
35.7655181884766 -44.487865447998
1.52739584445953 66.1834182739258
10.0763473510742 -49.0631675720215
0.760641396045685 -60.115837097168
35.1040725708008 -53.1094245910645
4.08598947525024 84.7822799682617
36.7877464294434 -42.2259330749512
-9.05131912231445 -54.4618797302246
11.1937694549561 24.7227592468262
28.5881576538086 -40.3376121520996
19.4687156677246 -25.4935531616211
-31.2322673797607 25.4531726837158
-38.107479095459 -9.05575561523438
40.2524681091309 -55.6437835693359
-38.8869400024414 -24.7076225280762
-51.5100479125977 -23.7850437164307
14.2239561080933 -1.5298445224762
3.43313956260681 -26.457239151001
-14.7513008117676 59.8690528869629
3.2206859588623 11.7626342773438
-11.1817827224731 7.24995851516724
-13.5163869857788 34.2488327026367
-3.65907788276672 -27.7040939331055
-13.2587223052979 40.5102844238281
-25.1148815155029 -1.82188069820404
42.614875793457 -55.443660736084
-26.7906265258789 24.2013778686523
27.493932723999 -41.520191192627
0.206918761134148 -42.6391639709473
-36.6595687866211 -24.2476444244385
-3.31427526473999 -6.8505425453186
10.03773021698 -37.2256546020508
22.3766479492188 -10.2656364440918
-2.72840595245361 46.2721061706543
26.7294540405273 -30.5466651916504
38.8881492614746 -44.215404510498
14.4415721893311 -63.5295181274414
-37.0411911010742 20.3775463104248
15.9359560012817 -32.611888885498
-1.22140038013458 8.00126934051514
2.23746418952942 54.3661193847656
-32.4431686401367 5.31508731842041
-6.95743560791016 -32.4719734191895
-3.33114576339722 41.5269088745117
5.78294277191162 -27.7950344085693
-15.9254083633423 -12.0912179946899
-1.86363780498505 32.3100433349609
-18.5041275024414 -34.950080871582
-19.3557910919189 58.7318649291992
-12.6471309661865 -28.8995933532715
-29.3336219787598 -6.75573492050171
-12.5778541564941 -29.3947620391846
-0.755422353744507 21.0363597869873
-53.4739990234375 -11.9781694412231
9.53775215148926 1.18791878223419
-5.44722270965576 3.71906733512878
9.39107418060303 -62.8244514465332
25.2141914367676 -40.1923370361328
-23.5486850738525 -55.1394195556641
36.624927520752 -51.204517364502
-1.41397559642792 64.4651794433594
31.1220169067383 32.1042709350586
28.4181709289551 -61.9741172790527
41.0682106018066 -6.71240520477295
7.91067838668823 39.9108734130859
-1.29828250408173 -3.32939910888672
-9.48875617980957 72.2016906738281
-10.4276132583618 38.6326713562012
-49.5756340026855 -17.0155220031738
-1.23361301422119 85.5276947021484
-9.40144634246826 62.9727249145508
1.43499612808228 59.0431213378906
-6.21810626983643 53.1744232177734
-40.3940620422363 29.7566719055176
25.2305107116699 -40.0765686035156
14.0889110565186 -52.4261283874512
-4.53575658798218 25.081392288208
-13.8788919448853 23.9843482971191
-36.7962684631348 9.24637508392334
15.9282217025757 31.1022338867188
11.8129873275757 73.0501480102539
14.2319955825806 -63.4841346740723
-1.97849798202515 66.7942733764648
-23.7449798583984 -56.3668365478516
-48.5560188293457 7.43255281448364
-20.9436798095703 -41.8449440002441
-14.2904224395752 23.3435115814209
10.4972915649414 49.6534957885742
-3.99147343635559 -50.0040283203125
18.211742401123 -51.417423248291
-5.88881301879883 -8.80561828613281
-1.51741051673889 -47.4201202392578
4.32817792892456 31.6414756774902
-24.871789932251 -7.62801742553711
-20.4118881225586 45.7798843383789
27.4483509063721 -26.8081016540527
-5.56537246704102 -29.4588851928711
-16.0065059661865 -65.9480972290039
-5.21497488021851 -25.0645942687988
40.4045066833496 -20.6414604187012
-45.9770278930664 22.8458423614502
-26.5258178710938 -0.407515108585358
28.5873775482178 -46.7295265197754
-44.2382164001465 10.3070650100708
-50.5521125793457 5.60726404190063
-7.19392871856689 -42.993709564209
-9.5423583984375 -38.5154151916504
-43.3997917175293 -7.75084924697876
9.01693916320801 -51.2871055603027
25.0602397918701 13.2396965026855
-29.0721530914307 11.8301572799683
-27.9322357177734 -3.03503251075745
1.96551692485809 36.5476722717285
-7.27517414093018 82.8970489501953
0.202581092715263 59.3200950622559
33.6214828491211 4.04944849014282
34.3057708740234 -22.4389762878418
-23.6837158203125 -9.76546096801758
-34.6332054138184 0.0546604692935944
39.3979530334473 -44.4877548217773
-22.6435832977295 7.02360010147095
29.4733123779297 -25.9379234313965
-33.2400817871094 5.08824300765991
-38.6765251159668 4.40700483322144
9.50541591644287 18.0882816314697
10.8913688659668 -52.0722389221191
32.0797653198242 -33.75390625
-6.8299732208252 1.5664496421814
27.8091373443604 -42.6121864318848
7.99870491027832 47.2746238708496
-47.7646560668945 19.1078987121582
-0.175539493560791 -6.24241399765015
-15.9347543716431 -23.7865829467773
3.07636833190918 -4.72971487045288
37.8014335632324 -23.3688488006592
18.0060062408447 42.4501457214355
13.2788887023926 11.630054473877
-7.13648319244385 -29.9890613555908
0.0621891058981419 87.6787338256836
7.98978853225708 -35.625072479248
};
\end{axis}

%% file: pgf/emb2.tex

\definecolor{color0}{rgb}{0.12156862745098,0.466666666666667,0.705882352941177}
\definecolor{color1}{rgb}{1,0.498039215686275,0.0549019607843137}

\begin{axis}[ name=emb2,at={($(emb1.east)+(1.5cm,0)$)},anchor=west,
tick align=outside,
tick pos=left,
title={Iteration 2},
x grid style={white!69.0196078431373!black},
xmin=-100, xmax=100,
xtick style={color=black},
y grid style={white!69.0196078431373!black},
ymin=-100, ymax=100,
ytick style={color=black},
xlabel={embedding $1$},
]
\addplot [draw=color0, fill=color0, mark=*, only marks]
table{%
x  y
-59.3989906311035 -3.88174295425415
-52.1980133056641 -29.9600734710693
-23.8700866699219 14.8747043609619
45.9644012451172 -72.8355331420898
18.4263343811035 -18.6554489135742
72.3069076538086 11.0514307022095
-12.3835382461548 62.7439842224121
1.45318341255188 45.9724082946777
38.1082382202148 -30.5641174316406
54.2327880859375 37.2595710754395
-50.9309349060059 -7.92653179168701
-22.5761680603027 66.7992248535156
58.8991279602051 -43.1910400390625
61.4539566040039 -49.7814102172852
58.2523498535156 -22.9961128234863
-10.0876932144165 76.8169403076172
20.1417694091797 11.577691078186
-25.0688610076904 72.656135559082
32.7763710021973 -68.5729446411133
43.774055480957 -47.319263458252
-8.18664264678955 -6.68431043624878
61.2906723022461 -17.7328300476074
-56.5929985046387 13.0510787963867
26.403600692749 19.5474452972412
-49.0229454040527 65.834098815918
80.6248397827148 -6.40851020812988
-36.9550437927246 57.4287910461426
-76.9757766723633 -9.87645435333252
-50.0519790649414 1.24364972114563
-45.7655029296875 -10.9576644897461
-68.0342025756836 -15.5926961898804
8.11697101593018 12.6379137039185
-34.6995468139648 49.5712051391602
27.6488857269287 6.658118724823
51.1296463012695 -16.5664176940918
-39.5692901611328 -2.23570156097412
46.1967811584473 2.90700578689575
-26.5735702514648 28.4042854309082
-48.2714958190918 82.0399551391602
-44.5700759887695 66.4654235839844
-32.1709747314453 -9.29514408111572
46.1248054504395 -63.8987846374512
34.651050567627 26.2713394165039
-15.9287643432617 -30.7664432525635
-14.581859588623 -33.2216491699219
20.257209777832 50.0617332458496
67.8618927001953 -8.28883647918701
3.37970495223999 52.6689147949219
55.7906455993652 -13.5965938568115
12.4347591400146 -1.88904201984406
26.3612022399902 25.7781429290771
-16.0075702667236 -5.73361730575562
-39.3100280761719 -25.3935527801514
-21.9485607147217 1.03800225257874
-48.8090553283691 12.6146001815796
0.795708119869232 -45.8782234191895
28.8898296356201 -70.4913940429688
15.1721210479736 -18.4402351379395
-13.7505655288696 -45.2579879760742
-40.8381652832031 81.0577087402344
36.7289543151855 -17.6083393096924
68.57666015625 -4.2831711769104
18.1650562286377 -37.947093963623
9.40101718902588 17.4554061889648
5.72623348236084 2.3608341217041
42.2327194213867 -44.2584915161133
44.8672027587891 22.535120010376
61.0741271972656 -14.1680221557617
56.1815605163574 -0.412944078445435
7.30303144454956 -0.810533285140991
47.6955337524414 -35.0373649597168
40.4625816345215 3.96519875526428
14.6635913848877 17.4803352355957
-39.4301948547363 33.9675979614258
-31.8878650665283 51.699592590332
-26.1130313873291 -2.08763408660889
-32.5507774353027 52.1364860534668
-58.2698554992676 7.10386037826538
29.452579498291 -37.9644889831543
19.7859439849854 -60.1365509033203
-25.514440536499 -15.6069660186768
-26.4914836883545 -7.46048307418823
0.879397511482239 -21.7175598144531
36.8755416870117 0.459950834512711
18.8285751342773 -26.4703998565674
-12.2708702087402 -30.0855579376221
3.25344800949097 -12.7354316711426
-20.5203437805176 62.6723480224609
-34.7725372314453 67.0757217407227
52.8978309631348 -15.226318359375
19.9529476165771 -57.9644927978516
-48.1834335327148 -30.4872341156006
-0.0812728181481361 31.3325386047363
-74.9028167724609 -4.52485084533691
19.489200592041 37.6603050231934
-25.2416934967041 -10.3873701095581
-46.3800582885742 14.1078805923462
-6.18200302124023 -8.57003402709961
50.0472946166992 -20.225923538208
-12.296971321106 20.688648223877
34.1648101806641 9.38941955566406
74.6129302978516 -43.6138229370117
-44.7393417358398 75.8535003662109
13.5938720703125 -11.826473236084
-7.36279535293579 14.8440847396851
-69.0080337524414 -18.5043964385986
67.8615875244141 -34.37939453125
58.6337203979492 3.96834206581116
-28.146427154541 46.5085678100586
-47.419319152832 44.5567626953125
43.3062591552734 -39.751335144043
5.81577110290527 20.1478080749512
21.1640739440918 26.4635200500488
-45.5399703979492 -32.1087379455566
55.7591743469238 -39.2527236938477
62.1665687561035 -35.8036651611328
34.757152557373 -1.43468940258026
8.12178897857666 -8.8945140838623
-31.5656967163086 26.0276870727539
39.818733215332 -20.8631439208984
-31.0394954681396 0.955881655216217
59.4547500610352 15.1434230804443
-13.8101634979248 11.6420450210571
3.70125460624695 -35.2027702331543
20.955249786377 -5.76408910751343
41.9285469055176 14.7503681182861
47.5923042297363 -39.3766899108887
-60.457691192627 -10.1247577667236
23.4971084594727 20.420202255249
33.7751235961914 3.99141979217529
5.2456636428833 -35.0663032531738
-36.9814491271973 5.46620464324951
79.1913528442383 -37.5640144348145
23.7385883331299 -22.9336051940918
-9.48399066925049 -44.8162460327148
-9.2557201385498 22.6291351318359
-29.7932548522949 -13.9379873275757
-28.2205944061279 57.7812080383301
-18.7021198272705 14.3815689086914
-71.0723114013672 -4.63226842880249
-46.9271354675293 73.2849044799805
34.3288383483887 1.27177536487579
54.7293128967285 5.22821712493896
-28.0949039459229 73.7138671875
-19.3658199310303 -0.0408369414508343
5.17434358596802 -36.330078125
50.9830741882324 -63.1721076965332
21.0291748046875 -26.8175106048584
-38.0682067871094 2.43904089927673
59.8169708251953 11.6546640396118
-39.4730186462402 68.2189407348633
34.4089736938477 4.44350862503052
44.4119720458984 14.3907823562622
-34.0979118347168 31.3583583831787
48.5569763183594 9.39108180999756
-42.116771697998 -29.8529357910156
36.0764198303223 -8.17378997802734
52.5448112487793 -65.7737121582031
48.6089057922363 9.35070991516113
6.83019685745239 29.0661010742188
-9.53365612030029 33.6760482788086
-50.4103050231934 31.8982334136963
-49.0275115966797 1.04745733737946
-34.6633987426758 12.2821025848389
-22.4950523376465 56.6228408813477
8.48038387298584 17.2510166168213
-7.01185131072998 5.18603992462158
33.6945381164551 -40.5916481018066
6.41020679473877 8.19158172607422
60.6389541625977 18.3586330413818
-16.3632221221924 -10.01673412323
12.9059152603149 6.38405561447144
-36.7435836791992 78.7849655151367
52.3454513549805 -21.5072422027588
-58.0724487304688 -19.9035682678223
-28.0434112548828 20.379114151001
33.6542091369629 -59.7471122741699
-10.1208429336548 -15.4552965164185
-35.8588447570801 -29.1452960968018
52.5030250549316 -12.3222036361694
0.999411880970001 -64.7572326660156
-14.4102468490601 -13.0091924667358
14.4581165313721 46.3643836975098
-44.3120460510254 14.0982732772827
31.4083499908447 30.6097412109375
-37.473876953125 62.8971214294434
52.4400405883789 -5.46604442596436
-29.3257350921631 -22.9098052978516
-44.0082740783691 -19.5236721038818
-39.9195709228516 83.4554138183594
-41.1397285461426 47.4687423706055
71.2676315307617 -39.4677200317383
9.29929733276367 31.0564708709717
13.9979190826416 -27.6271438598633
64.1859817504883 -22.318265914917
-7.88691139221191 54.1349792480469
-26.5304889678955 2.04200172424316
3.31811833381653 3.59719967842102
-42.2843589782715 -1.63987755775452
24.1633396148682 8.00628185272217
-56.2584266662598 -10.3211784362793
-18.6256942749023 7.24935674667358
54.142333984375 -14.5199251174927
-22.4391937255859 65.0281219482422
-70.4204330444336 -5.12173318862915
17.7595710754395 -22.6750507354736
-7.14825534820557 -33.5689582824707
82.3786926269531 -3.20420145988464
-47.616081237793 18.8638076782227
-2.37386131286621 28.125452041626
-11.4208097457886 -0.0132297649979591
34.711742401123 -34.218189239502
18.4841117858887 -29.8056125640869
13.507511138916 -3.65068030357361
71.8787841796875 -42.7243232727051
-42.2383460998535 86.946907043457
-8.73406028747559 17.3507213592529
-32.5736236572266 76.5084381103516
-38.7352294921875 52.3309555053711
-30.791130065918 -27.6454792022705
7.72406911849976 4.99949789047241
2.98966360092163 37.5302734375
19.5705814361572 -23.2182846069336
7.50603580474854 -9.76622009277344
24.1124496459961 4.98121356964111
47.9165267944336 -7.09372568130493
-40.3194046020508 28.8689002990723
27.4657535552979 41.9828071594238
-43.923755645752 84.5519180297852
39.5291137695312 -20.8607902526855
13.5486783981323 1.70230710506439
84.5519180297852 -10.5820827484131
-68.9961090087891 -16.0907115936279
-24.5514831542969 67.4720916748047
29.3040714263916 7.46895551681519
50.9293060302734 18.2334365844727
19.2450103759766 2.56654071807861
2.36419177055359 -32.7943534851074
58.6967010498047 -33.187557220459
-12.3060646057129 37.0289421081543
-19.6004123687744 51.8837280273438
27.5616912841797 -19.1526069641113
8.54753971099854 32.795970916748
-20.411205291748 38.3117446899414
27.8729419708252 -53.4984741210938
20.2859973907471 -75.4083099365234
-38.7425537109375 -8.03707218170166
-40.5596313476562 24.8831920623779
-74.3020935058594 -17.1867179870605
-68.6639785766602 26.908899307251
-21.4956398010254 2.73554587364197
86.8826675415039 -15.2385663986206
42.6218299865723 -56.2890586853027
7.10129880905151 -9.5288028717041
61.1347770690918 -42.054515838623
-20.9888610839844 63.3413543701172
-18.7487411499023 -8.84112644195557
-6.6768593788147 -11.5698852539062
-71.4968338012695 -10.892991065979
-59.9933280944824 19.2149543762207
-35.2800788879395 12.8048686981201
-4.63287448883057 43.563777923584
-59.1229133605957 -29.7552108764648
-43.6125640869141 -37.7944869995117
-56.3725509643555 -20.6695251464844
75.6030426025391 -29.4717330932617
5.3095908164978 -52.1880378723145
8.97394371032715 -16.0830268859863
0.115210600197315 20.1055583953857
-45.7031440734863 -30.0208969116211
90.2393264770508 -21.9501800537109
35.4823608398438 -11.3835020065308
-36.4400520324707 72.9027252197266
75.3703002929688 -11.3495836257935
11.3728199005127 37.541805267334
-57.6957740783691 20.2870807647705
-57.0439109802246 -7.85005283355713
60.869514465332 -33.4365005493164
-43.5703773498535 77.1906661987305
-66.8210525512695 9.15075016021729
-73.4523773193359 -15.3547801971436
-14.9303426742554 -2.92157173156738
40.2957916259766 25.0242786407471
-30.4353866577148 64.6221923828125
-32.5106201171875 -9.55502605438232
-24.8972797393799 35.9228134155273
-50.7080535888672 51.6497192382812
32.39990234375 -23.7859344482422
-15.7249851226807 77.4506225585938
73.3279266357422 3.21253037452698
-4.0570707321167 -59.8306617736816
60.2347984313965 -37.1465644836426
58.6439399719238 -15.4414167404175
-26.7096729278564 -17.4291725158691
70.7139663696289 -4.93206739425659
3.43090963363647 22.9970607757568
-0.446505397558212 -45.6635818481445
77.2161712646484 -33.7345581054688
-21.2819881439209 6.1115837097168
-25.5696620941162 -3.40481805801392
-0.153866097331047 -0.585077285766602
79.8786697387695 -20.2555236816406
9.64826107025146 17.471736907959
-46.8657493591309 -29.3196334838867
-13.5983533859253 74.8395004272461
42.6521835327148 -56.2629547119141
7.08123779296875 24.8181419372559
45.2584037780762 -28.9564304351807
0.400502055883408 50.9078102111816
-13.7732810974121 -22.2463226318359
-0.530891001224518 39.8518943786621
69.5099182128906 -16.6493759155273
-23.1634654998779 65.5608444213867
2.96448111534119 55.8111801147461
-6.8560619354248 57.3919143676758
63.5054397583008 12.0995321273804
56.0865325927734 -18.6970138549805
-47.8642959594727 18.298864364624
-30.8819255828857 59.1407737731934
-54.6786918640137 -17.7549934387207
65.6035766601562 -26.8683567047119
-25.8499526977539 34.9922714233398
37.8048515319824 -19.0297603607178
24.794319152832 -4.10150337219238
-40.2118682861328 -25.3640155792236
-33.5752296447754 51.8232307434082
-62.2018165588379 -16.1192016601562
21.9215412139893 -64.7725296020508
-35.4417266845703 53.1030464172363
37.1364517211914 44.0489273071289
9.50137424468994 12.9451131820679
24.1220436096191 4.96964073181152
11.7408227920532 -29.1793174743652
70.7820205688477 7.29525756835938
-33.4333457946777 49.856502532959
40.7781715393066 -17.9914989471436
13.5478572845459 -46.2429389953613
85.4374771118164 -48.3215560913086
-6.90166044235229 44.9188423156738
-13.8979005813599 85.4421768188477
-0.62027508020401 12.3474140167236
37.8450775146484 -23.5875968933105
8.81610107421875 -27.5583763122559
77.4424819946289 -48.46728515625
-33.0183944702148 -31.7651500701904
-20.971773147583 70.5363388061523
-35.6964645385742 -1.41628623008728
2.77059078216553 70.4951171875
-12.5862483978271 14.5554733276367
8.05156517028809 -58.4096984863281
-64.7065582275391 -13.555251121521
-32.9703407287598 80.4142532348633
-15.214207649231 -63.8855628967285
15.8186559677124 -43.2602920532227
10.2907552719116 -24.1504287719727
-20.4889297485352 48.5024490356445
-8.19021320343018 -6.67587184906006
-28.2075843811035 5.11156415939331
-11.5738077163696 39.6964340209961
-49.4079933166504 13.5674734115601
74.5886459350586 -43.5884590148926
-12.2257871627808 60.6362342834473
21.9207286834717 -64.7654571533203
12.5524625778198 -50.6407279968262
61.0741424560547 -42.0437049865723
-15.5830211639404 63.6794471740723
71.0634460449219 -56.695384979248
73.7618637084961 -7.17419862747192
-19.1107044219971 -4.95230340957642
-0.728427648544312 -44.0373001098633
71.606559753418 -48.5530128479004
-66.4302673339844 -0.327835977077484
-10.7471256256104 50.0019187927246
60.6331748962402 18.3463802337646
-38.1535606384277 2.63820719718933
-5.34000825881958 -0.672251045703888
9.36342430114746 5.27910947799683
65.4282684326172 -47.4585800170898
-47.7138748168945 53.549015045166
-10.6796932220459 -21.8291873931885
-38.0355033874512 -13.532434463501
-4.31324100494385 68.4206085205078
52.3415145874023 -52.331901550293
-38.8730964660645 2.78898906707764
-36.4955520629883 -18.5189056396484
54.8506660461426 -60.6088829040527
-20.2015151977539 11.5845308303833
8.0252161026001 -49.0894393920898
75.7655410766602 -16.1678199768066
-38.9667892456055 6.01471328735352
-7.40868806838989 24.3016700744629
30.6235427856445 1.3699768781662
76.5086898803711 -22.2330055236816
-5.25026321411133 67.5455932617188
83.2030792236328 -14.3316459655762
53.0123672485352 0.478307694196701
37.3485450744629 28.9958362579346
-30.0785903930664 -10.8698148727417
28.8088111877441 -54.5211334228516
23.149715423584 8.51925659179688
-2.51547741889954 65.6581497192383
-40.2030334472656 4.37924861907959
44.0576477050781 -34.693920135498
21.7848243713379 15.1445875167847
30.1651248931885 -72.7196197509766
16.5047492980957 35.4596176147461
-3.3954746723175 9.41968059539795
48.5329055786133 9.36345767974854
-44.454963684082 56.6736793518066
-55.5926475524902 7.89623165130615
-59.2917251586914 -6.7321310043335
-4.82204008102417 -26.7455005645752
-41.9510345458984 7.41239309310913
2.96942710876465 53.9298820495605
-14.2248554229736 -6.45009613037109
14.4392776489258 9.58361721038818
79.6598129272461 -10.8771724700928
59.2922515869141 -20.6027431488037
45.9611854553223 -72.8237991333008
37.0598754882812 23.4007244110107
-26.133716583252 56.1828422546387
-15.9582977294922 -49.3759765625
20.7195434570312 -68.2566680908203
27.3982067108154 6.6661319732666
-17.0087413787842 28.5222549438477
-8.58413887023926 -2.84308671951294
-36.9404563903809 76.9203033447266
-1.85023772716522 -17.5901622772217
-0.181286469101906 49.070240020752
-51.8755683898926 69.9951248168945
-23.908576965332 68.6470642089844
-16.1966762542725 51.388256072998
-25.7606239318848 59.8825874328613
-75.3806533813477 -7.03822660446167
86.5657348632812 -45.543571472168
56.5627784729004 -48.9869766235352
21.2611827850342 -29.2159099578857
-21.3960113525391 21.9705276489258
-38.3300743103027 -4.13372278213501
-32.0654830932617 1.74141180515289
-11.2358369827271 52.9550094604492
29.9986190795898 36.7527465820312
-31.6302547454834 67.0185928344727
44.2773094177246 -37.2058410644531
-53.0618934631348 6.88460969924927
16.5096740722656 6.99396133422852
-31.2355709075928 -20.650276184082
-16.3224124908447 45.5259017944336
0.616694331169128 -21.2407989501953
58.7178115844727 -26.9473896026611
-4.42249441146851 -26.278284072876
51.587158203125 -44.8588714599609
20.5828227996826 15.6665449142456
-52.6805267333984 -9.93984222412109
-0.176125377416611 -19.4638290405273
35.3063011169434 -42.8768882751465
-21.0656833648682 -34.971809387207
20.7217578887939 -68.2426681518555
39.7439041137695 -8.8104887008667
83.4271850585938 -48.7518692016602
5.44370746612549 11.2411460876465
-29.1349754333496 -2.91787981987
76.1913681030273 -39.3777084350586
-23.6166439056396 14.7327327728271
-53.2422943115234 -3.1880099773407
51.7555885314941 -40.0069274902344
-12.6554822921753 -59.8986587524414
33.7438125610352 -17.7167797088623
68.7686157226562 -28.5229053497314
4.53547811508179 -41.2315864562988
-15.0745630264282 -64.0912933349609
-43.7637214660645 -6.25004386901855
-12.5970230102539 -3.96502327919006
-32.4800605773926 74.8199081420898
-23.8639736175537 61.5917778015137
-1.77100360393524 2.17694187164307
62.9815330505371 -18.7691802978516
23.2446002960205 -13.5396299362183
-27.1018161773682 10.5413007736206
-11.4782886505127 -16.5731792449951
-34.7837753295898 -10.016471862793
42.3681678771973 -9.34893989562988
-14.7214765548706 8.48163414001465
-55.278938293457 -8.09577178955078
-4.13879919052124 -37.4389686584473
85.3874130249023 -48.2849464416504
77.0888595581055 -8.10247421264648
-17.085521697998 -42.2612609863281
-1.34474468231201 -50.919002532959
-15.7301654815674 -6.04484176635742
-53.9591827392578 -27.2284202575684
81.4442138671875 2.06626009941101
-19.7702484130859 -12.6803245544434
14.0538921356201 1.20696699619293
49.709041595459 -5.27758264541626
-33.1269493103027 36.5951271057129
24.6680717468262 12.796591758728
31.6101036071777 30.5448226928711
-15.4466800689697 68.1862030029297
30.5778465270996 -6.49753570556641
};
\addplot [draw=color1, fill=color1, mark=*, only marks]
table{%
x  y
-14.1562461853027 16.2737731933594
15.9061107635498 0.851377665996552
-14.5903673171997 32.9087791442871
24.7583923339844 -52.4581413269043
1.76987397670746 -18.1802673339844
12.3498573303223 -70.0374374389648
-10.3571968078613 74.6969299316406
-49.0271110534668 -17.9125480651855
22.1610584259033 -49.2137107849121
36.1605529785156 -17.489049911499
-18.2265892028809 -11.663556098938
-5.55481910705566 73.7786407470703
13.1064291000366 -44.5872497558594
46.4859504699707 -35.5428199768066
14.6336479187012 -65.2039794921875
-10.5388031005859 65.3314514160156
-6.78564691543579 26.3422622680664
0.797948896884918 76.6471176147461
35.6403961181641 -61.0181159973145
28.7877178192139 -49.7694931030273
-12.8304452896118 -14.748987197876
6.28536701202393 -48.0450553894043
4.0225887298584 26.9253406524658
49.203483581543 -13.8960771560669
10.4839687347412 79.8198547363281
1.43978202342987 -64.7868728637695
9.88216018676758 44.1859855651855
-54.984489440918 -9.38840389251709
-41.3730010986328 -12.4610357284546
-23.3574848175049 18.0422458648682
33.5815658569336 16.6903343200684
45.8693923950195 -34.5644378662109
-3.67634797096252 73.5894012451172
29.9919891357422 -14.1625623703003
18.2956447601318 -43.659553527832
-23.6551246643066 17.2828235626221
-0.523545205593109 -33.7187957763672
-3.46174550056458 25.0130176544189
-13.303617477417 61.720516204834
8.45828437805176 57.3310928344727
25.74001121521 -3.36165690422058
30.3130989074707 -61.7557830810547
-9.81210994720459 -66.5528717041016
-7.22028207778931 41.8842697143555
27.6234664916992 -26.0225315093994
-23.6735897064209 23.3374004364014
25.8328628540039 -44.4100875854492
-47.7951469421387 -8.54372787475586
-9.06405830383301 -16.9985446929932
3.21282887458801 -58.4978675842285
-43.9842987060547 -6.30048370361328
2.99738717079163 23.0593242645264
15.7514410018921 10.014009475708
-42.0615005493164 15.3201522827148
-0.909237623214722 -9.83153057098389
-10.9196891784668 -22.6806621551514
-28.9560070037842 -38.3873100280762
-21.7516994476318 37.6689529418945
-0.90399307012558 -12.597393989563
-10.7759971618652 79.7270126342773
6.95509386062622 -31.4997692108154
-1.24866032600403 -48.9488410949707
1.08647096157074 -73.665901184082
31.7287845611572 -2.52046322822571
31.3026351928711 -4.13925886154175
-9.25134658813477 -46.2501754760742
38.2947578430176 -5.81643438339233
0.778681576251984 -26.6699714660645
13.3758659362793 -49.5960464477539
-2.4627513885498 -37.982551574707
18.2552795410156 -45.910270690918
36.0335083007812 -43.0681915283203
-45.2637023925781 0.117252051830292
-20.7245254516602 27.6910781860352
4.71429300308228 43.474235534668
-8.54316520690918 29.0524082183838
-10.6342191696167 62.4147338867188
-25.5382080078125 2.76588034629822
-20.9436187744141 -10.1609754562378
33.8535652160645 -8.42477703094482
-41.7106094360352 17.5954570770264
-42.6732711791992 30.4063396453857
6.28234338760376 7.22087717056274
-4.92992496490479 -56.8990135192871
-11.8856220245361 38.4550132751465
8.65184688568115 0.349969923496246
30.8050785064697 -12.5851440429688
-8.84524250030518 21.4997024536133
8.73474788665771 78.5139312744141
13.9132356643677 -56.6436729431152
15.94407081604 19.3828907012939
19.3564701080322 -4.94871473312378
-48.4773941040039 -23.1676368713379
-11.1856412887573 7.2453989982605
-1.52089750766754 -30.7742691040039
1.84971594810486 37.3751029968262
-5.7638635635376 8.2220287322998
10.7414426803589 1.55196952819824
-15.4240303039551 -29.0445556640625
22.3318481445312 27.5244808197021
37.5331916809082 -62.9953689575195
27.2599811553955 -48.8286933898926
-9.75624656677246 65.166145324707
-0.649470210075378 13.6928176879883
-28.4043292999268 20.3937892913818
39.9670448303223 27.7662830352783
19.4506244659424 -2.32283234596252
-16.0902843475342 -60.8875999450684
8.23221015930176 55.2768592834473
7.78045177459717 63.0177726745605
33.5156669616699 -37.2890853881836
24.4366474151611 13.5936975479126
30.8764934539795 -20.917142868042
45.1145668029785 20.4626750946045
25.9215526580811 -48.9129219055176
11.8244342803955 -50.2865715026855
-17.0146446228027 -19.9688949584961
11.3605842590332 -39.7386283874512
-31.621208190918 4.6064395904541
2.80386805534363 -40.207405090332
-8.1839075088501 30.9129390716553
13.5803813934326 -66.5836944580078
-17.5457954406738 24.6219291687012
-30.2811622619629 -15.9232082366943
-10.8703584671021 -2.81906247138977
-2.95088005065918 -38.0437088012695
21.6684036254883 -66.0403289794922
-16.8177680969238 -2.28541779518127
11.2980442047119 -4.48066854476929
16.5028839111328 -10.287504196167
-32.2802047729492 -13.6332426071167
-47.3634262084961 -11.6237506866455
17.3558483123779 -76.4602737426758
4.3209376335144 -12.1043319702148
-21.5901622772217 3.09332180023193
21.5516510009766 25.0099468231201
-24.203426361084 25.5471076965332
-8.62772560119629 67.5758895874023
8.06674289703369 33.1486778259277
-4.65688323974609 -1.56843042373657
6.29899215698242 67.7486572265625
-10.5629920959473 -22.7748241424561
26.409646987915 -51.4696846008301
12.7960977554321 76.7983474731445
-39.6081924438477 -1.63019394874573
2.3358690738678 -28.6222839355469
44.9914283752441 -53.0035552978516
-47.1891784667969 -3.51396441459656
-17.2614307403564 14.4307317733765
27.7461719512939 -12.9975690841675
-14.8937530517578 70.0156936645508
-19.1048164367676 15.080099105835
29.5145778656006 -40.4345779418945
-11.9497404098511 66.8091354370117
31.4417381286621 -9.81137275695801
40.2962608337402 11.2909832000732
-4.69818592071533 -16.7863788604736
25.4435920715332 -46.8182754516602
-13.724157333374 -19.9166126251221
4.88891696929932 24.9782314300537
17.5421257019043 25.3467330932617
-1.54870390892029 27.5834064483643
-35.4704093933105 -5.87784862518311
-12.174446105957 11.2807664871216
-17.2362785339355 55.619499206543
-22.6500720977783 -28.7325820922852
-38.5535430908203 12.7513599395752
15.0858554840088 -60.9970588684082
40.2745246887207 10.7555618286133
2.10408425331116 -43.9000816345215
2.43483734130859 15.0179214477539
3.69317245483398 -32.3478851318359
-12.121976852417 82.5328826904297
29.1320610046387 1.27591419219971
32.0222129821777 25.2803649902344
-46.8451690673828 4.17712068557739
36.4329147338867 -49.6340103149414
-12.5656356811523 -47.868968963623
2.3995053768158 3.609938621521
-14.246506690979 -22.5724544525146
25.0124397277832 -25.8872509002686
-0.254881918430328 31.9352989196777
-12.4840927124023 23.8146667480469
-31.6277961730957 -15.385142326355
-35.646427154541 -21.0412502288818
9.02097988128662 61.0992965698242
23.1384620666504 -61.568775177002
5.37125873565674 3.65267586708069
1.76970970630646 -8.3620548248291
-11.1585922241211 80.6489181518555
-34.6866912841797 -28.5431785583496
-2.43847513198853 -54.868049621582
-2.0417959690094 -1.06925773620605
40.1018676757812 4.54694747924805
31.3149070739746 -5.08900690078735
2.91170954704285 46.0739555358887
-16.4742374420166 19.0722599029541
4.82551193237305 -25.9199619293213
-48.4812698364258 1.47474801540375
-22.6564407348633 -5.44893312454224
-39.1985664367676 -8.78360557556152
-41.8020248413086 15.4914474487305
38.3638916015625 -26.1201820373535
-16.8518733978271 77.0881042480469
-35.7138023376465 16.9489593505859
-7.21964645385742 -45.8839645385742
34.8170280456543 -12.6864891052246
26.6888599395752 -59.169303894043
-19.6623344421387 19.8142642974854
-7.39723062515259 13.1846542358398
45.6438179016113 -4.15494441986084
0.111052051186562 -33.7903213500977
-3.46785235404968 -9.97296714782715
26.0671405792236 0.227064251899719
26.5278739929199 4.71392631530762
-7.33442544937134 81.3036270141602
-35.3015899658203 31.5906181335449
1.35896062850952 58.5727119445801
1.5297212600708 71.1305923461914
-10.010570526123 23.5384063720703
18.9875087738037 -6.82228851318359
6.80627346038818 23.0084571838379
4.11152935028076 -34.8415184020996
-12.7648687362671 -47.0462188720703
15.9349784851074 -21.9359931945801
28.6154327392578 3.10245203971863
-28.0933208465576 1.6114673614502
9.18317985534668 47.0225067138672
-12.45578956604 68.5944442749023
21.3322200775146 -70.7522659301758
30.226448059082 2.85382008552551
-0.521462440490723 -47.6239852905273
37.0535774230957 23.7878074645996
-1.67553639411926 61.4517555236816
34.4644203186035 -12.1803646087646
28.5689296722412 -22.4197235107422
-32.4310493469238 -18.4039688110352
39.0668754577637 -12.8985385894775
45.6493759155273 4.99477672576904
11.5254974365234 41.074592590332
-8.21777820587158 38.5960464477539
21.1782970428467 7.94684505462646
-15.5823822021484 20.4763259887695
-7.84456348419189 33.6721115112305
14.9224367141724 -44.2074966430664
14.3978500366211 -32.9720993041992
-43.2316627502441 -20.4248332977295
-29.3081855773926 20.3523845672607
-20.6603851318359 -3.9294376373291
-1.38323223590851 47.1030387878418
-27.3588199615479 39.5297660827637
24.0691299438477 -50.2185173034668
-45.5678291320801 -25.494592666626
-8.38327789306641 -2.32737755775452
0.616246342658997 -39.3792991638184
-6.9112434387207 35.1056709289551
36.5442810058594 18.2156276702881
-6.54022693634033 22.064790725708
-42.7417068481445 27.1025981903076
-2.0668785572052 -15.1366767883301
-32.1374626159668 -23.2694110870361
-51.4727516174316 -14.646125793457
9.67971038818359 10.4945592880249
34.4755973815918 12.4918699264526
-50.3344345092773 -4.68861103057861
19.0577411651611 13.9575262069702
11.7904024124146 -37.3484649658203
10.4059038162231 -48.5614318847656
-6.00288057327271 -8.90501689910889
34.664249420166 12.0318851470947
17.9467601776123 -66.4420394897461
21.5221843719482 -59.7803802490234
11.1702671051025 74.4532623291016
15.7115955352783 -64.0511245727539
6.62478828430176 38.1891136169434
-39.778018951416 -25.5806522369385
-39.5218887329102 14.0598936080933
1.42190909385681 -53.7553443908691
-10.9876232147217 69.3366317749023
-7.99409246444702 20.985725402832
-33.2599143981934 27.005298614502
-21.3398609161377 23.3442535400391
18.7808227539062 -75.8148574829102
-9.13871479034424 49.2513580322266
-22.5335121154785 -29.2492485046387
11.6988506317139 75.239616394043
-6.66132497787476 69.9680023193359
20.1837615966797 -72.3973007202148
4.53920030593872 63.4861526489258
27.4785194396973 -56.5249290466309
31.1390323638916 7.47712659835815
-2.16996502876282 -58.0512161254883
32.9060592651367 -49.1143951416016
-13.7541513442993 -4.27445602416992
24.6208076477051 4.66902160644531
-20.3429908752441 3.30881142616272
-29.2848739624023 -26.0461730957031
26.9002685546875 -46.6657180786133
-17.4212207794189 34.3454704284668
0.481921970844269 5.13405418395996
-6.70317602157593 16.1393127441406
-14.9225292205811 -52.051643371582
-12.7421054840088 5.92016887664795
-44.0190124511719 -1.4125919342041
5.69720506668091 56.2550811767578
-5.33053112030029 -28.6760196685791
-43.5013122558594 1.70440280437469
-16.7730884552002 -50.5043869018555
-54.7032814025879 -14.4245710372925
-16.3418769836426 25.6480083465576
-53.2781944274902 -15.987491607666
23.5342407226562 -35.0567817687988
6.20785188674927 64.8274536132812
-51.5338935852051 -10.556471824646
5.68330335617065 19.8132553100586
15.3247947692871 -3.25546836853027
-26.7605438232422 12.6324501037598
-17.1904792785645 -1.1943895816803
-5.54897451400757 63.4965934753418
35.7032318115234 12.815821647644
5.95640516281128 -67.5386276245117
-17.6672134399414 44.9177665710449
4.53430223464966 -28.5955810546875
14.6802082061768 -20.8955821990967
-14.3905334472656 11.6104049682617
9.05849838256836 74.4791107177734
-36.0064239501953 1.14881777763367
37.4585456848145 -18.2941722869873
-14.5936479568481 61.6761436462402
-20.5579738616943 -20.6917114257812
17.1725883483887 -6.50464487075806
-6.07322120666504 -24.980224609375
-3.1837592124939 13.4426889419556
6.84798860549927 -43.4136352539062
0.68505322933197 44.5727310180664
17.0947704315186 -64.4154815673828
-2.5180516242981 -64.8597412109375
8.40808582305908 -50.0519676208496
-53.5180053710938 -16.5135402679443
-9.29903316497803 51.9289207458496
23.4430599212646 -1.8377548456192
29.9761981964111 -53.7202911376953
-6.89573097229004 -49.0051002502441
17.1100425720215 -58.1905899047852
11.3685636520386 -11.402551651001
-3.10602593421936 65.9060668945312
-39.5226516723633 13.0065221786499
-1.43580591678619 69.1976776123047
-4.13788604736328 30.8474102020264
15.560097694397 -49.7316246032715
-39.9275054931641 11.8809366226196
2.13128209114075 64.247444152832
-12.0617790222168 48.4711875915527
4.42822170257568 -52.078182220459
35.5346374511719 -20.2437839508057
2.57543063163757 59.2436637878418
-39.2273483276367 13.9640045166016
-39.0242462158203 22.9863262176514
-12.306568145752 3.77273058891296
-30.7462024688721 1.61940360069275
35.7655181884766 -44.487865447998
1.52739584445953 66.1834182739258
10.0763473510742 -49.0631675720215
0.760641396045685 -60.115837097168
35.1040725708008 -53.1094245910645
4.08598947525024 84.7822799682617
36.7877464294434 -42.2259330749512
-9.05131912231445 -54.4618797302246
11.1937694549561 24.7227592468262
28.5881576538086 -40.3376121520996
19.4687156677246 -25.4935531616211
-31.2322673797607 25.4531726837158
-38.107479095459 -9.05575561523438
40.2524681091309 -55.6437835693359
-38.8869400024414 -24.7076225280762
-51.5100479125977 -23.7850437164307
14.2239561080933 -1.5298445224762
3.43313956260681 -26.457239151001
-14.7513008117676 59.8690528869629
3.2206859588623 11.7626342773438
-11.1817827224731 7.24995851516724
-13.5163869857788 34.2488327026367
-3.65907788276672 -27.7040939331055
-13.2587223052979 40.5102844238281
-25.1148815155029 -1.82188069820404
42.614875793457 -55.443660736084
-26.7906265258789 24.2013778686523
27.493932723999 -41.520191192627
0.206918761134148 -42.6391639709473
-36.6595687866211 -24.2476444244385
-3.31427526473999 -6.8505425453186
10.03773021698 -37.2256546020508
22.3766479492188 -10.2656364440918
-2.72840595245361 46.2721061706543
26.7294540405273 -30.5466651916504
38.8881492614746 -44.215404510498
14.4415721893311 -63.5295181274414
-37.0411911010742 20.3775463104248
15.9359560012817 -32.611888885498
-1.22140038013458 8.00126934051514
2.23746418952942 54.3661193847656
-32.4431686401367 5.31508731842041
-6.95743560791016 -32.4719734191895
-3.33114576339722 41.5269088745117
5.78294277191162 -27.7950344085693
-15.9254083633423 -12.0912179946899
-1.86363780498505 32.3100433349609
-18.5041275024414 -34.950080871582
-19.3557910919189 58.7318649291992
-12.6471309661865 -28.8995933532715
-29.3336219787598 -6.75573492050171
-12.5778541564941 -29.3947620391846
-0.755422353744507 21.0363597869873
-53.4739990234375 -11.9781694412231
9.53775215148926 1.18791878223419
-5.44722270965576 3.71906733512878
9.39107418060303 -62.8244514465332
25.2141914367676 -40.1923370361328
-23.5486850738525 -55.1394195556641
36.624927520752 -51.204517364502
-1.41397559642792 64.4651794433594
31.1220169067383 32.1042709350586
28.4181709289551 -61.9741172790527
41.0682106018066 -6.71240520477295
7.91067838668823 39.9108734130859
-1.29828250408173 -3.32939910888672
-9.48875617980957 72.2016906738281
-10.4276132583618 38.6326713562012
-49.5756340026855 -17.0155220031738
-1.23361301422119 85.5276947021484
-9.40144634246826 62.9727249145508
1.43499612808228 59.0431213378906
-6.21810626983643 53.1744232177734
-40.3940620422363 29.7566719055176
25.2305107116699 -40.0765686035156
14.0889110565186 -52.4261283874512
-4.53575658798218 25.081392288208
-13.8788919448853 23.9843482971191
-36.7962684631348 9.24637508392334
15.9282217025757 31.1022338867188
11.8129873275757 73.0501480102539
14.2319955825806 -63.4841346740723
-1.97849798202515 66.7942733764648
-23.7449798583984 -56.3668365478516
-48.5560188293457 7.43255281448364
-20.9436798095703 -41.8449440002441
-14.2904224395752 23.3435115814209
10.4972915649414 49.6534957885742
-3.99147343635559 -50.0040283203125
18.211742401123 -51.417423248291
-5.88881301879883 -8.80561828613281
-1.51741051673889 -47.4201202392578
4.32817792892456 31.6414756774902
-24.871789932251 -7.62801742553711
-20.4118881225586 45.7798843383789
27.4483509063721 -26.8081016540527
-5.56537246704102 -29.4588851928711
-16.0065059661865 -65.9480972290039
-5.21497488021851 -25.0645942687988
40.4045066833496 -20.6414604187012
-45.9770278930664 22.8458423614502
-26.5258178710938 -0.407515108585358
28.5873775482178 -46.7295265197754
-44.2382164001465 10.3070650100708
-50.5521125793457 5.60726404190063
-7.19392871856689 -42.993709564209
-9.5423583984375 -38.5154151916504
-43.3997917175293 -7.75084924697876
9.01693916320801 -51.2871055603027
25.0602397918701 13.2396965026855
-29.0721530914307 11.8301572799683
-27.9322357177734 -3.03503251075745
1.96551692485809 36.5476722717285
-7.27517414093018 82.8970489501953
0.202581092715263 59.3200950622559
33.6214828491211 4.04944849014282
34.3057708740234 -22.4389762878418
-23.6837158203125 -9.76546096801758
-34.6332054138184 0.0546604692935944
39.3979530334473 -44.4877548217773
-22.6435832977295 7.02360010147095
29.4733123779297 -25.9379234313965
-33.2400817871094 5.08824300765991
-38.6765251159668 4.40700483322144
9.50541591644287 18.0882816314697
10.8913688659668 -52.0722389221191
32.0797653198242 -33.75390625
-6.8299732208252 1.5664496421814
27.8091373443604 -42.6121864318848
7.99870491027832 47.2746238708496
-47.7646560668945 19.1078987121582
-0.175539493560791 -6.24241399765015
-15.9347543716431 -23.7865829467773
3.07636833190918 -4.72971487045288
37.8014335632324 -23.3688488006592
18.0060062408447 42.4501457214355
13.2788887023926 11.630054473877
-7.13648319244385 -29.9890613555908
0.0621891058981419 87.6787338256836
7.98978853225708 -35.625072479248
};
\end{axis}

%% file: pgf/emb3.tex

\definecolor{color0}{rgb}{0.12156862745098,0.466666666666667,0.705882352941177}
\definecolor{color1}{rgb}{1,0.498039215686275,0.0549019607843137}

\begin{axis}[name=emb3,at={($(emb2.east)+(1.5cm,0)$)},anchor=west,
tick align=outside,
tick pos=left,
title={Iteration 3},
x grid style={white!69.0196078431373!black},
xmin=-100, xmax=100,
xtick style={color=black},
y grid style={white!69.0196078431373!black},
ymin=-100, ymax=100,
ytick style={color=black},
xlabel={embedding $1$},
]
\addplot [draw=color0, fill=color0, mark=*, only marks]
table{%
x  y
-38.1394691467285 31.613826751709
-53.6026954650879 -14.2612895965576
-29.946475982666 33.599796295166
43.2039375305176 -66.5453186035156
12.5446929931641 -21.9658164978027
68.8552169799805 -12.7214450836182
-12.6025838851929 57.0704498291016
-53.712158203125 50.2754211425781
-13.3913583755493 -71.1761016845703
50.3764343261719 -12.2465028762817
-37.2645683288574 24.069860458374
-12.7079315185547 85.753532409668
48.7055549621582 -52.2421340942383
33.4609756469727 -74.3191986083984
-0.855190992355347 -78.4004135131836
-25.1809043884277 76.975456237793
-3.43423438072205 -27.8494071960449
-8.61722278594971 90.1910400390625
40.9777412414551 -25.7496242523193
26.5266361236572 -64.3477401733398
-39.0742073059082 7.02745676040649
33.7514724731445 -60.0735054016113
2.4723813533783 44.8706893920898
-15.5153427124023 -38.6645202636719
-4.93873691558838 64.718864440918
67.6318740844727 -52.4998588562012
-6.92963600158691 52.3849296569824
-48.0399360656738 39.1386222839355
-45.9315071105957 18.8996658325195
-35.0982666015625 17.4722232818604
-65.8723220825195 -7.31351327896118
-19.0517826080322 -25.0586223602295
-3.76817226409912 72.2202682495117
6.28907632827759 -18.4101486206055
42.3331336975098 -36.8180198669434
-40.2684173583984 23.7639465332031
45.2283477783203 -3.98420166969299
-25.8336925506592 37.1723289489746
-27.4597282409668 96.5063934326172
-18.9838085174561 79.5008850097656
-37.7121696472168 9.63839530944824
48.1750068664551 -81.0767593383789
-9.77233982086182 -37.4815979003906
-3.50150156021118 -16.470817565918
10.0994081497192 -24.2324275970459
-37.8442115783691 15.879864692688
58.307258605957 -35.4715576171875
-54.6654815673828 43.1385650634766
37.2571792602539 -37.181095123291
31.350004196167 -28.7315216064453
-28.4784984588623 -19.2070274353027
13.7431869506836 4.42524766921997
-44.5649108886719 -19.0002880096436
-9.99794578552246 17.1870651245117
19.9524269104004 42.3049697875977
14.1887044906616 -55.5313758850098
38.196533203125 -27.6626472473145
15.9178924560547 -21.5863933563232
26.4974422454834 35.4914016723633
-30.9731216430664 93.8189315795898
42.5893630981445 -20.1126232147217
43.778263092041 -46.2909164428711
22.1556911468506 -52.0119590759277
2.38426494598389 7.5336799621582
-24.7656936645508 -20.6058464050293
35.3400192260742 -23.9489898681641
66.6178741455078 -32.7973365783691
43.3757247924805 -43.657772064209
64.9415512084961 -20.6347427368164
24.0774478912354 -16.0055561065674
26.2596035003662 -66.0740127563477
8.24983596801758 -57.0617904663086
41.0042686462402 31.988109588623
-37.0005149841309 30.9146671295166
7.52376222610474 71.2180938720703
-21.2676887512207 1.80480265617371
6.43966484069824 71.8707733154297
-33.6614112854004 37.2737045288086
4.03257322311401 11.8498697280884
1.37358283996582 -63.8693046569824
-20.0376815795898 -12.9888000488281
-31.7740745544434 2.048579454422
21.9119434356689 -26.4639911651611
25.125280380249 -69.0963897705078
24.9211597442627 -2.26449275016785
-33.5107879638672 -33.891918182373
-5.59192609786987 -8.7601432800293
-4.58844900131226 83.5942153930664
-11.1740865707397 81.3536605834961
40.0559425354004 -36.861255645752
1.97088754177094 -61.903938293457
-7.61819124221802 -12.6712026596069
34.7032203674316 7.66340351104736
-44.825325012207 24.6952781677246
44.327766418457 28.7206268310547
-35.0405807495117 3.41243886947632
7.31040096282959 42.1818580627441
-39.188907623291 -29.8076305389404
47.9076232910156 -35.7289810180664
-18.1970710754395 27.9760723114014
9.13891315460205 -58.6497383117676
42.3337173461914 -70.6091842651367
-23.8636131286621 96.0698471069336
16.2560710906982 -29.3453712463379
-23.8527126312256 8.11795520782471
-68.3238754272461 -9.06933689117432
51.321891784668 -70.819938659668
-2.28497552871704 -55.461669921875
-8.83580303192139 61.7178153991699
-0.449237138032913 71.9764404296875
17.2892475128174 -83.9834899902344
1.39302980899811 18.2058410644531
58.1275978088379 14.7701826095581
-46.8673133850098 -21.4403953552246
42.1339302062988 -62.4418449401855
46.2196388244629 -59.5730514526367
30.4819927215576 -13.4465007781982
-13.6441688537598 -20.2770881652832
-38.5393447875977 35.7407188415527
45.977222442627 -25.5616397857666
-25.9071807861328 19.2602024078369
60.039867401123 -29.1267070770264
-14.1425113677979 -1.37825894355774
30.5405025482178 4.71726894378662
-19.9467449188232 -18.3487854003906
41.9758338928223 13.02659034729
38.3454284667969 -53.7497062683105
-41.213752746582 27.1752910614014
-12.913703918457 8.38034248352051
16.8278999328613 -13.512975692749
28.5461044311523 6.15626287460327
10.1599054336548 27.4758434295654
58.8793106079102 -40.6910820007324
54.3530807495117 6.58582592010498
36.0898399353027 32.0747451782227
4.17863178253174 26.5792655944824
9.28135585784912 43.4979972839355
-12.1076326370239 60.2258071899414
-25.6478233337402 10.3897533416748
-47.7434883117676 31.372386932373
-18.2250442504883 84.034782409668
18.3378868103027 -15.7327976226807
-5.95386457443237 -56.4809494018555
-26.4644298553467 74.5657730102539
-10.4727745056152 14.1857204437256
29.3909492492676 4.70043802261353
31.0635375976562 -44.7365684509277
1.66425395011902 -32.8970947265625
25.1708297729492 25.7900638580322
62.6711959838867 -26.6533088684082
-20.0307960510254 86.5952758789062
17.8175258636475 -12.5754108428955
44.3590393066406 9.1757640838623
-26.5661659240723 57.6567611694336
-27.0862407684326 -53.4224815368652
0.294786274433136 -18.4522399902344
-1.58074474334717 -68.6221160888672
33.7198486328125 -43.891170501709
-27.0496826171875 -53.4135971069336
5.92521858215332 52.7913055419922
13.6704082489014 53.3023452758789
-65.5870361328125 32.112419128418
-45.2146263122559 18.1487731933594
-1.59105050563812 19.1618003845215
-13.799015045166 66.6532745361328
-37.8113594055176 35.3197250366211
-14.3940544128418 12.8407535552979
22.0605392456055 -43.6653633117676
28.1170082092285 31.1274700164795
-24.824275970459 -63.451286315918
-7.68308210372925 11.8374633789062
4.19253158569336 -17.8641319274902
-27.073766708374 89.4548645019531
46.8346366882324 -18.5323944091797
-63.8396301269531 -5.50583505630493
-27.3914089202881 29.3708534240723
12.2721738815308 -82.489616394043
-4.53950977325439 -5.51422643661499
-47.5697822570801 -14.7177829742432
38.0959167480469 -34.8125228881836
28.5089092254639 -25.4980506896973
1.71449613571167 15.714542388916
-55.0056686401367 28.6125793457031
8.23566818237305 35.8725814819336
-11.1053514480591 3.4276020526886
-10.0473222732544 83.3941802978516
-5.15028619766235 -42.4866752624512
20.0696201324463 26.9091491699219
-60.161693572998 5.21272563934326
-32.4673957824707 93.0444717407227
12.7834510803223 76.5851058959961
45.5559539794922 -70.8861846923828
19.9395809173584 15.0962162017822
30.1063022613525 -0.00187792582437396
17.0699348449707 -73.3328247070312
-35.2190361022949 65.5659408569336
-11.4004726409912 20.4762744903564
-29.0197792053223 -23.83544921875
-32.1931381225586 25.7250785827637
38.7032623291016 -62.955810546875
-42.7379302978516 14.6658849716187
-21.9184112548828 9.08452701568604
38.7588768005371 -37.2719802856445
-13.1447925567627 84.4067306518555
-49.4044799804688 30.0773124694824
49.5279655456543 5.44973516464233
26.4664134979248 27.7290306091309
21.0434226989746 -71.137092590332
12.0383243560791 36.307201385498
-6.23784780502319 32.3255348205566
3.37404918670654 -12.5146999359131
25.2211894989014 -32.8194580078125
22.953987121582 0.493030250072479
30.2494583129883 -28.0496368408203
43.7219886779785 -73.1155700683594
-32.2084426879883 99.3870239257812
-19.5050182342529 31.1167736053467
-10.4076976776123 67.3212890625
4.89601850509644 74.6052474975586
-44.8527412414551 -35.122428894043
0.546008169651031 -22.8661727905273
2.47039437294006 52.4669876098633
19.3886375427246 -21.244104385376
-14.0611267089844 -19.1557483673096
8.10796642303467 -6.23268795013428
-1.05648612976074 -24.1120510101318
14.4251794815063 48.0271148681641
50.2996215820312 27.5838146209717
-29.659761428833 100.271675109863
45.2347869873047 -24.9034805297852
-15.594407081604 -9.87313842773438
63.54931640625 -43.0539054870605
-66.1623611450195 -8.30381011962891
-22.0107803344727 69.4758911132812
8.23962879180908 -19.3886470794678
42.8504829406738 1.49837386608124
4.44296932220459 16.5928535461426
31.8377990722656 3.29185390472412
45.3569374084473 -52.5974617004395
10.7846965789795 18.4056739807129
-12.0877342224121 70.2258529663086
39.3390426635742 -8.19195938110352
35.2952690124512 13.6486787796021
-31.3393173217773 36.8481674194336
17.8032932281494 -54.2725067138672
19.4336528778076 -34.5730895996094
-38.9771537780762 8.76212596893311
12.3397207260132 50.1087341308594
-52.9529457092285 8.00139999389648
-40.8578948974609 77.552864074707
-5.50185680389404 10.1921539306641
11.0463771820068 -75.0285797119141
24.8085994720459 -45.1326179504395
-14.2496957778931 -18.8525314331055
45.0313987731934 -62.1998519897461
-19.7280101776123 67.7234191894531
9.85436248779297 3.7086079120636
-20.7725849151611 -6.93229293823242
-46.6237030029297 32.465202331543
25.0375213623047 45.0499267578125
13.4750423431396 30.957498550415
-50.5957374572754 49.9657936096191
-52.6459655761719 -7.80616569519043
-51.4101104736328 -23.8153686523438
-56.3958168029785 -3.2066171169281
57.8110885620117 -60.891716003418
10.1035051345825 -44.4147109985352
-8.42291355133057 -8.11262607574463
18.3250331878662 -3.58542823791504
-6.49575710296631 -11.6973361968994
74.0573425292969 -40.7488288879395
-6.53106451034546 -42.1663436889648
-22.363000869751 75.9916305541992
64.2273788452148 -52.0037231445312
5.84508609771729 5.3161678314209
26.2159309387207 42.6140556335449
-49.5345115661621 -6.12801218032837
40.661376953125 -9.74051666259766
-24.3730621337891 97.4279022216797
-59.1583290100098 14.2267484664917
-4.06825256347656 7.41548681259155
26.6340370178223 -10.9292545318604
11.9424705505371 -40.6544876098633
-17.780065536499 65.0289535522461
-32.4643363952637 -4.12269926071167
-25.9056491851807 72.1544418334961
6.92263793945312 79.1510391235352
49.4272193908691 -28.2797622680664
-34.4816589355469 75.2138061523438
71.8999176025391 -24.5428867340088
-16.0472507476807 -67.2366180419922
44.1531105041504 -58.4207305908203
54.3291816711426 -34.4892616271973
-43.436653137207 -9.28324317932129
61.328784942627 -32.7675819396973
5.13684034347534 21.8402271270752
38.7042999267578 21.2779865264893
54.6459312438965 -67.0537338256836
-2.41563391685486 6.84422016143799
-21.3201274871826 7.70546245574951
-33.0243072509766 -22.8013172149658
61.1666297912598 -59.6950988769531
2.48933815956116 7.42455339431763
-47.0866394042969 -17.8100681304932
-31.6077995300293 73.1261672973633
24.8690338134766 -45.1874465942383
19.2101383209229 13.7217378616333
41.5094795227051 -55.3200798034668
-52.4504585266113 44.1950378417969
-2.12331056594849 -2.18581628799438
-57.3916893005371 48.7923316955566
59.2741279602051 -52.2526206970215
-13.6992664337158 84.401611328125
-54.495491027832 39.4044494628906
-32.0199699401855 63.7198753356934
66.4733276367188 -26.3631954193115
40.8371734619141 -40.5921211242676
12.2283792495728 35.4713935852051
-15.7243003845215 74.8287963867188
-61.3050270080566 -5.57661294937134
21.5492534637451 -69.9052124023438
-38.6072616577148 -6.27425765991211
42.6711807250977 -22.38671875
18.3546333312988 -1.14922869205475
-44.7204132080078 -17.8332176208496
7.5306601524353 73.1889038085938
-65.15673828125 2.07821869850159
3.07175540924072 -56.9403533935547
-12.7544460296631 89.5094528198242
18.0809230804443 12.0003776550293
-17.6186809539795 -22.8658409118652
8.15689849853516 -6.10607194900513
27.2385501861572 -0.26924654841423
69.5884399414062 -16.3200302124023
-5.67017841339111 71.0647048950195
45.0351715087891 -20.8775730133057
-9.96841621398926 -67.7342987060547
5.87092113494873 -77.4882965087891
-48.7112998962402 48.769962310791
-42.0888366699219 67.2918395996094
-38.8545608520508 -7.24172115325928
40.877815246582 -33.1480445861816
8.8345193862915 -31.8501129150391
58.3881340026855 -78.823371887207
4.09927225112915 -9.52323818206787
-15.212441444397 77.7001800537109
-45.0432014465332 -0.293186783790588
-30.497184753418 57.0464019775391
-15.0403175354004 -7.9351749420166
-8.27572250366211 -46.9445266723633
-43.5069847106934 2.26918578147888
-31.6455707550049 84.1911010742188
39.8779754638672 42.6427383422852
5.87359380722046 -50.168701171875
4.82944965362549 -32.5957527160645
-3.38232970237732 58.7161750793457
-39.1681518554688 8.289870262146
-13.0693683624268 15.5048398971558
-9.72932052612305 44.0320854187012
16.9627475738525 39.0360374450684
42.3536338806152 -70.62841796875
-11.1209278106689 58.9555511474609
3.07072257995605 -56.9399490356445
-14.2981376647949 -53.9906845092773
45.0191116333008 -62.1866455078125
1.80027556419373 63.9484596252441
44.5248260498047 -81.9046096801758
70.6742782592773 -36.5772514343262
-13.7155637741089 28.8691425323486
1.49007666110992 -14.4703788757324
41.9619140625 -78.0130310058594
-53.1061820983887 15.3728923797607
3.06283020973206 59.7040519714355
-24.8203468322754 -63.4504508972168
26.7129383087158 25.1772556304932
-28.5505561828613 6.84880828857422
-0.232716098427773 -21.0809326171875
37.363655090332 -74.4818649291992
2.05301260948181 79.9070739746094
-0.206202208995819 -5.97018623352051
28.7667007446289 38.0559692382812
12.0599460601807 59.102180480957
12.6312551498413 -55.9592742919922
27.1617183685303 24.5364856719971
-29.0958271026611 -0.213039562106133
31.9514236450195 -53.6208801269531
-14.1672925949097 2.63664603233337
7.76836967468262 -42.5276565551758
16.2425079345703 -66.1114501953125
12.9869813919067 25.8958435058594
0.877788424491882 25.3167819976807
-11.4175853729248 -34.5107116699219
59.113166809082 -56.0027618408203
11.3286457061768 59.9462203979492
65.7841949462891 -59.2986793518066
-17.5019264221191 -49.2155685424805
-16.786771774292 -42.9056625366211
-29.7154579162598 42.1907577514648
17.3424854278564 -53.596435546875
-5.56077194213867 4.65688705444336
13.2926635742188 61.9591979980469
0.470447510480881 31.4481430053711
20.6972923278809 -83.3771286010742
-7.63357162475586 6.73862838745117
17.5468044281006 -41.3467483520508
34.4164733886719 39.2596588134766
-28.8807392120361 -3.94574332237244
-27.1128635406494 -53.3744583129883
-9.46680068969727 75.7124099731445
-54.6072120666504 13.985445022583
-40.0881538391113 14.9287281036377
-1.17712998390198 -12.2241201400757
14.5106477737427 28.6589984893799
-54.3072128295898 42.1922264099121
6.40880393981934 -12.6565437316895
-4.87662124633789 -28.8560810089111
70.3321304321289 -56.743480682373
28.0657577514648 -67.945426940918
43.3446922302246 -66.5370330810547
-6.54024791717529 -37.0669403076172
-9.77907180786133 72.471061706543
46.8657302856445 20.6045360565186
4.7654128074646 -54.7908592224121
6.22988653182983 -18.4198818206787
-4.29327583312988 50.7126045227051
-5.7031192779541 -1.02196419239044
-26.6669654846191 85.2631454467773
-10.9744749069214 -15.4197626113892
-53.7070045471191 46.8463363647461
-20.2498741149902 95.3013000488281
-0.0186635926365852 81.1195068359375
-1.57753801345825 64.4960174560547
-16.5441207885742 69.3137283325195
-44.490550994873 14.7975206375122
5.32721996307373 -80.4375152587891
36.8004150390625 -70.1955490112305
22.6455154418945 -27.0034675598145
-3.09143900871277 12.9956531524658
-16.4380340576172 16.5790214538574
-24.9045295715332 19.3737888336182
4.76603984832764 61.9727973937988
-1.84666478633881 -75.4796752929688
-19.9632911682129 74.706428527832
23.9617595672607 -83.5851058959961
-44.7421226501465 20.2052726745605
2.5480842590332 -16.1237907409668
22.6331157684326 28.4964466094971
-7.39057922363281 17.1661987304688
-7.66574192047119 -17.1754703521729
36.6205978393555 -42.1029891967773
-0.562049984931946 -11.4344415664673
50.3992080688477 -44.7954635620117
-6.17042970657349 6.48294305801392
-43.7702789306641 28.1427001953125
-9.24235916137695 -17.0689868927002
19.7906303405762 -47.0327835083008
-34.7844161987305 -6.93713998794556
4.75787162780762 -54.7947654724121
55.7065658569336 -21.3418102264404
5.31485557556152 -75.4134521484375
1.23454737663269 27.8672657012939
-26.5256328582764 3.07332587242126
58.2487373352051 -43.5976066589355
-15.7939767837524 0.4960897564888
-46.4825439453125 12.22190284729
38.688533782959 -66.2474212646484
26.926399230957 -28.6456241607666
41.6075172424316 -17.0335750579834
24.5791969299316 -72.6562194824219
3.50666761398315 -36.807445526123
40.107837677002 42.6313400268555
-42.4969367980957 1.15924549102783
30.5738162994385 -7.54047250747681
-28.5729141235352 94.7233581542969
-14.3268375396729 71.5219879150391
-32.0162506103516 -14.1073999404907
31.9764556884766 -61.8346138000488
-1.3437591791153 -31.0363140106201
0.728518486022949 15.9316301345825
50.7395324707031 -16.1572265625
-40.8276557922363 7.55084705352783
46.742748260498 -12.1976375579834
-17.8813190460205 -0.137362480163574
-28.2801761627197 22.536922454834
30.6036033630371 13.8386344909668
5.91314125061035 -77.46484375
72.8910522460938 -32.7019882202148
11.4678926467896 44.8068313598633
24.6549530029297 -32.7375640869141
13.9002208709717 4.44688129425049
-53.6840133666992 -10.6989717483521
-4.12124395370483 -81.6131134033203
-29.5052967071533 -12.4665374755859
-14.85276222229 -8.3715877532959
34.2997589111328 -12.0776596069336
-22.7897701263428 50.4521560668945
-2.79966449737549 6.88044023513794
-11.0801057815552 3.41671252250671
-29.6593284606934 71.5669784545898
-9.83220863342285 -42.3832473754883
};
\addplot [draw=color1, fill=color1, mark=*, only marks]
table{%
x  y
-14.1562461853027 16.2737731933594
15.9061107635498 0.851377665996552
-14.5903673171997 32.9087791442871
24.7583923339844 -52.4581413269043
1.76987397670746 -18.1802673339844
12.3498573303223 -70.0374374389648
-10.3571968078613 74.6969299316406
-49.0271110534668 -17.9125480651855
22.1610584259033 -49.2137107849121
36.1605529785156 -17.489049911499
-18.2265892028809 -11.663556098938
-5.55481910705566 73.7786407470703
13.1064291000366 -44.5872497558594
46.4859504699707 -35.5428199768066
14.6336479187012 -65.2039794921875
-10.5388031005859 65.3314514160156
-6.78564691543579 26.3422622680664
0.797948896884918 76.6471176147461
35.6403961181641 -61.0181159973145
28.7877178192139 -49.7694931030273
-12.8304452896118 -14.748987197876
6.28536701202393 -48.0450553894043
4.0225887298584 26.9253406524658
49.203483581543 -13.8960771560669
10.4839687347412 79.8198547363281
1.43978202342987 -64.7868728637695
9.88216018676758 44.1859855651855
-54.984489440918 -9.38840389251709
-41.3730010986328 -12.4610357284546
-23.3574848175049 18.0422458648682
33.5815658569336 16.6903343200684
45.8693923950195 -34.5644378662109
-3.67634797096252 73.5894012451172
29.9919891357422 -14.1625623703003
18.2956447601318 -43.659553527832
-23.6551246643066 17.2828235626221
-0.523545205593109 -33.7187957763672
-3.46174550056458 25.0130176544189
-13.303617477417 61.720516204834
8.45828437805176 57.3310928344727
25.74001121521 -3.36165690422058
30.3130989074707 -61.7557830810547
-9.81210994720459 -66.5528717041016
-7.22028207778931 41.8842697143555
27.6234664916992 -26.0225315093994
-23.6735897064209 23.3374004364014
25.8328628540039 -44.4100875854492
-47.7951469421387 -8.54372787475586
-9.06405830383301 -16.9985446929932
3.21282887458801 -58.4978675842285
-43.9842987060547 -6.30048370361328
2.99738717079163 23.0593242645264
15.7514410018921 10.014009475708
-42.0615005493164 15.3201522827148
-0.909237623214722 -9.83153057098389
-10.9196891784668 -22.6806621551514
-28.9560070037842 -38.3873100280762
-21.7516994476318 37.6689529418945
-0.90399307012558 -12.597393989563
-10.7759971618652 79.7270126342773
6.95509386062622 -31.4997692108154
-1.24866032600403 -48.9488410949707
1.08647096157074 -73.665901184082
31.7287845611572 -2.52046322822571
31.3026351928711 -4.13925886154175
-9.25134658813477 -46.2501754760742
38.2947578430176 -5.81643438339233
0.778681576251984 -26.6699714660645
13.3758659362793 -49.5960464477539
-2.4627513885498 -37.982551574707
18.2552795410156 -45.910270690918
36.0335083007812 -43.0681915283203
-45.2637023925781 0.117252051830292
-20.7245254516602 27.6910781860352
4.71429300308228 43.474235534668
-8.54316520690918 29.0524082183838
-10.6342191696167 62.4147338867188
-25.5382080078125 2.76588034629822
-20.9436187744141 -10.1609754562378
33.8535652160645 -8.42477703094482
-41.7106094360352 17.5954570770264
-42.6732711791992 30.4063396453857
6.28234338760376 7.22087717056274
-4.92992496490479 -56.8990135192871
-11.8856220245361 38.4550132751465
8.65184688568115 0.349969923496246
30.8050785064697 -12.5851440429688
-8.84524250030518 21.4997024536133
8.73474788665771 78.5139312744141
13.9132356643677 -56.6436729431152
15.94407081604 19.3828907012939
19.3564701080322 -4.94871473312378
-48.4773941040039 -23.1676368713379
-11.1856412887573 7.2453989982605
-1.52089750766754 -30.7742691040039
1.84971594810486 37.3751029968262
-5.7638635635376 8.2220287322998
10.7414426803589 1.55196952819824
-15.4240303039551 -29.0445556640625
22.3318481445312 27.5244808197021
37.5331916809082 -62.9953689575195
27.2599811553955 -48.8286933898926
-9.75624656677246 65.166145324707
-0.649470210075378 13.6928176879883
-28.4043292999268 20.3937892913818
39.9670448303223 27.7662830352783
19.4506244659424 -2.32283234596252
-16.0902843475342 -60.8875999450684
8.23221015930176 55.2768592834473
7.78045177459717 63.0177726745605
33.5156669616699 -37.2890853881836
24.4366474151611 13.5936975479126
30.8764934539795 -20.917142868042
45.1145668029785 20.4626750946045
25.9215526580811 -48.9129219055176
11.8244342803955 -50.2865715026855
-17.0146446228027 -19.9688949584961
11.3605842590332 -39.7386283874512
-31.621208190918 4.6064395904541
2.80386805534363 -40.207405090332
-8.1839075088501 30.9129390716553
13.5803813934326 -66.5836944580078
-17.5457954406738 24.6219291687012
-30.2811622619629 -15.9232082366943
-10.8703584671021 -2.81906247138977
-2.95088005065918 -38.0437088012695
21.6684036254883 -66.0403289794922
-16.8177680969238 -2.28541779518127
11.2980442047119 -4.48066854476929
16.5028839111328 -10.287504196167
-32.2802047729492 -13.6332426071167
-47.3634262084961 -11.6237506866455
17.3558483123779 -76.4602737426758
4.3209376335144 -12.1043319702148
-21.5901622772217 3.09332180023193
21.5516510009766 25.0099468231201
-24.203426361084 25.5471076965332
-8.62772560119629 67.5758895874023
8.06674289703369 33.1486778259277
-4.65688323974609 -1.56843042373657
6.29899215698242 67.7486572265625
-10.5629920959473 -22.7748241424561
26.409646987915 -51.4696846008301
12.7960977554321 76.7983474731445
-39.6081924438477 -1.63019394874573
2.3358690738678 -28.6222839355469
44.9914283752441 -53.0035552978516
-47.1891784667969 -3.51396441459656
-17.2614307403564 14.4307317733765
27.7461719512939 -12.9975690841675
-14.8937530517578 70.0156936645508
-19.1048164367676 15.080099105835
29.5145778656006 -40.4345779418945
-11.9497404098511 66.8091354370117
31.4417381286621 -9.81137275695801
40.2962608337402 11.2909832000732
-4.69818592071533 -16.7863788604736
25.4435920715332 -46.8182754516602
-13.724157333374 -19.9166126251221
4.88891696929932 24.9782314300537
17.5421257019043 25.3467330932617
-1.54870390892029 27.5834064483643
-35.4704093933105 -5.87784862518311
-12.174446105957 11.2807664871216
-17.2362785339355 55.619499206543
-22.6500720977783 -28.7325820922852
-38.5535430908203 12.7513599395752
15.0858554840088 -60.9970588684082
40.2745246887207 10.7555618286133
2.10408425331116 -43.9000816345215
2.43483734130859 15.0179214477539
3.69317245483398 -32.3478851318359
-12.121976852417 82.5328826904297
29.1320610046387 1.27591419219971
32.0222129821777 25.2803649902344
-46.8451690673828 4.17712068557739
36.4329147338867 -49.6340103149414
-12.5656356811523 -47.868968963623
2.3995053768158 3.609938621521
-14.246506690979 -22.5724544525146
25.0124397277832 -25.8872509002686
-0.254881918430328 31.9352989196777
-12.4840927124023 23.8146667480469
-31.6277961730957 -15.385142326355
-35.646427154541 -21.0412502288818
9.02097988128662 61.0992965698242
23.1384620666504 -61.568775177002
5.37125873565674 3.65267586708069
1.76970970630646 -8.3620548248291
-11.1585922241211 80.6489181518555
-34.6866912841797 -28.5431785583496
-2.43847513198853 -54.868049621582
-2.0417959690094 -1.06925773620605
40.1018676757812 4.54694747924805
31.3149070739746 -5.08900690078735
2.91170954704285 46.0739555358887
-16.4742374420166 19.0722599029541
4.82551193237305 -25.9199619293213
-48.4812698364258 1.47474801540375
-22.6564407348633 -5.44893312454224
-39.1985664367676 -8.78360557556152
-41.8020248413086 15.4914474487305
38.3638916015625 -26.1201820373535
-16.8518733978271 77.0881042480469
-35.7138023376465 16.9489593505859
-7.21964645385742 -45.8839645385742
34.8170280456543 -12.6864891052246
26.6888599395752 -59.169303894043
-19.6623344421387 19.8142642974854
-7.39723062515259 13.1846542358398
45.6438179016113 -4.15494441986084
0.111052051186562 -33.7903213500977
-3.46785235404968 -9.97296714782715
26.0671405792236 0.227064251899719
26.5278739929199 4.71392631530762
-7.33442544937134 81.3036270141602
-35.3015899658203 31.5906181335449
1.35896062850952 58.5727119445801
1.5297212600708 71.1305923461914
-10.010570526123 23.5384063720703
18.9875087738037 -6.82228851318359
6.80627346038818 23.0084571838379
4.11152935028076 -34.8415184020996
-12.7648687362671 -47.0462188720703
15.9349784851074 -21.9359931945801
28.6154327392578 3.10245203971863
-28.0933208465576 1.6114673614502
9.18317985534668 47.0225067138672
-12.45578956604 68.5944442749023
21.3322200775146 -70.7522659301758
30.226448059082 2.85382008552551
-0.521462440490723 -47.6239852905273
37.0535774230957 23.7878074645996
-1.67553639411926 61.4517555236816
34.4644203186035 -12.1803646087646
28.5689296722412 -22.4197235107422
-32.4310493469238 -18.4039688110352
39.0668754577637 -12.8985385894775
45.6493759155273 4.99477672576904
11.5254974365234 41.074592590332
-8.21777820587158 38.5960464477539
21.1782970428467 7.94684505462646
-15.5823822021484 20.4763259887695
-7.84456348419189 33.6721115112305
14.9224367141724 -44.2074966430664
14.3978500366211 -32.9720993041992
-43.2316627502441 -20.4248332977295
-29.3081855773926 20.3523845672607
-20.6603851318359 -3.9294376373291
-1.38323223590851 47.1030387878418
-27.3588199615479 39.5297660827637
24.0691299438477 -50.2185173034668
-45.5678291320801 -25.494592666626
-8.38327789306641 -2.32737755775452
0.616246342658997 -39.3792991638184
-6.9112434387207 35.1056709289551
36.5442810058594 18.2156276702881
-6.54022693634033 22.064790725708
-42.7417068481445 27.1025981903076
-2.0668785572052 -15.1366767883301
-32.1374626159668 -23.2694110870361
-51.4727516174316 -14.646125793457
9.67971038818359 10.4945592880249
34.4755973815918 12.4918699264526
-50.3344345092773 -4.68861103057861
19.0577411651611 13.9575262069702
11.7904024124146 -37.3484649658203
10.4059038162231 -48.5614318847656
-6.00288057327271 -8.90501689910889
34.664249420166 12.0318851470947
17.9467601776123 -66.4420394897461
21.5221843719482 -59.7803802490234
11.1702671051025 74.4532623291016
15.7115955352783 -64.0511245727539
6.62478828430176 38.1891136169434
-39.778018951416 -25.5806522369385
-39.5218887329102 14.0598936080933
1.42190909385681 -53.7553443908691
-10.9876232147217 69.3366317749023
-7.99409246444702 20.985725402832
-33.2599143981934 27.005298614502
-21.3398609161377 23.3442535400391
18.7808227539062 -75.8148574829102
-9.13871479034424 49.2513580322266
-22.5335121154785 -29.2492485046387
11.6988506317139 75.239616394043
-6.66132497787476 69.9680023193359
20.1837615966797 -72.3973007202148
4.53920030593872 63.4861526489258
27.4785194396973 -56.5249290466309
31.1390323638916 7.47712659835815
-2.16996502876282 -58.0512161254883
32.9060592651367 -49.1143951416016
-13.7541513442993 -4.27445602416992
24.6208076477051 4.66902160644531
-20.3429908752441 3.30881142616272
-29.2848739624023 -26.0461730957031
26.9002685546875 -46.6657180786133
-17.4212207794189 34.3454704284668
0.481921970844269 5.13405418395996
-6.70317602157593 16.1393127441406
-14.9225292205811 -52.051643371582
-12.7421054840088 5.92016887664795
-44.0190124511719 -1.4125919342041
5.69720506668091 56.2550811767578
-5.33053112030029 -28.6760196685791
-43.5013122558594 1.70440280437469
-16.7730884552002 -50.5043869018555
-54.7032814025879 -14.4245710372925
-16.3418769836426 25.6480083465576
-53.2781944274902 -15.987491607666
23.5342407226562 -35.0567817687988
6.20785188674927 64.8274536132812
-51.5338935852051 -10.556471824646
5.68330335617065 19.8132553100586
15.3247947692871 -3.25546836853027
-26.7605438232422 12.6324501037598
-17.1904792785645 -1.1943895816803
-5.54897451400757 63.4965934753418
35.7032318115234 12.815821647644
5.95640516281128 -67.5386276245117
-17.6672134399414 44.9177665710449
4.53430223464966 -28.5955810546875
14.6802082061768 -20.8955821990967
-14.3905334472656 11.6104049682617
9.05849838256836 74.4791107177734
-36.0064239501953 1.14881777763367
37.4585456848145 -18.2941722869873
-14.5936479568481 61.6761436462402
-20.5579738616943 -20.6917114257812
17.1725883483887 -6.50464487075806
-6.07322120666504 -24.980224609375
-3.1837592124939 13.4426889419556
6.84798860549927 -43.4136352539062
0.68505322933197 44.5727310180664
17.0947704315186 -64.4154815673828
-2.5180516242981 -64.8597412109375
8.40808582305908 -50.0519676208496
-53.5180053710938 -16.5135402679443
-9.29903316497803 51.9289207458496
23.4430599212646 -1.8377548456192
29.9761981964111 -53.7202911376953
-6.89573097229004 -49.0051002502441
17.1100425720215 -58.1905899047852
11.3685636520386 -11.402551651001
-3.10602593421936 65.9060668945312
-39.5226516723633 13.0065221786499
-1.43580591678619 69.1976776123047
-4.13788604736328 30.8474102020264
15.560097694397 -49.7316246032715
-39.9275054931641 11.8809366226196
2.13128209114075 64.247444152832
-12.0617790222168 48.4711875915527
4.42822170257568 -52.078182220459
35.5346374511719 -20.2437839508057
2.57543063163757 59.2436637878418
-39.2273483276367 13.9640045166016
-39.0242462158203 22.9863262176514
-12.306568145752 3.77273058891296
-30.7462024688721 1.61940360069275
35.7655181884766 -44.487865447998
1.52739584445953 66.1834182739258
10.0763473510742 -49.0631675720215
0.760641396045685 -60.115837097168
35.1040725708008 -53.1094245910645
4.08598947525024 84.7822799682617
36.7877464294434 -42.2259330749512
-9.05131912231445 -54.4618797302246
11.1937694549561 24.7227592468262
28.5881576538086 -40.3376121520996
19.4687156677246 -25.4935531616211
-31.2322673797607 25.4531726837158
-38.107479095459 -9.05575561523438
40.2524681091309 -55.6437835693359
-38.8869400024414 -24.7076225280762
-51.5100479125977 -23.7850437164307
14.2239561080933 -1.5298445224762
3.43313956260681 -26.457239151001
-14.7513008117676 59.8690528869629
3.2206859588623 11.7626342773438
-11.1817827224731 7.24995851516724
-13.5163869857788 34.2488327026367
-3.65907788276672 -27.7040939331055
-13.2587223052979 40.5102844238281
-25.1148815155029 -1.82188069820404
42.614875793457 -55.443660736084
-26.7906265258789 24.2013778686523
27.493932723999 -41.520191192627
0.206918761134148 -42.6391639709473
-36.6595687866211 -24.2476444244385
-3.31427526473999 -6.8505425453186
10.03773021698 -37.2256546020508
22.3766479492188 -10.2656364440918
-2.72840595245361 46.2721061706543
26.7294540405273 -30.5466651916504
38.8881492614746 -44.215404510498
14.4415721893311 -63.5295181274414
-37.0411911010742 20.3775463104248
15.9359560012817 -32.611888885498
-1.22140038013458 8.00126934051514
2.23746418952942 54.3661193847656
-32.4431686401367 5.31508731842041
-6.95743560791016 -32.4719734191895
-3.33114576339722 41.5269088745117
5.78294277191162 -27.7950344085693
-15.9254083633423 -12.0912179946899
-1.86363780498505 32.3100433349609
-18.5041275024414 -34.950080871582
-19.3557910919189 58.7318649291992
-12.6471309661865 -28.8995933532715
-29.3336219787598 -6.75573492050171
-12.5778541564941 -29.3947620391846
-0.755422353744507 21.0363597869873
-53.4739990234375 -11.9781694412231
9.53775215148926 1.18791878223419
-5.44722270965576 3.71906733512878
9.39107418060303 -62.8244514465332
25.2141914367676 -40.1923370361328
-23.5486850738525 -55.1394195556641
36.624927520752 -51.204517364502
-1.41397559642792 64.4651794433594
31.1220169067383 32.1042709350586
28.4181709289551 -61.9741172790527
41.0682106018066 -6.71240520477295
7.91067838668823 39.9108734130859
-1.29828250408173 -3.32939910888672
-9.48875617980957 72.2016906738281
-10.4276132583618 38.6326713562012
-49.5756340026855 -17.0155220031738
-1.23361301422119 85.5276947021484
-9.40144634246826 62.9727249145508
1.43499612808228 59.0431213378906
-6.21810626983643 53.1744232177734
-40.3940620422363 29.7566719055176
25.2305107116699 -40.0765686035156
14.0889110565186 -52.4261283874512
-4.53575658798218 25.081392288208
-13.8788919448853 23.9843482971191
-36.7962684631348 9.24637508392334
15.9282217025757 31.1022338867188
11.8129873275757 73.0501480102539
14.2319955825806 -63.4841346740723
-1.97849798202515 66.7942733764648
-23.7449798583984 -56.3668365478516
-48.5560188293457 7.43255281448364
-20.9436798095703 -41.8449440002441
-14.2904224395752 23.3435115814209
10.4972915649414 49.6534957885742
-3.99147343635559 -50.0040283203125
18.211742401123 -51.417423248291
-5.88881301879883 -8.80561828613281
-1.51741051673889 -47.4201202392578
4.32817792892456 31.6414756774902
-24.871789932251 -7.62801742553711
-20.4118881225586 45.7798843383789
27.4483509063721 -26.8081016540527
-5.56537246704102 -29.4588851928711
-16.0065059661865 -65.9480972290039
-5.21497488021851 -25.0645942687988
40.4045066833496 -20.6414604187012
-45.9770278930664 22.8458423614502
-26.5258178710938 -0.407515108585358
28.5873775482178 -46.7295265197754
-44.2382164001465 10.3070650100708
-50.5521125793457 5.60726404190063
-7.19392871856689 -42.993709564209
-9.5423583984375 -38.5154151916504
-43.3997917175293 -7.75084924697876
9.01693916320801 -51.2871055603027
25.0602397918701 13.2396965026855
-29.0721530914307 11.8301572799683
-27.9322357177734 -3.03503251075745
1.96551692485809 36.5476722717285
-7.27517414093018 82.8970489501953
0.202581092715263 59.3200950622559
33.6214828491211 4.04944849014282
34.3057708740234 -22.4389762878418
-23.6837158203125 -9.76546096801758
-34.6332054138184 0.0546604692935944
39.3979530334473 -44.4877548217773
-22.6435832977295 7.02360010147095
29.4733123779297 -25.9379234313965
-33.2400817871094 5.08824300765991
-38.6765251159668 4.40700483322144
9.50541591644287 18.0882816314697
10.8913688659668 -52.0722389221191
32.0797653198242 -33.75390625
-6.8299732208252 1.5664496421814
27.8091373443604 -42.6121864318848
7.99870491027832 47.2746238708496
-47.7646560668945 19.1078987121582
-0.175539493560791 -6.24241399765015
-15.9347543716431 -23.7865829467773
3.07636833190918 -4.72971487045288
37.8014335632324 -23.3688488006592
18.0060062408447 42.4501457214355
13.2788887023926 11.630054473877
-7.13648319244385 -29.9890613555908
0.0621891058981419 87.6787338256836
7.98978853225708 -35.625072479248
};
\end{axis}